\RequirePackage[hyphens]{url}
\documentclass[sigconf, nonacm]{acmart}

\usepackage{booktabs}
\AtBeginDocument{%
	\providecommand\BibTeX{{%
			\normalfont B\kern-0.5em{\scshape i\kern-0.25em b}\kern-0.8em\TeX}}}

\copyrightyear{2022}
 \acmYear{2022}
 \setcopyright{acmcopyright}\acmConference[SIGMOD '22]{Proceedings of the 2022 International Conference on Management of Data}{}{}
 \acmPrice{15.00}
 \acmDOI{}
 \acmISBN{}

\usepackage{tikz}
\usetikzlibrary{patterns}

\usepackage[multiple]{footmisc}
\usepackage{booktabs}
\usepackage{dsfont}
\usepackage[utf8]{inputenc}
\usepackage{balance}
\usepackage{siunitx}
\usepackage{multirow}
\usepackage{graphicx}
\usepackage{listings}
\usepackage{commath}
\usepackage{epstopdf}
\usepackage{color}
\usepackage{url}
\usepackage{caption}
\usepackage{alltt}
\usepackage{bbm}
\usepackage{mathrsfs}
\usepackage{float}
\usepackage{subcaption}
\usepackage{fancyvrb}
\usepackage[linesnumbered,ruled,vlined]{algorithm2e}
\usepackage{algpseudocode}

\usepackage{mathtools}
\usepackage{capt-of}
\usepackage{colortbl}
\usepackage{xcolor}
\usepackage{xfrac}
\usepackage{footnote}
 \usepackage{enumitem}
\usepackage{array}
\usepackage{arydshln}
\usepackage{cleveref}
\definecolor{mygreen}{rgb}{0,0.6,0}
\definecolor{myred}{rgb}{0.6,0,0}
\definecolor{mygray}{rgb}{0.5,0.5,0.5}
\definecolor{mymauve}{rgb}{0.58,0,0.82}
\definecolor{myblue}{rgb}{0,0,1}

\newcommand{\reva}[1]{#1}
\newcommand{\revb}[1]{#1}
\newcommand{\revc}[1]{#1}

\newcommand{\revd}[1]{#1}

\usepackage{listings}
\lstnewenvironment{VerbatimText}[1][]{
    
    \lstset{fancyvrb=true,basicstyle=\footnotesize,captionpos=b,xleftmargin=2em,#1}
}{}

\PassOptionsToPackage{hyphens}{url}\usepackage{hyperref}

\newcommand{\features}{\ensuremath{\mb X}}
\newcommand{\classifier}{\ensuremath{f}}
\newcommand{\Dom}[1]{\textsc{Dom}(#1)}
\newcommand{\labelDom}{Y}
\newcommand{\predictLabelDom}{\hat{Y}}
\newcommand{\underlyingDist}{\mc{D}}
\newcommand{\predictedLabel}{\hat{y}}
\newcommand{\datap}{\mb{x}}
\newcommand{\tpoint}{\mb{z}}
\newcommand{\alabel}{y}
\newcommand{\learner}{\mc{A}}
\newcommand{\modelDom}{\Theta}
\newcommand{\protectedAttr}{S}

\newcommand{\pat}{\phi}
\newcommand{\patset}{\Phi}
\newcommand{\patDom}{\Phi_{\dtrain}}
\newcommand{\patI}[1]{\dtrain({#1})}
\newcommand{\apatI}{\patI{\pat}}
\newcommand{\attr}[1]{\ensuremath{\texttt{\upshape{#1}}}}
\newcommand{\cnst}[1]{\ensuremath{\textsf{#1}}}
\newcommand{\supthresh}{\tau}

\DeclareMathOperator*{\argmax}{argmax}
\DeclareMathOperator*{\argmin}{argmin}

\newcommand{\pr}{{\tt \mathrm{Pr}}}
\newcommand{\sys}{\textsc{Gopher}\xspace}

\newcommand{\mc}[1]{\mathcal{#1}}

\newcommand{\ignore}[1]{}

\definecolor{black}{rgb}{0,0,0}
\definecolor{grey}{rgb}{0.8,0.8,0.8}
\definecolor{red}{rgb}{1,0,0}
\definecolor{green}{rgb}{0,1,0}
\definecolor{darkgreen}{rgb}{0,0.5,0}
\definecolor{darkpurple}{rgb}{0.5,0,0.5}
\definecolor{darkdarkpurple}{rgb}{0.3,0,0.3}
\definecolor{blue}{rgb}{0,0,1}
\definecolor{shadegreen}{rgb}{0.95,1,0.95}
\definecolor{shadeblue}{rgb}{0.95,0.95,1}
\definecolor{shadered}{rgb}{1,0.85,0.85}
\definecolor{shadegrey}{rgb}{0.85,0.85,0.85}
\definecolor{oddRowGrey}{rgb}{0.80,0.80,0.80}
\definecolor{evenRowGrey}{rgb}{0.85,0.85,0.85}
\definecolor{lightpurple}{rgb}{0.88,1.0,1.0}

\newcommand{\indep}{\mbox{$\perp\!\!\!\perp$}}

\newcommand{\RNum}[1]{\uppercase\expandafter{\romannumeral #1\relax}}
\newcommand{\mb}[1]{{\mathbf{#1}}}
\newtheorem{defn}{Definition}[section]

\newtheorem{col}[defn]{Colollary}

\newcommand{\proj}[1]{{\Pi}}
\newcommand{\sel}[1]{{\sigma}}

\newcommand{\cut}[1]{}
\newcommand{\eat}[1]{}

\usepackage[multiple]{footmisc}

\theoremstyle{remark}

\def\infinity{\rotatebox{90}{8}}
\newcommand{\bias}{\mathcal{F}}
\newcommand{\lossz}{L}
\newcommand{\lossD}{\mathcal{L}}
\newcommand{\hessian}{\mathcal{H}_\theta}
\newcommand{\grad}{\nabla_\theta}

\newcommand{\dtrain}{\mb D}
\newcommand{\dtest}{\mb D_{test}}
\newcommand{\dinter}{\mb{S}}
\newcommand{\udtrain}{\dtrain^{\bar{\mb S}}} %\dtrain^{\bar{\mb S}
\newcommand{\upar}{\theta_{\bar{\mb S}}} %
\newcommand{\influence}{\mathcal{I}}
\newcommand{\responsibility}{\mathcal{R}}
\newcommand{\params}{\ensuremath{\theta}}
\newcommand{\oparam}{\ensuremath{\theta}^*}
\newcommand{\support}{Sup}
\newcommand{\score}{U}
\newcommand{\topk}[1]{\textsc{top-}{#1}}
\newcommand{\cont}{C}
\newcommand{\cthresh}{c}

\SetKwInput{KwInput}{Input}
\SetKwInput{KwOutput}{Output}

\newcommand{\partitle}[1]{\par\noindent\textbf{{#1}.}}
\newcommand{\parsubtitle}[1]{\smallskip\noindent\textit{{#1}}}
\newcommand{\thead}[1]{\cellcolor{black}\textbf{\textcolor{white}{#1}}}

\newcommand{\latticelabels}[1]{\tikz[baseline=(X.base)]\node [fill=black!20, draw, rectangle, font=\sffamily, inner sep=2pt] (X) {#1};}

\renewcommand\footnotetextcopyrightpermission[1]{} % removes footnote with conference information in first column

\author{Romila Pradhan\textsuperscript{*}}
\affiliation{\institution{Purdue University}\city{West Lafayette}\state{IN}\country{USA}}
\email{rpradhan@purdue.edu}

\author{Jiongli Zhu }
\affiliation{\institution{University of California, San Diego}\city{La Jolla}\state{CA}\country{USA}}
\email{jiz143@ucsd.edu}

\author{Boris Glavic}
\affiliation{\institution{Illinois Institute of Technology}\city{Chicago}\state{IL}\country{USA}}
\email{bglavic@hawk.iit.edu}

\author{Babak Salimi}
\affiliation{\institution{University of California, San Diego}\city{La Jolla}\state{CA}\country{USA}}
\email{bsalimi@ucsd.edu}

\thanks{Work done while the author was affiliated with the University of California, San Diego.}

\newbool{Techreport}
\newrobustcmd{\ifnottechreport}[1]{\ifbool{Techreport}{}{#1}}
\newrobustcmd{\iftechreport}[1]{\ifbool{Techreport}{#1}{}}

\boolfalse{Techreport}

\begin{document}
%\fancyhead{}
\title{Interpretable Data-Based Explanations for Fairness Debugging}

%\titlenote{Produces the permission block, and
%  copyright information}
%\subtitlenote{The full version of the author's guide is available as
 % \texttt{acmart.pdf} document}

%\renewcommand\footnotetextcopyrightpermission[1]{} % removes footnote with conference information in first column
%\pagestyle{plain}

% The default list of authors is too long for headers.
%\renewcommand{\shortauthors}{B. Trovato et al.}

\begin{abstract}
A wide variety of fairness metrics and eXplainable Artificial Intelligence (XAI) approaches have been proposed in the literature to identify bias in machine learning models that are used in critical real-life contexts.
However, merely reporting on a model's bias, or generating explanations using  existing XAI techniques  is insufficient to locate and eventually mitigate sources of  bias.
We introduce \sys, a system that % , unlike others,
produces {\em  compact}, {\em interpretable} and {\em causal explanations} for bias or unexpected model behavior by identifying {\em coherent subsets of the training data} that are {\em root-causes} for this behavior.
Specifically, we introduce the concept of {\em causal responsibility}
that quantifies the extent to which {\em intervening} on training data by {\em removing or updating subsets of it} can resolve the bias.
Building on this concept, we develop an efficient approach for generating the top-$k$ patterns that explain model bias that utilizes techniques from the machine learning (ML) community to approximate causal responsibility 
and uses pruning rules to manage the large search space for patterns. Our experimental evaluation 
 demonstrates the effectiveness of \sys in generating interpretable explanations for identifying and debugging sources of bias.
\end{abstract}

% %%%% to enable/disable page number and ACM format
% \settopmatter{printfolios=true}
\settopmatter{printacmref=false}

\maketitle

\section{Introduction}
\label{sec:intro}
% 1st paragraph: Explainability and explainable AI.
Machine learning (ML) is being increasingly applied to decision-making in sensitive domains such as finance, healthcare, crime prevention and justice management. Although it  has the potential to overcome undesirable aspects
of human decision-making, concern continues to mount that its opacity perpetuates systemic biases and discrimination reflected in training data~\cite{fb-housing, self-driving-cars,10.1145/2702123.2702520}.  This concern
% gives rise to
has caused
increasing regulations and demands for generating human understandable explanations for the behavior of ML algorithms--- an issue addressed by the rapidly growing field of \textit{eXplainable Artificial Intelligence} (\textit{XAI}); see \cite{molnar2020interpretable} for a recent survey.

The development of XAI methods is motivated by technical, social and ethical objectives~\cite{kasirzadeh2021reasons, datta2016algorithmic, binns2018s, lepri2018fair, kemper2019transparent} including: 
(1)~providing users with actionable insights to change the results of algorithms in the future, (2)~facilitating the identification of sources of harms such as bias and discrimination, and (3)~providing the ability to debug ML algorithms and models by identifying errors or biases in training data that result in adverse and unexpected behavior.

Most XAI research to date has focused on generating {\bf feature-based explanations},  which quantify the extent to which input feature values contribute to an ML model's predictions. In this direction, methods based on feature importance quantification~\cite{lipovetsky2001analysis, vstrumbelj2014explaining,
lundberg2017unified,lundberg2018consistent,datta2016algorithmic,merrick2019explanation,frye2019asymmetric,aas2019explaining}, surrogate models, causal and counterfactual methods~\cite{lundberg2017unified,ribeiro2016should,ribeiro2018anchors}, and logic-based approaches~~\cite{shih2018symbolic,ignatiev2020towards,darwiche2020reasons} are designed to reveal the dependency patterns between the input features and output of ML algorithms. These methods differ in terms of whether they address correlational, causal, counterfactual or contrastive patterns. While such explanations satisfy the aforementioned objectives~(1) and (2) if we only consider the test data as a source of bias, they fall short in generating diagnostic explanations that let users trace unexpected or discriminatory algorithmic behavior back to its {\em training data}. Hence, they fail to satisfy objective~(3) if the training data is the source of the bias.
% and do not fulfill objective (4). 
Indeed, information solely about output-input feature dependency is insufficient % to arm developers and practitioners with actionable diagnostic explanations that
to explain discriminatory and unexpected algorithmic decisions in terms of {\em data errors and biases} introduced during data collection and preparation. While a feature-based explanation can identify which features of a test data point are correlated with a misprediction or bias, it does not explain why the model exhibits this bias. To explain such behavior, we must % dig deeper and
trace the bias back to the data used to train the  model. For example, feature-based approaches cannot generate explanations of the form: \textit{``The primary source of gender bias for this classifier, which decides about loan applications, is its training data which  exhibits bias against the credit scores of unmarried females who are house owners.''}

%%%%%%%%%%%%%%%%%%%%%%%%%%%%%%%%%%%%%%%%
\begin{figure*}[t]
    \centering
    % \includegraphics[scale=0.37]{figs/example.pdf}\\[-2mm]
        %%%%%%%%%%%%%%%%%%%%%%%%%%%%%%%%%%%%%%%%
    \begin{subfigure}{0.34\linewidth}
      \includegraphics[width=1.0\linewidth]{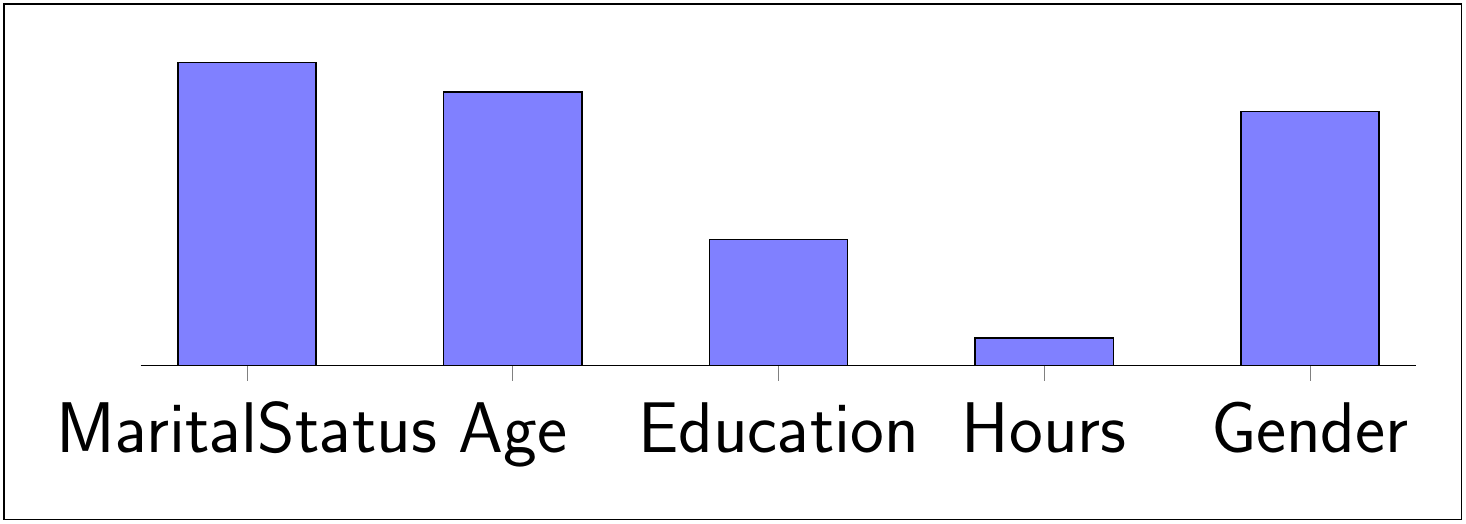}
      \caption{Feature-based explanations}
    \end{subfigure}
    %%%%%%%%%%%%%%%%%%%%%%%%%%%%%%%%%%%%%%%%
    \begin{subfigure}{0.65\linewidth}
      \includegraphics[width=1.0\linewidth]{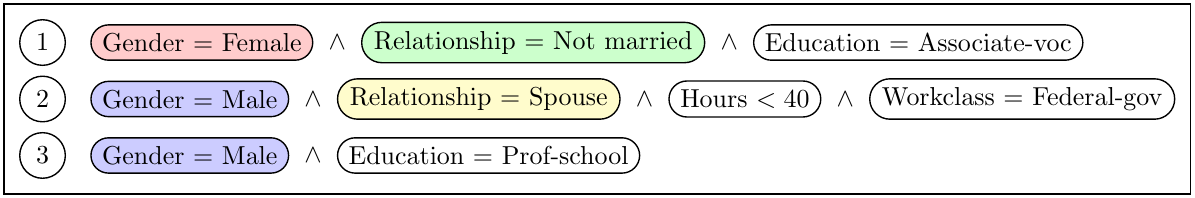}
      \caption{Top-k explanations generated by \sys}
    \end{subfigure}
    %%%%%%%%%%%%%%%%%%%%%%%%%%%%%%%%%%%%%%%%
    \begin{subfigure}{0.34\linewidth}
      \includegraphics[width=1.0\linewidth]{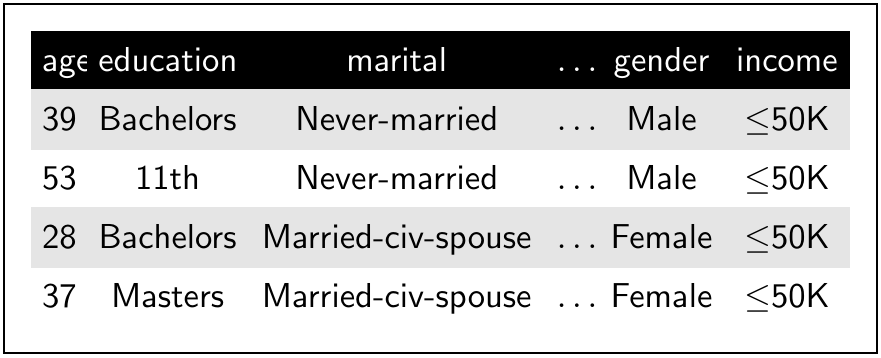}
      \caption{Instance-based explanations}
    \end{subfigure}
    \begin{subfigure}{0.65\linewidth}
      \includegraphics[width=1.0\linewidth]{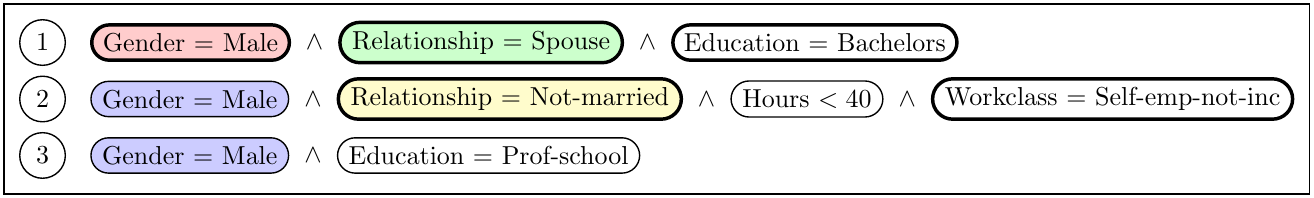}
      \caption{Update-based explanations for the corresponding top-k explanations}
    \end{subfigure}
    %%%%%%%%%%%%%%%%%%%%%%%%%%%%%%%%%%%%%%%%
    \vspace{-3mm}
    \caption{An overview of explanations generated by XAI approaches for an ML algorithm built using the UCI Adult dataset (see \Cref{sec:exp} for details), which predicts if an individual earns $\geq50$K/year. (a)~Feature-based explanations rank attributes in descending order of importance toward model behavior. (c) Instance-based explanations output a list of data points most responsible for model behavior. \sys\ generates two kinds of explanations: (b)~top-$k$ explanations identify the patterns most \textit{causally} responsible for model bias, and (d)~updates to the top-k explanations that reduce the model's bias by the most.}
    \label{fig:system}
    \vspace{1mm}
\end{figure*}
%%%%%%%%%%%%%%%%%%%%%%%%%%%%%%%%%%%%%%%%

In this paper, we take the first step toward developing a framework
for generating {\em diverse}, {\em compact} and {\em interpretable} {\bf training data-based explanations} that identity which  parts of the training data are responsible for an unexpected and discriminatory behavior of an ML model. Specifically, we introduce \sys, a system that finds \textit{patterns} which compactly describe cohesive sets of training data points that, when eliminated, reduce the bias of the model.
In other words, {\em if the ML algorithm had been trained on the modified training data, it would not have exhibited the unexpected or undesirable behavior or would have exhibited this behavior to a lesser degree.} Explanations generated by our framework, which complement existing approaches in XAI,  are crucial for helping system developers and ML practitioners to debug ML algorithms by identifying data errors and bias in training data, such as {\em measurement errors} and {\em misclassifications}~\cite{gianfrancesco2018potential,jacobs2021measurement,wang2021fair}, {\em data imbalance}~\cite{farrand2020neither}, {\em missing data} and {\em selection bias}~\cite{fernando2021missing,martinez2019fairness,mehrabi2019survey}, covariate shift~\cite{rezaei2020robust,singh2021fairness},
{\em technical biases} introduced during data preparation~\cite{Stoyanovich2020ResponsibleDM}, and {\em poisonous data} points injected through {\em adversarial attacks}~\cite{solans2020poisoning, jagielski2020subpopulation,mehrabi2020exacerbating,goel2021importance}. It is known in the algorithmic fairness literature that information about the source of bias is critically needed to build fair ML algorithms because none of the current bias mitigation solutions fits all situations~\cite{singh2021fairness, wang2021fair,farrand2020neither,goel2021importance,Fu2020AIAA}. %We illustrate using an example:

We use the example shown below to illustrate the difference between feature- and data-based techniques, that explain bias based on the contribution of a subset of the training data.

% %%%%%%%%%%%%%%%%%%%%%%%%%%%%%%%%%%%%%%%%

%%%%%%%%%%%%%%%%%%%%%%%%%%%%%%%%%%%%%%%%
\begin{example}\label{ex:example1}
  Consider a classifier that predicts whether individuals described by their attributes (such as gender, education level, marital status, working hours, etc.) earn more than $50$K a year. A system developer uses the classifier to predict the income of individuals and, while analyzing the results, finds an unexpected negative result (earning less than $50$K) for a female user:
  \vspace{-.3cm}
  \begin{center}
    \resizebox{1\linewidth}{!}{
      \begin{tabular}{|c|c|c|c|c|c|c|}
        \thead{age} & \thead{education} & \thead{marital} & \thead{workclass} & \thead{race} & \thead{gender} & \thead{hours} \\
        34 & Bachelors& Never-married& Private& Black& Female& 40 \\ \hline
      \end{tabular}
    }
  \end{center}

Based on her education ($\attr{education}=\text{Bachelors}$), marital status ($\attr{marital}=\text{Never-married}$), (\attr{age}=$34$), and other features, this user should be classified as earning more than $50$K. Upon closer examination, the developer realizes that the classifier violates the commonly used fairness metric of statistical parity with respect to the protected group $\attr{gender}=Female$. Statistical parity requires that the probability of being classified as the positive class (earning more than $50$K in our example) is the same for individuals from the protected group as it is for individuals from the privileged group (we present fairness definitions in \Cref{sec:prelim}).

 The developer uses an existing XAI package, such as LIME~\cite{ribeiro2016should} or SHAP~\cite{lundberg2017unified}, to explain the algorithm's output for the user and learns that the user's gender and unmarried status were most responsible for the faulty prediction. We show examples of such explanations in \Cref{fig:system}(a) and (c).
These explanations, however, do not help the developer identify the source of the model's bias. In contrast, \sys identifies cohesive subsets of the training data that are most responsible for the bias and compactly describes the subsets using patterns that are conjunctions of predicates. \Cref{fig:system}(c) shows the top-3 explanations \sys produces. Pattern $\pat_2$ states that a sub-population that strongly impacts model bias is male spouses who work in the federal government for  less than 40 hours per week. Indeed, in the Adult Income dataset (see \Cref{sec:exp}), this subpopulation is the major source of gender bias, because it reports household income; hence, due to the bias in data collection,  married males becomes highly correlated with high-income. In addition to producing such explanations, \sys also identifies homogeneous updates for sub-populations that can maximally improve model bias. \Cref{fig:system}(d) shows one such update (with bold outline). For instance, model bias would significantly decline if we changed $\attr{relationship}$ status from ``Spouse'' to ``Not married'' and $\attr{Workclass}$ from federal employee to self-employed in the subset of the training data corresponding to $\pat_2$. Such updates mean that our approach not only identifies which \textit{subsets of training data} are responsible for the bias, but also which  \textit{features of the data points in these subsets} are responsible.
% age=34, education=Bachelors, marital=Never-married, workclass=Private, race=Black, gender=Male, hours=40
%\babak{The examples should illustrate senarios in %which bias is due to data quality issues such as messurment bias, misclassification or data poisoning attacks. You can show both an individual for whom the model returns totally unexpected result and show the model violates both euqlized odds and demographoc parity. Then show that feature attribution methods such as LIME, SHAP, Lweis and counterafctual explanations can not generate interesting and insightful local and global explanations that could assist system developer to undestand the source of bias. Finally, show explanations generated by your systems addrsse this shortcomming.}
\end{example}
%%%%%%%%%%%%%%%%%%%%%%%%%%%%%%%%%%%%%%%%

%
%%%%%%%%%%%%%%%%%%%%%%%%%%%%%%%%%%%%%%%

%%%%%%%%%%%%%%%%%%%%%%%%%%%%%%%%%%%%%%%%%
\partitle{Root causes and responsibility}
Our first contribution is to formalize {\em root causes} of bias and the {\em causal responsibility} of a subset of the training data for the bias of an ML model. Responsibility is measured as the difference
between the bias of the original model and that of a new model trained on the {\em modified} training data, obtained by either removing or updating the subset.
We define data-based explanations for bias as the top-k subsets of the training data that have the highest causal responsibility toward model bias.

%%%%%%%%%%%%%%%%%%%%%%%%%%%%%%%%%%%%%%%%
\partitle{Pattern-based explanations} Such ``raw explanations" based on subsets of data points may overwhelm rather than inform end-users. More compact and coherent descriptions % of specific subsets of the  training data
% responsible for model bias
are needed. Furthermore, sources of bias and discrimination % due to data errors
in training data are typically not randomly distributed across different sub-populations; rather they manifest systematic errors in data collection, selection, feature engineering, and curation~\cite{gianfrancesco2018potential,jacobs2021measurement,wang2021fair, fernando2021missing,martinez2019fairness,mehrabi2019survey,parikh2019addressing}. In other words, more often than not, \textit{certain cohesive training data subsets are responsible for bias}.
% is often a manifestation
% the fact that data errors are , thereby certain cohesive subsets of training data are responsible for bias.
% Therefore, o
Our system generates explanations in terms of {\em compact patterns} expressed as first-order predicates that describe homogenous training data subsets.
By identifying commonalities within training data subsets, which are the main contributors to model bias, such explanations unearth the root causes of bias. To ensure that the generated explanations are
{\em diverse}, our system filters explanations that refer to similar subsets of training data.

\partitle{Computational challenges} Computing such explanations is challenging for two main reasons. First, computing the causal responsibility of a given subset of data requires retraining the ML model, which is expensive.  Second, it is computationally prohibitive to explore all possible subsets of training data to identify the top-$k$ explanations. To address the first challenge, we trace the model's predictions through its learning algorithm back to the training data. To do so, we approximate the change in model parameters by using either {\em influence functions}~\cite{cook-influence} or by assuming that the updated model parameters can be computed by applying a {\em single gradient decent step} to the original model. First-order (FO) approximations of influence functions have recently been proven useful in ML model debugging~\cite{pmlr-v70-koh17a} by identifying the top-k training data points responsible for model mis-predictions. However, they do not effectively capture the ground truth influence of training data subsets because of correlations between data points. To better approximate the change in model parameters, we utilize \textit{second-order (SO) influence functions}, which were shown to better correlate with ground truth subset influence~\cite{icml2020_4261}. While even SO influence functions can provide relatively poor estimates of the influence of large portions of the training data, the approximation error of SO influence functions is typically much better for coherent sets of data points as described by our patterns.
%we benefit from the facts that
% (1) patterns describe coherent subsets of the training data, and (2) the approximation error of SO influence functions is typically much better for such coherent sets of data points.
To address the second challenge, we develop a \textit{lattice-based search algorithm} based on ideas from frequent itemset mining~\cite{10.5555/645920.672836} to discover frequent patterns. Our algorithm identifies coarse-grained subsets of training data that are \textit{influential} and successively refines them to discover smaller influential subsets. We introduce a novel \textit{quality metric} for explanations, and optimizations that further prune the search space and efficiently generate the top-k explanations.

\partitle{Update-based explanations} \sys is unique in that, in addition to generating explanations in terms of data removal, it also considers explanations based on updating data points. This aspect of \sys is motivated by the observation that not all features of a data point are erroneous and responsible for the bias. % We generate explanations based on updates (potentially on a subset of features) to reduce bias.
% Specifically,
To identify an update that mitigates bias,
we search for a \textit{homogeneous} perturbation of some of the feature of the training data points described by a pattern (explanation) that leads to a maximal reduction in model bias. By homogeneous we mean that all data points in the subset are updated using the same perturbation vector, e.g., we change the marital status from \emph{never-married} to \emph{divorced}. % (i.e., all data points described by the pattern are uniformly perturbed).
We formulate the problem of computing {\em update-based explanations} as a constrained optimization problem that searches for a homogeneous update to a subset of training data in order to maximize bias reduction, and we solve it using {\em projected gradient descent}~\cite{Bubeck2015ConvexOA}.

\noindent We make the following contributions:
\begin{itemize}[leftmargin=*]
\item We formalize \textit{interventions} on training data that remove or update training data points and estimate their effect on ML model bias. Utilizing intervention, we define \textit{predicate-based causal explanations} to identify training data subsets that contribute significantly to ML model bias (\Cref{sec:problem}).
\item We approximate the effect of training data deletion on model bias (\Cref{sec:intervention}) to avoid % computationally
expensive % model
retraining.
    \item We propose a principled approach for generating top-k explanations based on frequent \textit{pattern mining} and use \textit{pruning techniques} to reduce the size of the  search space (\Cref{sec:topk}).
    \item We introduce \textit{update-based, actionable  explanations}. These explanations identify not only which parts of the training data are responsible for the bias but also how to reduce or "repair" the bias. We formalize the problem of finding the most effective explanation as a constrained optimization problem that maximizes bias reduction. We propose a new algorithm based on projected gradient descent to determine the best \textit{homogeneous} update for a responsible training data subset (\Cref{sec:repair}).
    \item We conducted extensive experiments using % three
      real-world datasets to evaluate: (1)~the end-to-end performance of \sys, (2)~the accuracy of influence estimation techniques, and (3)~the quality and interpretability of  explanations  (\Cref{sec:exp}). We show that \sys generates explanations that are consistent with insights from existing literature.
    % \boris{Need to be a bit careful with how to phrase the last point, even though it is important to state}\boris{Add major take-aways from the experimental results here}
\end{itemize}

%%%%%%%%%%%%%%%%%%%%%%%%%%%%%%%%%%%%%%%%
\cut{
\begin{figure}
  \centering
  \begin{tabular}{|l|l|} \hline
    \textbf{Fairness Metric}                                 & \textbf{Description}   \\ \hline
    % \rowcolor{black!20}
    Statistical Parity \cite{calders2009building}            & $S \indep \hat{Y}$     \\
    a.k.a.  Demographic Parity \cite{dwork2012fairness}      &                        \\ \hline
    % \rowcolor{black!20}
    Equalized Odds (EO) \cite{hardt2016equality}             & $S \indep\hat{Y}| Y $  \\
    a.k.a.  Disparate Mistreatment  \cite{zafar2017fairness} &                        \\ \hline
    Predictive Parity (PP)\cite{chouldechova2017fair}        & $S \indep Y| \hat{Y} $ \\
    a.k.a. Outcome Test  \cite{simoiu2017problem}            &                        \\\hline
  \end{tabular}                                                                       \\[-3mm]
  \caption{\textmd{Fairness notions and their description in terms of the conditional independence statement.}} \label{tbl:asdef}
\end{figure}
}
%%%%%%%%%%%%%%%%%%%%%%%%%%%%%%%%%%%%%%%%

%%%%%%%%%%%%%%%%%%%%%%%%%%%%%%%%%%%%%%%%%%%%%%%%%%%%%%%%%%%%%%%%%%%%%%%%%%%%%%%%
\vspace{-2mm}
\section{Background and Assumptions}
\label{sec:prelim}
We now present pertinent background information on classification and algorithmic fairness. % and formulate the problem of generating training data-based explanations for bias.

%\subsection{Preliminaries and Background}
%\label{sec:notation}
% \noindent\textbf{Notation.}
%%%%%%%%%%%%%%%%%%%%%%%%%%%%%%%%%%%%%%%%%%%%%%%%%%%%%%%%%%%%%%%%%%%%%%%%%%%%%%%%
\partitle{Classification}
We consider the problem of {\em binary classification}, which is the focus of most literature on algorithmic fairness. Consider a training dataset of $n$ examples $\dtrain=\{\tpoint_i = (\datap_i, \alabel_i)\}_{i=1}^n$, with domain $\Dom{\features} \times \Dom{Y}$, drawn from an unknown data distribution $\underlyingDist$, where $\features$  denotes a set of discrete and continuous features and $\labelDom = \{0, 1\}$ is a binary label to be predicted. Here, the goal is to find a classifier  $\classifier: \Dom{\features} \to \predictLabelDom$ that associates each data point $\datap$ with a \textit{predicted label} $\predictedLabel \in \predictLabelDom = \{0,1\}$ that maximizes accuracy $E_{\datap, \alabel \sim  \mc{D}}[\mathbb{1}(\classifier(\datap)=\alabel)]$, i.e., the expectation of the fraction of data points from the unknown data distribution $\underlyingDist$ that are correctly predicted by $\classifier$.
%we consider classifiers that are functions $\classifier_{\params}$ that are parameterized by a \textit{model} $\params$ from a space of possible models $\modelDom$
Note that since we have access to only a sample $\dtrain$ of $\underlyingDist$, model performance is determined over the training data as a substitute for calculating
$E_{\datap, \alabel \sim  \mc{D}}[\mathbb{1}(\classifier(\datap)=\alabel)]$.
% $f_{\theta}(x)$, where $f: f_{\theta}(x) \rightarrow \hat{Y}$ is a function that is parameterized by $\theta$ and $\predictedLabel$ denotes the {\em predicted} class label $(Dom(\hat{Y}) =\{0, 1\})$, which correctly classifies as many data points as possible, .
Typically, $f(.)$ is a member of a family of functions $f_{\params}$ that are parameterized by $\theta$ with domain $\modelDom$. A learning algorithm $\learner$ uses $\dtrain$ and learns parameters $\oparam \in \modelDom$ that maximize the {\em empirical accuracy} $\sum_{i=1}^{n} \mathbb{1}(\classifier_{\params}(\datap)=\alabel)$. Learning algorithms typically use some {\em loss function} $\lossz(\tpoint, \params)$ and minimize the {\em empirical loss} $\lossD(\dtrain, \params) = \frac{1}{n} \sum_{i=1}^{n} \lossz(\tpoint, \params)$. We denote the optimal parameters $\oparam$ of a classifier  as $\params$ when it is clear from the context. \revc{Throughout this paper, we consider
% only
learning algorithms that use a loss function $\lossD(\theta)$ that is twice-differentiable,
% and strictly convex in $\theta$
which covers % several
classification algorithms such as % as those based on
logistic regression~\cite{sklearn}, support vector machines (SVM)~\cite{sklearn}, and feedforward neural networks~\cite{fastai}.}

\partitle{Algorithmic fairness}
Consider a binary classifier $\classifier_{\params}$ with output $\hat{Y}$ and a {\em protected} attribute $\protectedAttr \in \features$, such as gender or race. Without loss of generality, we interpret $\predictLabelDom=1$ as a {\em favorable} (positive) prediction and $\predictLabelDom =0$ as an {\em unfavorable} (negative) prediction. To simplify the exposition, we assume $\Dom{\protectedAttr} = \{0, 1\}$, where $\protectedAttr=1$ indicates a {\em privileged} and $\protectedAttr=0$ indicates a {\em protected} group (e.g., males and non-males, respectively). Algorithmic fairness aims to ensure that the classifier $\classifier$ makes fair predictions devoid of discrimination with respect to the protected attribute(s).  Many fairness definitions have been proposed to quantify the bias of a binary classifier (see~\cite{10.1145/3194770.3194776} for a recent survey). The best-known ones are based on \emph{associative} relationships between the protected attribute and the classifier's outcome. These definitions are unlike {\em causal notions} of fairness, which incorporate background knowledge about the underlying data-generative process and define fairness in terms of the causal influence of the protected attribute on the classifier's outcome~\cite{kusner2017counterfactual,kilbertus2017avoiding,nabi2018fair,salimi2019interventional}.
% In this paper,
We focus on three of the most widely used associational notions of fairness:

%(summarized in Table~\ref{tbl:asdef})
%%%%%%%%%%%%%%%%%%%%%%%%%%%%%%%%%%%%%%%%
\parsubtitle{Statistical parity} requires an algorithm to classify both
the protected and privileged group with the same probability:
\[\pr(\predictLabelDom=1|\protectedAttr=1)=\pr(\predictLabelDom=1|\protectedAttr=0).\]

%%%%%%%%%%%%%%%%%%%%%%%%%%%%%%%%%%%%%%%%
% \parsubtitle{Equal opportunity} requires both protected and privileged groups to have the same false negative rate: \[\pr(\predictLabelDom=0|\labelDom=1, \protectedAttr=1) = \pr(\predictLabelDom=0|\labelDom=1, \protectedAttr=0).\]
\parsubtitle{Equal opportunity} requires both the protected and privileged group to have the same true positive rate: \[\pr(\predictLabelDom=1|\labelDom=1, \protectedAttr=1) = \pr(\predictLabelDom=1|\labelDom=1, \protectedAttr=0).\]

%%%%%%%%%%%%%%%%%%%%%%%%%%%%%%%%%%%%%%%%
\parsubtitle{Predictive parity} requires that both protected and privileged groups have the same predicted positive value (PPV), i.e., correct positive predictions have to be independent of the protected attribute:
 \[\pr(\labelDom=1|\predictLabelDom=1, \protectedAttr=1) = \pr(\labelDom=1|\predictLabelDom=1, \protectedAttr=0).\]

%\boris{Is that correct to say?:} BABAK: YES
In practice, these fairness notions are often evaluated by estimating the probabilities on a test dataset $\dtest$. Note that % the techniques developed in this paper
our techniques also work for other associational and causal notions of fairness.

%%%%%%%%%%%%%%%%%%%%%%%%%%%%%%%%%%%%%%%%%%%%%%%%%%%%%%%%%%%%%%%%%%%%%%%%%%%%%%%%
\section{Problem Statement}
\label{sec:problem}
In this section, we introduce \textit{root causes} for bias and
%\sandy{Should the following say "removal-based" explanations?}
define data-based explanations for the bias of a classifier. For now, we focus on generating explanations based on removal of data points from training data.  In \Cref{sec:repair}, we extend our framework to support explanations which update data points.
% , not  removing them.

Consider a training dataset $\dtrain$, a learning algorithm $\learner$, a classifier $\classifier_{\params}$ trained by running $\learner$ on $\dtrain$, and a fairness metric (\Cref{sec:prelim})  $\bias: \params \times  \dtest \rightarrow \mathbb{R}$ which we refer to as {\em bias}. Bias quantifies the {\em fairness violation} of the classifier on a testing data $\dtest$ that is unseen during the training process, such that if $\bias(\params, \dtest) > 0$, the classifier is {\em biased}, and is {\em unbiased} otherwise. For instance, for statistical parity we may define $\bias$ as $\pr(\predictLabelDom=1|S=1)-\pr(\predictLabelDom=1|S=0)$, where $\pr(\predictLabelDom=1|S=s)$, for $s \in \{0,1\}$, is estimated on $\dtest$ using empirical probabilities. % First, we describe how we quantify the degree to which a training data point, or a group thereof, is responsible for a classifier's bias.

\partitle{Intervening on training data}
We evaluate the effect of a subset of the training data $\dinter \subseteq \dtrain$ on the bias of a classifier $\classifier_{\params}$ using an {\em intervention} that removes $\dinter$ from $\dtrain$. We then assess whether the removal reduces the bias of a {\em new classifier} trained on the adjusted ("intervened") training data. More specifically, let $\udtrain=\dtrain \setminus \dinter$ be the intervened training data and $\upar$ be the new model trained on $\udtrain$ (using the same learning algorithm $\learner$). The effect of $\dinter$ on the bias of $\classifier_{\upar}$ can be measured by comparing the bias of the original and updated classifiers, i.e., by comparing $\bias(\params, \dtest)$ and $\bias(\upar, \dtest)$.
% We are now ready to define the root causes of classifier bias.
%\babak{Ions.}
% For ease of notation, we represent the bias of an algorithm in terms of its parameters and the dataset over which the algorithm makes predictions, i.e.,  instead of $\mathcal{F}(\mb T, \mathcal{A}(T))$, we write $\mathcal{F}_{\mb T, \theta}$.  $\mathcal{F}_{\theta, \mb T}$
% For more details on algorithmic fairness removed update based intervetions above because it is not clear how it could be used to define root causes if the update vector is not specified. I think update-based explanations should be defined either lated in this section or in the their own dedicated secti and the different fairness definitions, please see Section~\ref{sec:fairness-definitions}.

%\babak{It is not clear where $T$ comes from all of a suden. Once you define fairness in the baklcground you can sat your formulations can caltures all notions of fairness we are inetrsted in in this paper and also say you esstimate these fairness violations on test data. This is indeed emperical violation of fainress definitions on test data.}

%%%%%%%%%%%%%%%%%%%%%%%%%%%%%%%%%%%%%%%%
\begin{definition}[\textbf{Root cause of bias}]
 A subset $\dinter \subseteq \dtrain$ of the training data $\dtrain$ is a \textit{root cause} of the bias of a classifier $\classifier_{\params}$ if:
    \[0 \leq  \bias({\upar, \dtest}) < \bias({\params, \dtest}),\]
where  $\upar$ is a classifier trained on $\udtrain=\dtrain \setminus \dinter$.  \end{definition}
%%%%%%%%%%%%%%%%%%%%%%%%%%%%%%%%%%%%%%%%

%\babak{I stopped here. I can;t digest the following definition. $f$ comes out of the blue. revise teh following for bias and then say in words you define the same messures for accuricy and loss as well.}
Next, we introduce a metric to quantify the causal responsibility of a training data subset toward classifier bias.

%%%%%%%%%%%%%%%%%%%%%%%%%%%%%%%%%%%%%%%%
\begin{definition}[\textbf{Causal responsibility}]\label{def:responsibility}
We measure the \textit{causal responsibility} of a training data subset $\dinter \subseteq \dtrain$ toward the bias $\bias$ of a classifier $\classifier_\params$ w.r.t. a fairness metric as
\[
  \responsibility_{\bias}(\dinter) = \dfrac{\bias\big(\params , \dtest \big) - \bias\big(\upar, \dtest \big)}{\bias\big(\params, \dtest\big).}
\]
\end{definition}
%%%%%%%%%%%%%%%%%%%%%%%%%%%%%%%%%%%%%%%%

Intuitively, causal responsibility captures the relative difference between the bias of the original model and that of the new model obtained by intervening on training data. It quantifies the degree of contribution $\dinter$ has on model bias. Note that $-\infinity < \responsibility_\bias(\dinter) < 1$. If $\dinter $ is a root cause of bias, then $\bias\big(\params , \dtest\big) - \bias\big(\upar, \dtest \big)>0$; hence,  $0 <\responsibility_{\bias}(\dinter)<1$. The larger the value of $\responsibility_{\bias}$, the greater
%\sandy{Babak:  can you fix the inappropriate italics below?}
the responsibility of $\dinter$
toward bias. If $\responsibility_\bias(\dinter) \leq 0$, then removing $\dinter$ from training data either does not change bias or further exacerbates it.
We define data-based explanations based on causal responsibility. To
formalize the notion of pattern-based explanations, we first introduce patterns and define an interestingness score for patterns.

%%%%%%%%%%%%%%%%%%%%%%%%%%%%%%%%%%%%%%%%
\begin{definition}[\textbf{Pattern}]\label{def:pattern}
 A {\em pattern} $\pat$ is a conjunction of predicates $\pat=\bigwedge_i \pat_i$ where each $\pat_i$ is % an atomic predicate
 of the form $ [X \text{ \texttt{op} } c]$, where $X \in \features$ is a feature, $\cthresh$ is a constant and \texttt{op} $\in \{=, <, \leq, >, \geq\}$.
\end{definition}
%%%%%%%%%%%%%%%%%%%%%%%%%%%%%%%%%%%%%%%%

Patterns represent training data subsets. For example, the pattern
\[
  \pat = (\attr{gender}=\cnst{`Female'}) \land (\attr{age} < 45)
\]
describes data points where \attr{gender} is \cnst{`Female'} and \attr{age} is less than $45$. We use $\patDom$ to denote the set of all patterns defined over the training data $\dtrain$ and denote the set of data points satisfying pattern $\pat$ by $\apatI$.

%%%%%%%%%%%%%%%%%%%%%%%%%%%%%%%%%%%%%%%%
\begin{definition}[\textbf{Support of a pattern}]\label{def:pat-support}
  The {\em support} of pattern $\pat$ is denoted by $\support(\pat)$ and is defined as the fraction of data points that satisfy $\pat$, i.e.,
  \[
    \support(\pat) = \dfrac{|\apatI|}{|\dtrain|.}
  \]
\end{definition}
%%%%%%%%%%%%%%%%%%%%%%%%%%%%%%%%%%%%%%%%

% \begin{definition}
%     \textbf{(Utility of a subset)} The utility of a subset of data instances $\mb S \in \mb D$
%     % over fairness metric $\mathcal{F}$
%     is measured as:
%     $$ U(\mb S) = \dfrac{1}{\mid S \mid} \cdot \left(\dfrac{\mathcal{F}_{\mb D'}- \mathcal{F}_{\mb D}}{\mathcal{F}_{\mb D}}\right) \cdot \left(\dfrac{L(\mb D, \theta)}{L(\mb D', \theta') - L(\mb D, \theta)}\right)$$
%     where $\mb D'$ represents the altered database instance obtained either by removing data instances in $\mb S$ (i.e., $\mb D' = \mb D \setminus \mb S$) or by updating data instances in $\mb S$ (i.e., $\mb D' = \mb S^*$), and $\theta'$ represents the parameters of the model trained over $\mb D'$.
% \end{definition}

%%%%%%%%%%%%%%%%%%%%%%%%%%%%%%%%%%%%%%%%
\begin{definition}[\textbf{Interestingness of a pattern}]\label{def:interestingness}
Given a biased classifier $\classifier_\params$ and a pattern $\pat$, we define the {\em interestingness} of $\pat$ to explain the bias of the classifier $\classifier_\params$ as:
    % Rf -ve, Rl -ve best
    % Rf -ve, Rl +ve ok
    % Rf +ve, Rl -ve bad
    % Rf +ve, Rl +ve worst
    \begin{equation*}
        \score(\pat) = \dfrac{\responsibility_{\bias}\big(\apatI \big)}{\support(\pat).}
    \end{equation*}
\end{definition}
%%%%%%%%%%%%%%%%%%%%%%%%%%%%%%%%%%%%%%%%

The intuition behind our definition of interestingness is that if two patterns result in the same reduction in model bias, we are more interested in the pattern that requires fewer changes to the data (less support).
Thus, interestingness measures the average bias reduction per data point covered by a pattern.

In addition to finding patterns with high interestingness scores, we also aim to return a \revb{\textit{diverse} set of patterns that have little overlap in terms of the data points that they contain.} Diversity helps us avoid returning patterns that are too similar to each other, i.e., that differ only in minor details. This is a desirable property because patterns with large overlap convey similar information and, thus, lead to redundancy. \revb{We capture the notion of diversity through a containment score as defined next.}
% The interestingness measure prioritizes patterns that bring about minimal changes to the data and, therefore, have a higher interestingness score.

% We are interested in patterns that reduce bias; however, note that intervening on data instances may increase the empirical loss. To preserve the loss of the updated model, we only permit interventions that cause the resulting increase (if any) in loss to lie within a predefined threshold. Toward this goal, we define \textit{explanations} as patterns that have a positive interestingness score and preserve the loss of the updated model.

%%%%%%%%%%%%%%%%%%%%%%%%%%%%%%%%%%%%%%%%
\begin{definition}[\textbf{Containment score}]\label{def:containment}
  Given patterns $\pat$ and $\pat'$, the containment score $\cont$ measures the fraction of $\pat$ contained in $\pat'$:
%   their intersection:
  $$\cont(\pat, \pat') = \dfrac{|\apatI \cap \patI{\pat'}|}{|\apatI|.}$$
  Furthermore, for a pattern $\pat$ and a set of patterns $\patset$, we define
  $$\cont(\pat,\patset) = \max_{\pat' \in \patset} \cont(\pat,\pat').$$
\end{definition}
%%%%%%%%%%%%%%%%%%%%%%%%%%%%%%%%%%%%%%%%

The containment score quantifies the overlap between the data represented by a pair of patterns.
% Note that we are abusing notation in the definition. Note that we treat a patterns as sets of comparisons for the sake of this definition, e.g.,  $|\pat|$ to denotes the number of conjuncts in $\pat$. % and $\pat \cap \pat'$
% $\cont(\phi_a, \phi_b) \in [0,1]$.
Similar patterns have a higher fraction of overlapping data points (a containment score close to 1) and largely convey the same information differently. Dissimilar or diverse patterns, % on the other hand,
  have a containment score closer to 0.

%\boris{We can potentially safe some space and reduce notation by not defining explanation candidates. We just define topk based on $\patDom$ directly.}

%We are now ready to define an candidate explanation in terms of a pattern and its interestingness score as:
%%%%%%%%%%%%%%%%%%%%%%%%%%%%%%%%%%%%%%%%
%\begin{definition}[\textbf{(Candidate Explanations)}]
%  A candidate explanation $\expl{\pat}$ is defined in terms of pattern $\pat$ and its interestingness score $\score(\pat)$ in the form of a tuple $\expl{\pat} = \langle \pat, \score(\pat)\rangle$.
%\end{definition}
%%%%%%%%%%%%%%%%%%%%%%%%%%%%%%%%%%%%%%%%

%We use $\Expl{\dtrain} = \{\expl{\pat} = \langle \pat, \score(\pat) \rangle\}_{\pat \in \patDom}$ to denote the set of all candidate explanations for a training dataset $\dtrain$. We are interested in explanations that are \textit{interesting} and \textit{diverse} by way of covering distinct influential portions of the training data.
We are now ready to formalize the problem of generating the top-$k$ diverse and interesting data-based explanations: % for classifier:

%%%%%%%%%%%%%%%%%%%%%%%%%%%%%%%%%%%%%%%%
\begin{definition}[\textbf{Top-$k$ explanations}]\label{def:top-k-explanations}
  Generating the {\em top-$k$ data-based explanations} for the bias $\bias$ of a classifier requires  finding the  top-$k$ most responsible and diverse explanations. Formally, given a containment threshold $\cthresh$, the goal is to compute a set $\topk{k} \subseteq \patDom$ of explanation candidates as defined below.
  % such that {\em for all} $\expl{\pat} \in \topk{k}$ and $\expl{\pat'} \in \Expl{\dtrain} \setminus   \topk{k}$
  % either $\score(\pat')<\score(\phi)$ or   $\cont(\phi, \phi')> c$.
%\boris{make one line if need to save space:}
  \begin{align*}
    \topk{1}   & = \argmax_{\pat \in \patDom} \score(\pat) \\
    \topk{i+1} & = \argmax_{\pat \in \patDom \land \cont(\pat,\topk{i}) < \cthresh} \score(\pat), \ \text{for} \  \ 1<i \leq k.
  \end{align*}

  %  \[ \forall \expl_\phi \in \Expl_k, \expl_{\phi'} \in \Expl_{\mb D} \setminus   \Expl_{\mb D}:   \
  %  \]
   % such that for all $\expl_\phi \in \Expl_k$ there exist no $\expl_{\phi'} \in \Expl_{\mb D}$ such that  $\score(\phi')>\score(\phi)$ and $\cont(\phi_a, \phi_b)< c$, where $c$ is a given containment threshold.
   % a set of candidate explanations $\Expl_{\mb D}$ and a threshold $c$, the top-$k$ explanations $\Expl_k \subseteq \Expl_{\mb D}$ for bias of a classifier are sorted in descending order of $\score(\phi)$ and denoted by:
   % $$\Expl_{k} = \{\expl(\phi) \mid \expl(\phi) \in \Expl_{\mb D} \text{ and } \forall \text{ } \expl(\phi') \in \Expl_{\mb D}, \phi' \neq \phi, C(\phi, \phi') < c\}$$
   % such that for each pattern $\phi$,
    % $\responsibility_{\mathcal{L}} \geq \alpha_L$
    % and $\responsibility_{\mathcal{F}} \geq \alpha_F$ where
    % $\alpha_L$ and
    % $\alpha_F$ is the user-defined threshold for change in bias of the model.
\end{definition}
%%%%%%%%%%%%%%%%%%%%%%%%%%%%%%%%%%%%%%%%

%\boris{Technically, $\topk{k}$ should be parameterized based on a few things ($\dtrain$, $bias$). Not sure whether it is better to be explicit there ($\topk{k}(\dtrain,\bias,\cthresh)$) or be concise.}

Intuitively, $\topk{k}$ is computed by iteratively including into the current result set the next candidate with the highest score, skipping any candidate patterns whose overlap with one of the explanations in the result we have computed so far exceeds the threshold $\cthresh$. Note that $\topk{k}$ is not well-defined if multiple patterns have the same score. % To resolve this ambiguity of our definition,
We impose an arbitrary order over $\patDom$ to break ties.

Two major challenges complicate efforts to efficiently compute top-$k$ explanations. First,  computing the causal responsibility of a pattern % for classifier bias
is computationally intensive: we have to train a new classifier on the intervened training data obtained by removing the subset of data points covered by the pattern. Second, % to identify the top-$k$ explanations,
it is computationally prohibitive to exhaustively explore the space of all candidate explanations, which is exponential in the number of features.
\section{Computing Top-k Explanations}
\label{sec:solution}
This section describes our methods for addressing the challenges  of % efficiently
computing top-$k$ explanations for the bias of an ML model. % ~(cf.~\Cref{sec:problem}).
First, we describe how to efficiently approximate the causal responsibility of a   training data subset, % on bias of a classifier
 without having to retrain classifiers (\Cref{sec:intervention}). Then, we develop a lattice-based search algorithm which utilizes the approximation methods for estimating causal responsibility to efficiently generate top-$k$ explanations (\Cref{sec:topk}).

%Next, we describe our solution to address the above challenges.
%%%%%%%%%%%%%%%%%%%%%%%%%%%%%%%%%%%%%%%%%%%%%%%%%%%%%%%%%%%%%%%%%%%%%%%%%%%%%%%%
\subsection{Approximating Causal Responsibility}
\label{sec:intervention}
% Influence functions~\cite{cook-influence} are a classic technique from robust statistics that measure how the optimal model parameters depend on training data instances.
We now describe two methods for approximating the causal responsibility of a subset of the training data $\dtrain$ on classifier bias without retraining the classifier. The first method uses influence functions (\Cref{sec:influencef}). The second is a simple, yet effective, approximation based on gradient descent (\Cref{sec:gdapprox}).

%{\em influence functions} ~\cite{cook-influence}, which we review next.

%We use influence functions~\cite{cook-influence} to quantify the effect of an intervention that are widely used in model debugging~\cite{pmlr-v70-koh17a} to measure how the optimal model parameters change with a change in training data instances. We next  describe details of influence functions.

%%%%%%%%%%%%%%%%%%%%%%%%%%%%%%%%%%%%%%%%%%%%%%%%%%%%%%%%%%%%%%%%%%%%%%%%%%%%%%%%
\subsubsection{Influence function approximation}
\label{sec:influencef}

 Recall from \Cref{sec:prelim} that the optimal model $\oparam$ is the element of the set of possible models $\modelDom$ minimizing the empirical risk, i.e.,
 \begin{equation}
    \oparam = \underset{\theta \in \modelDom}{\argmin} \hspace{2mm} \lossD(\theta) =  \underset{\theta \in \modelDom}{\argmin} \hspace{2mm} \frac{1}{n} \sum_{i=1}^n \lossz(\mb z_i, \theta).
    \label{eq:params}
\end{equation}

Let $\grad \lossD(\theta)$ and $\hessian = \grad^2 \lossD(\theta) = \frac{1}{n} \sum_{i=1}^n \nabla_{\theta}^2 L(\mb z_i, \theta)$ be the gradient and the Hessian of the loss function, respectively. Under the assumption that the empirical risk $\lossD(\theta)$ is twice-differentiable and strictly convex, it is guaranteed that $\hessian$ exists and is positive-definite, which further implies $\hessian^{-1}$ exists.\footnote{Please see~\cite{pmlr-v70-koh17a} for a discussion on relaxing these assumptions on the loss function.}

%\subsubsection{Effect of Removing a data instance}
%\label{sec:if:first-order}
The influence of a data instance $\tpoint \in \dtrain$ is measured in terms of the effect of its removal from $\dtrain$ on model parameters $\oparam$. We first measure the effect of up-weighting $\tpoint$ by some small $\epsilon$; as we will see later, removing $\tpoint$ is equivalent to up-weighting it by $\epsilon=-\frac{1}{n}$, where $n$ is the number of data instances in $\dtrain$. Up-weighting $\tpoint$ by $\epsilon$ leads to a new set of optimal model parameters:
\begin{equation}
    \theta_{\epsilon}^* = \underset{\theta \in \modelDom}{\argmin} \hspace{2mm} L(\theta) + \epsilon L(\tpoint, \theta).
    \label{eq:updatedparams}
\end{equation}

We are interested in estimating how model parameters change due to an $\epsilon$ change in $\tpoint$. Let $\Delta_{\epsilon} = \theta_{\epsilon}^* - \oparam$ denote the {\em change in model parameters}. Note that because $\oparam$ is independent of $\epsilon$, we can capture the change in model parameters in terms of $\theta_\epsilon$:
\begin{equation}
    \frac{d\theta_{\epsilon}^*}{d\epsilon} = \frac{d\Delta_{\epsilon}}{d\epsilon}.
    \label{eq:paramchange}
\end{equation}

%%%%%%%%%%%%%%%%%%%%%%%%%%%%%%%%%%%%%%%%%%%%%%%%%%%%%%%%%%%%%%%%%%%%%%%%%%%%%%%%
\partitle{Influence of a single data point}
Since $\theta_{\epsilon}^*$ minimizes the updated training loss (Equation~\eqref{eq:updatedparams}), the gradient of the loss function at $\theta_{\epsilon}^*$ should be zero, i.e.,
\begin{equation}
    \grad \lossD(\theta_{\epsilon}^*) + \epsilon \grad \lossz(\tpoint, \theta_{\epsilon}^*) = 0.
    \label{eq:diffloss}
\end{equation}

%if

Let us denote the gradient of the loss function as $g(\epsilon, \theta)$; hence, the LHS of Equation~\eqref{eq:diffloss} reads $g(\epsilon, \theta_{\epsilon}^*)$.  The key idea behind influence functions is that up-weighting a data point $\mb z$ by a very small $\epsilon$ does not significantly change the optimal parameters, i.e., if $\epsilon \rightarrow 0$, then $\theta_{\epsilon}^* \rightarrow \oparam$. Therefore, using the first principles, $g(\epsilon, \theta_{\epsilon}^*)$ can be effectively approximated by Taylor's expansion of $g(\epsilon, \theta)$ at $\theta^*$, the optimal parameters of the original model. ~\footnote{Recall that Taylor’s expansion of function $u$ about the point $t$ with
increment $h$ is given by $u(t+h)=u(t)+ u'(t)h + u''(t) h^2 \ldots . $ Hence, in our case $g(\epsilon, \theta^*_{\epsilon})=g(\epsilon, \theta^*+\Delta_{\epsilon} ) \approx [\grad \lossD (\oparam) + \epsilon \grad \lossz(\tpoint, \oparam)] + [\grad^2 \lossD(\oparam) + \epsilon \grad^2 \lossz(\tpoint, \oparam)] \Delta_{\epsilon} $.} By plugging this approximation into the LHS of Equation~\eqref{eq:diffloss},  we obtain:
\begin{equation}
    [\grad \lossD (\oparam) + \epsilon \grad \lossz(\tpoint, \oparam)] + [\grad^2 \lossD(\oparam) + \epsilon \grad^2 \lossz(\tpoint, \oparam)] \Delta_{\epsilon} \approx 0.\\
    % \nabla_{\theta}^2 L(\theta_{\epsilon}^*)\frac{d\theta_{\epsilon}^*}{d\epsilon} + \epsilon \cdot \nabla_{\theta}^2 L(\tpoint, \theta_{\epsilon}^*) \frac{d\theta_{\epsilon}^*}{d\epsilon} + \nabla_{\theta} L(\tpoint, \theta_{\epsilon}^*) = 0
    \label{eq:if_1}
\end{equation}
Solving for $\Delta_{\epsilon}$, we get
\begin{equation}
    \Delta_{\epsilon} = -[\grad \lossD(\oparam) + \epsilon \grad \lossz(\tpoint, \oparam)][\grad^2 \lossD(\oparam) + \epsilon \grad^2 \lossz(\tpoint, \oparam)]^{-1}.
    \label{eq:deltaEpsilon}
\end{equation}
From Equation~\eqref{eq:params}, since $\oparam$ minimizes $\lossD(\theta)$, we obtain $\grad \lossD(\oparam) = 0$. Ignoring higher orders of $\epsilon$,
\begin{equation}
    \Delta_{\epsilon} = - \grad \lossz(\tpoint, \oparam) \grad^2 \lossD(\oparam)^{-1}\epsilon.
    \label{eq:deltaEpsilon_2}
\end{equation}
From Equations~\eqref{eq:paramchange} and~\eqref{eq:deltaEpsilon_2}, as $\epsilon \rightarrow 0$, the influence of up-weighting a data instance $\tpoint \in \dtrain$ by $\epsilon$ on the model parameters is computed as:
\resizebox{1\linewidth}{!}{
  \begin{minipage}{1.05\linewidth}
    \begin{align}
      \influence_{\theta}(\tpoint) &= \frac{d\theta_{\epsilon}^*}{d\epsilon}\biggr\rvert_{\epsilon=0}
      % \frac{\Delta_{\epsilon}}{d\epsilon}\biggr\rvert_{\epsilon=0}
      % \nonumber\\
      =-\grad^2 \lossD(\oparam)^{-1}\grad \lossz(\tpoint, \oparam)
      % \nonumber\\
      = -\hessian^{-1}\grad \lossz(\tpoint, \oparam)
      \label{eq:if}
    \end{align}
  \end{minipage}
}\\

To remove data instance $\tpoint$, we consider up-weighting it by $\epsilon = -\frac{1}{n}$. The change in model parameters due to removing $\tpoint$ can, therefore, be linearly approximated by computing $d\theta_{\epsilon}^* \approx -\frac{1}{n} \influence_{\theta}(\tpoint)$.  %Intuitively, first-order influence function approximate the parameters of the new model }

\partitle{Influence of subsets}
Using first-order (FO) influence function approximation (Equation~\eqref{eq:if}), the effect of removing a subset of data instances $\mb S \subseteq \dtrain$ on model parameters can be obtained by:
\begin{equation}
    \influence^{(1)}_\theta(\mb S) = \sum \limits_{\tpoint \in \mb S} \influence_\theta(\tpoint).
    \label{eq:if_subset}
\end{equation}

The approximation shown in \Cref{eq:if_subset} is quite accurate when the parameters of the updated model are close to those of the original model: because we estimate the effect of removing a set of data points by summing their individual effects, which means that we assume that the effect of removing one data point on the model is independent of the effect of removing any other data point. Put differently, when approximating the influence of a data point, we assume that the removal of other data points will not affect the model. While in general this assumption does not hold, it leads to an acceptable approximation if a small fraction of training data points is removed, but accuracy will decline when a larger fraction of training data is removed.
% When the parameters of the updated model are significantly different from the original model parameters, such as when a significant fraction of the training  data instances are removed, first-order influence approximations are no longer accurate.
For such large model perturbations, using higher-order terms can reduce approximation errors significantly.
This issue can be alleviated by computing the effect of uniformly up-weighting data points in a subset of training data by some small $\epsilon$, using the same idea as influence functions but considering higher-order optimality criteria~\cite{icml2020_4261}. Specifically, in the derivation of influences, we do not ignore second-order terms of $\epsilon$
% in Equations~\eqref{eq:if_1} and~\eqref{eq:deltaEpsilon_2}
for an even more accurate estimation of the \textit{group influence} of up-weighting a subset of training data points on model parameters, and we obtain the influence as:
\begin{equation} \small
    \influence^{(2)}_\theta(\mb S) = \left(\dfrac{1}{|\mb \dtrain|-|\mb S|}\right) \influence^{(1)}_\theta(\mb S) + \left(\dfrac{|\mb S|}{|\mb \dtrain|-|\mb S|}\right) \influence'_\theta(\mb S),
    \label{eq:inf:second-order}
\end{equation}
where
{\small
$$\influence'_\theta(\mb S) =
% \dfrac{|\mb S|}{|\mb \dtrain|-|\mb S|}
\left(\mb I - \hessian^{-1}\dfrac{1}{|\mb S|} \sum \limits_{\tpoint \in \mb S} \grad^2 \lossz(\tpoint, \theta) \right) \influence^{(1)}(\mb S)
$$
}

We refer the readers to \cite{icml2020_4261} for detailed derivations of the above expression for the group influence of a subset, but note that the second term ($\influence'$) in the second-order (SO) approximation captures correlations among data points in $\mb S$ through a function of gradients and Hessians of the loss function at the optimal model parameters. When training data points are correlated, the SO group influence function is more informative and captures the ground truth influence more accurately. We empirically compare the accuracy of first- and second-order influence function approximations in \Cref{sec:exp}.

%%%%%%%%%%%%%%%%%%%%%%%%%%%%%%%%%%%%%%%%%%%%%%%%%%%%%%%%%%%%%%%%%%%%%%%%%%%%%%%%
\partitle{Causal responsibility}
Influence functions can be adopted to approximate causal responsibility using the chain rule of differentiation, as follows. We can estimate the effect of up-weighting data points in $\mb S$ on a function $f(\theta)$ as:
\begin{align}
    \influence_{f}(\mb S) &= \frac{df(\theta_{\epsilon}^*)}{d\epsilon}\biggr\rvert_{\epsilon=0}= \frac{df(\theta_{\epsilon}^*)}{d\theta}\frac{d(\theta_{\epsilon}^*)}{d\epsilon}\biggr\rvert_{\epsilon=0}
    % \nonumber\\
    % &= \nabla_{\theta}f(\oparam)^\top \frac{d(\theta_{\epsilon}^*)}{d\epsilon}\biggr\rvert_{\epsilon=0} \nonumber\\
     = \nabla_{\theta}f(\oparam)^\top  \influence_{\theta}(\mb S),
    % = -\nabla_{\theta}f(\oparam)^\top H_{\theta}^{-1}\nabla_{\theta} L(\tpoint, \oparam)
    \label{eq:if:function}
\end{align}
where $\influence_{\theta}(\mb S)=\influence_{\theta}^{(1)}(\mb S)$ or $\influence_{\theta}^{(2)}(\mb S)$ depending on whether the influence is computed using first- or second-order group influence function approximations, respectively.

Given a fairness metric $\bias$,
$\influence_\bias(\mb S)$ captures the effect of intervening on $\mb S$ on the bias of the classifier and is used as the numerator in \Cref{def:responsibility} to compute the responsibility of  $\mb S$. % toward bias.

%%%%%%%%%%%%%%%%%%%%%%%%%%%%%%%%%%%%%%%%%%%%%%%%%%%%%%%%%%%%%%%%%%%%%%%%%%%%%%%%
\subsubsection{Gradient-based approximation}
\label{sec:gdapprox}
\revd{As shown in~\cite{pmlr-v70-koh17a}, FO influence function approximations can be used to update individual training data points to maximize test loss. However, they deviate from ground truth influence for a \textit{group} of data points because they do not capture the data correlations in a group~\cite{basu2020second}. SO influence functions, on the other hand, capture correlations but have only been explored for the case of subset removal~\cite{basu2020second}.

To generate update-based explanations, we introduce an alternative approach for approximating causal responsibility. We assume that the updated model parameters are obtained through one step of gradient descent (which we use in \Cref{sec:repair}) and empirically find explanations that are more accurate than even SO influence approximations (details in \Cref{exp:qual:responsibility})}.

Note that as in \Cref{eq:params}, the updated model parameters when a subset of data points is removed is given by:
\begin{equation}
    \upar =  \underset{\params \in \modelDom}{\argmin} \hspace{2mm} \left(\lossD(\mb D, \params) - \frac{1}{|\mb S|} \sum_{\tpoint \in \mb S} \lossz(\tpoint, \params)\right).
    \label{eq:params:updated}
  \end{equation}
The preceding equation can be solved using gradient descent by taking repeated steps in the direction of the steepest descent of the loss function (for data points in $\mb D \setminus \mb S$). However, inspired by the existing literature in adversarial ML attacks~\cite{DBLP:journals/corr/abs-2006-14026}, which assumes that model parameters do not change significantly when a small subset of data points is poisoned, we assume minimal change in model parameters. % due to the same circumstance.
% This assumption is line with existing literature in adversarial ML attacks~\cite{DBLP:journals/corr/abs-2006-14026} that assumes a small change in model parameters due to poisoning of a very small subset of data points.
Thus, the change in model parameters is approximated in terms of a {\em single step of gradient descent}, as follows:
%%%%%%%%%%%%%%%%%%%%
\begin{align}
    \upar = \params - \eta\left( \grad \lossD(\mb D, \oparam) - \dfrac{1}{n} \sum_{i=1}^{|\mb S|} \grad \lossz\left(\tpoint_i, \params\right)\right),
    \label{eq:one-step-GD}
\end{align}
%%%%%%%%%%%%%%%%%%%%
where $\eta$ is the learning rate for the gradient step.

The effect of removing $\mb S$ on bias is then computed as the difference in test bias before and after $\mb S$ is removed: $\influence(\mb S) =\bias(\upar, \dtest) - \bias(\params, \dtest)$. While our single-step gradient descent approach can be used to estimate subset influence, this may not be a good idea % for generating top-$k$ explanations
for learning algorithms that use more efficient techniques than gradient descent. We, therefore, use this approach mainly for generating update-based explanations in Section~\ref{sec:repair}.

%We empirically show the effectiveness of this approach in~\Cref{sec:exp}.

%%% Local Variables:
%%% mode: latex
%%% TeX-master: "main"
%%% End:

% \vspace{-5mm}
\subsection{Lattice-Based Search}
\label{sec:topk}

%This section considers the problem of generating interesting pattern-based explanations, and outlines our algorithm to solve the problem of finding the top-k explanations as defined in Definition~\ref{def:top-k-explanations}.
Having discussed several techniques for efficiently approximating influence, we now introduce our algorithm for finding the top-k explanations according to \Cref{def:top-k-explanations}.
As discussed in \Cref{sec:intro}, the na\"ive approach for computing the top-k explanations by evaluating all possible patterns is exponential in the number of attributes, and, thus, is % computationally
infeasible even when we estimate the influence of patterns.

\begin{figure}[t]
\centering
		\includegraphics[scale=0.7]{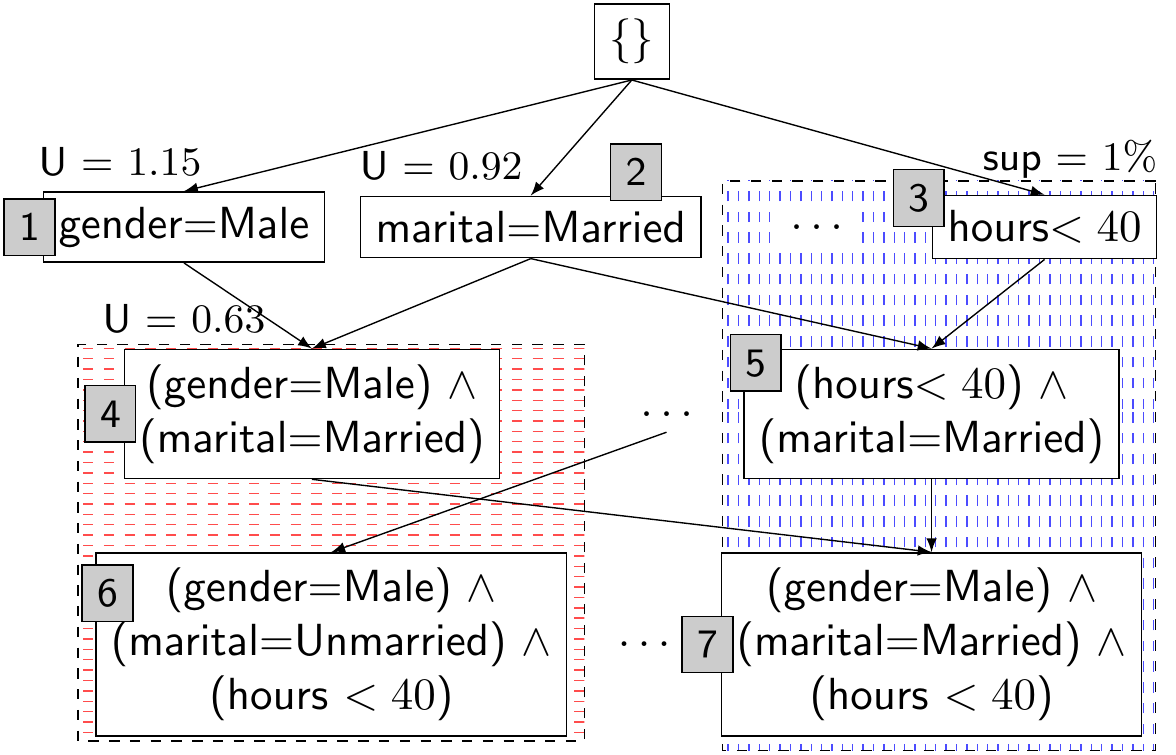}
	\vspace*{-.2cm} \caption{\revb{An overview of the lattice structure and search with support threshold $\tau=5\%$.}}
	\vspace{-5mm}
	\label{fig:lattice}
\end{figure}

% Toward this objective, we must first clarify what makes a pattern \textit{influential}. A pattern $\pat$ is considered influential if it: (1)~consists of at least a minimum pre-defined fraction of data instances, i.e., $\support(\pat) \geq \supthresh$, (2)~is a root cause of bias, and (3)~is maximally interesting, i.e., there exists no finer pattern $\phi'$ with additional predicates $\phi_k$ such that $\phi' = \phi \wedge \phi_k$ and $\score(\phi') \geq \score(\phi)$.
%\sandy{Reminder to complete new figure and provide a caption.}
%%%%%%%%%%%%%%%%%%%%%%%%%%%%%%%%%%%%%%%%
%\begin{figure}[t]%
%	\centering
%	%\includegraphics[width=0.6\columnwidth]{plots/lattice-structure.pdf}
%	\caption{Visualization of the lattice structure for the example in ...\romila{Do we need this?} \babak{not a priority anymore}}
%\end{figure}
%%%%%%%%%%%%%%%%%%%%%%%%%%%%%%%%%%%%%%%%

Toward this objective, we propose \texttt{ComputeCandidates}, an algorithm that takes the training data as input and generates a list of candidate patterns. \revb{Figure~\ref{fig:lattice} shows an example of part of the lattice search space for patterns.} \texttt{ComputeCandidates} generates explanations starting with patterns each having a single predicate. The algorithm then iteratively generates patterns with $i$ predicates by merging two patterns with $i-1$ predicates that differ in only one predicate.
% \tikz{\draw[text=red,thick, draw=red]  rectangle{1};}
 \revb{
%  For instance, it merges patterns $\pat_1 = \{ (\attr{hours} < 40) \land (\attr{marital} = Married) \}$ and $\pat_2 = \{ (\attr{marital} = Married) \land (\attr{gender} = Male) \}$ into a pattern $\pat = \{ (\attr{hours} < 40) \land (\attr{gender} = Male) \land (\attr{marital} = Married) \}$.
 For instance, it merges patterns \latticelabels{4} and \latticelabels{5} in Figure~\ref{fig:lattice} into pattern \latticelabels{7}.}
 The idea is similar to frequent itemset mining~\cite{10.5555/645920.672836,10.1145/335191.335372} in data mining where itemsets with $n$ items are generated by successively merging candidate itemsets of smaller size.

By itself, building patterns bottom-up does not reduce the size of the search space.
We propose two heuristics to address this issue. The first heuristic is that we assume as input a  support threshold $\supthresh$ and only consider patterns whose support is above $\supthresh$. The rationale for this heuristic is that patterns with low support describe only a small portion of the training data and, thus, are unlikely to identify systematic issues. \revb{In our experiments in~\cite{techreport}, with $\supthresh$ as small as $1\%$, we did not observe  patterns with low support reducing bias by much,
% as shown in Table 3 of~\cite{techreport},
% Even for $\supthresh=1\%$,
and found
much larger patterns (with $\sim 27\%$ support) dominating the top-k.} For a candidate pattern $\pat$ generated by merging patterns $\pat_i$ and $\pat_j$ we have $\sup(\pat) \leq \sup(\pat_i)$ and $\sup(\pat) \leq \sup(\pat_j)$. Thus, pruning $\pat_i$ prunes the entire sub-lattice whose root is $\pat_i$. \revb{In \Cref{fig:lattice}, pattern \latticelabels{3} is pruned since its support ($1\%$) is less than the threshold ($5\%$). As a result, patterns  formed by merging pattern \latticelabels{3} with another pattern, including patterns \latticelabels{5} and \latticelabels{7} and the entire sub-tree shaded blue \tikz[baseline=(X.base)]\node [pattern=vertical lines, pattern color = blue!70, rectangle, text width=2mm, draw](){};, are also pruned.
% pruning pattern $\pat_1 = \{ (\attr{hours} < 40) \land (\attr{marital} = Married) \}$ also prunes other patterns including  $\pat = \{ (\attr{hours} < 40) \land  (\attr{marital} = Married) \land (\attr{gender} = Male) \}$ formed by merging $\pat_1$ with another pattern.
}

% To see why this has to be the case, note that $\pat$ contains the same predicates as $\pat_i$ ($\pat_j$) plus one additional predicate. Thus, any training data point in $\patI{\pat}$, i.e., that fulfills the predicates of $\pat$, is also in $\patI{\pat_i}$ ($\patI{\pat_j}$).

Our second heuristic is to prune patterns during merging.
Consider a pattern $\pat$ generated by merging patterns $\pat_i$ and $\pat_j$. We only consider $\pat$ if its responsibility exceeds the responsibility of the two patterns it is generated from: $\responsibility(\pat) > \responsibility(\pat_i)$ and $\responsibility(\pat) > \responsibility(\pat_j)$. Note that if this is the case, then  $\score(\pat) > \max(\score(\pat_i), \score(\pat_j))$, because $\score(\pat) = \frac{\responsibility_\bias(\patI{\pat})}{\support(\pat)}
$ and as mentioned above $\support(\pat) \leq \min(\support(\pat_i), \support(\pat_j))$. \revb{
In \Cref{fig:lattice},
% the pattern $\pat = \{ (\attr{gender} = Male) \land (\attr{marital} = Married) \}$ is generated from patterns $\pat_1 = \{ \attr{gender} = Male \}$ and $\pat_2 = \{ \attr{marital} = Married \}$ if $\score(\pat) > \score(\pat_1)$ and $\score(\pat) > \score(\pat_2)$.
we do not generate \latticelabels{4} (and subsequently its successors, shaded red \tikz[baseline=(X.base)]\node [pattern=horizontal lines, pattern color = red!70, rectangle, text width=2mm, draw](){}; ) because $U($pattern \latticelabels{4}$)$ is less than both $U($pattern \latticelabels{1}$)$ and $U($pattern \latticelabels{2}$)$.
} The rationale for this heuristic is that patterns with more predicates are harder to interpret. Thus, an increase in the number of predicates should be justified by an increased impact on the bias of the model.
% \boris{I don't think that my explanation above is very convincing yet.}

We now discuss our algorithm in more detail (pseudocode is shown in \Cref{alg:eg}).
We start by generating all possible single predicate patterns (\Cref{alg:ge-one-pred}) by iterating through all features $X \in \features$ and each possible value $val$ for the feature and generate three types of comparisons: $X < val$, $X = val$, and $X > val$.
Note that for features with a large number of possible values, we can apply binning techniques to reduce the number of candidate patterns. Additionally, binning has the advantage of preventing the generation of almost identical explanations (e.g., $\attr{hours} < 40$ and $\attr{hours} < 42$).

We then iteratively create patterns (\Cref{alg:ge-iterate}) of size $i$ by merging two patterns of size $i-1$ that  differ in only one predicate to generate a candidate pattern of size $i$. The algorithm terminates if no new candidates have been produced in the current iteration. For each generated candidate pattern $\pat$ we test if its support is larger than the threshold $\supthresh$ and its responsibility is larger than that of the patterns it was derived from.
If both the conditions are fulfilled then we
include $\pat$ in $\patset_i$, the set of candidate patterns of size $i$. Finally, we return all candidate explanations, which is the union of all sets $\patset_i$ generated so far.

Note that while merging patterns, we do not have to  consider patterns that are {\em conflicting}. Patterns $\pat_i$ and $\pat_j$ are conflicting if they both contain a predicate on an attribute $X$ and the conjunction of these predicates is unsatisfiable.
In \Cref{fig:lattice}, patterns
% $\pat_1 =  \{\attr{marital} = Married\}$ and $\pat_2 =  \{\attr{marital} = Unmarried\}$
\latticelabels{6} and \latticelabels{7}
are conflicting. The merged pattern $\pat = $\latticelabels{6} $\cup$ \latticelabels{7} has zero support, and we do not need to consider it further.

{\small

%%%%%%%%%%%%%%%%%%%%%%%%%%%%%%%%%%%%%%%%%%%%%%%%%%%%%%%%%%%%%%%%%%%%%%%%%%%%%%%%
\begin{algorithm}[t]
  \KwInput{Training data $\dtrain$, support threshold $\supthresh$}
  \KwOutput{Candidate explanations}

  $\patset_1 \gets \emptyset$\Comment{Initialize set of one-predicate explanations} \label{alg:ge-one-pred}

  \For{$X \in \features$}{
    \For{$val \in \pi_{X}(\dtrain)$}{
      \For{$\pat \in \{\{X=val\},\{X<val\}, \{X>val\}\}$}{
        \If {$\support(\pat) > \supthresh$}{
          $\patset_1 \gets \patset_1 \cup \{\pat\}$
        }
        % \tcc*{\textsf{identify root nodes}}
      }
    }
  }
  % $\patset_2 \gets \emptyset$

  % \For{$\pat_i, \pat_j \in \patset_1$}{
  %   \If{$\pat_i$.col <> $\pat_j$.col} {
  %     \If{$\support(\pat_i \cup \pat_j) > \supthresh$}{
  %       $\patset_2 \gets \patset_2 \cup \{\pat_i \cup \pat_j\}$
  %     }
  %   }
  % }

  $i \gets 2$ \label{alg:ge-iterate}

  \While{$\patset_{i-1} \neq \emptyset$}{
    $\patset_{i} \gets \emptyset$

    \For{$\pat_i,\pat_j \in \patset_{i-1}$}{
      \If{$|\pat_i \cap \pat_j| = i-2 \land \support(\pat_i \cup \pat_j) \geq \supthresh$}{
        $\pat = \pat_i \cup \pat_j$

        \If{$\influence(\pat) > \influence(\pat_i) \land \influence(\pat) > \influence(\pat_j)$}{

          $\patset_{i} \gets \patset_{i} \cup \{\pat\}$
        }
      }
    }
    $i \gets i + 1$
  }

  \Return $\bigcup_{j=0}^{i} \patset_i$

  \caption{ComputeCandidates}
  \label{alg:eg}
\end{algorithm}
%%%%%%%%%%%%%%%%%%%%%%%%%%%%%%%%%%%%%%%%%%%%%%%%%%%%%%%%%%%%%%%%%%%%%%%%%%%%%%%%
}

%%%%%%%%%%%%%%%%%%%%%%%%%%%%%%%%%%%%%%%%%%%%%%%%%%%%%%%%%%%%%%%%%%%%%%%%%%%%%%%%
\partitle{Diversity of explanations}
We use \Cref{alg:topk} to compute a diverse set of top-k explanations as defined in \Cref{def:top-k-explanations}. We first use \texttt{ComputeCandidates} to generate a set of candidate explanations, which is then sorted based on the interestingness score $\score$. We iterate over the set in sort order, include patterns into the result, and skip patterns whose overlap with any of the previously added patterns exceeds the threshold $\cthresh$.

\reva{We are interested in patterns that, when removed or modified, reduce bias maximally without significantly affecting model accuracy. Instead of directly optimizing for minimal accuracy loss, we penalize patterns with high support (or, low interestingness score). As an extreme example, we can remove the entire data (that has the maximum support) and obtain a perfectly unbiased classifier that makes random guesses. However, such a pattern has no explanatory value. This intuition aligns with the notion of minimality in database repair
% , where the goal is 
that seeks
to find minimal subsets of data responsible for an inconsistency.
% Explanations may be utilized  to identify training data points that should be manually investigated which would not be possible for patterns with high support.
Optimizing for bias reduction and accuracy loss simultaneously is an interesting direction for future work.

Note that \sys's explanations describe training data subsets that may have potential data quality errors or point to historical biases reflected in training data or bias that was introduced during data collection. These errors are not detectable using standard data cleaning algorithms such as outlier detection. \sys can be complemented with existing error detection mechanisms or external sources of information on data provenance to expose the errors in the identified subsets.
}

% Note that simply sorting the top-k explanations based on their interestingness score may result in redundant explanations that have a number of data points in common. To ensure our explanations are interesting and cover as many diverse patterns as possible,

% \romila{we need an algorithm to generate the top-k using containment, as defined in 3.6.}. \jiongli{Since we are generating the most interesting explanations, the interestingness score has the top priority, and the diversity of patterns comes next. Based on this idea, we first sort the candidate patterns by their interestingness scores. To decide/evaluate whether patterns are dissimilar enough, we set a threshold $\cthresh$ for containment score, and two patterns are defined to be similar if their containment score exceeds $\cthresh$. Then, we keep iterating over the sorted set of candidate patterns and adding patterns into the top-k set if they are  dissimilar to all patterns already there. We break the iteration when the top-k set is full (size equals k) because any following patterns will not have interestingness scores higher than any of patterns in the top-k set. According to experimental results, this procedure cost much less time compared to lattice search, but it successfully filtered out redundant patterns. (\Cref{alg:topk})}

%%%%%%%%%%%%%%%%%%%%%%%%%%%%%%%%%%%%%%%%%%%%%%%%%%%%%%%%%%%%%%%%%%%%%%%%%%%%%%%%
{\small
\begin{algorithm}[t]
	\KwInput{Training dataset $\dtrain$, support threshold $\supthresh$, containment threshold $\cthresh$, desired number of top explanations $k$}
    \KwOutput{Top-k explanations}

    $\patset_{cand} \gets \Call{ComputeCandidates}{\dtrain, \supthresh}$

    $\patset_{top-k} \gets \emptyset$ 	\Comment{Set of top-k explanations}

    \For{$\pat \in$ \Call{sort-by-score}{$\patset_{cand}$}}{
        \If {$\neg\,\exists \pat_j \in \patset_{top-k}: \cont(\pat,\pat_j) > \cthresh$}{
		    $\patset_{top-k} \gets \patset_{top-k} \cup \pat_j$
		}
	    \If {$|\patset_{top-k}| = k$}{
		    \Return $\patset_{top-k}$
		}
	}

    \caption{Generate Top-k Explanations}
	\label{alg:topk}
\end{algorithm}
}
%%%%%%%%%%%%%%%%%%%%%%%%%%%%%%%%%%%%%%%%%%%%%%%%%%%%%%%%%%%%%%%%%%%%%%%%%%%%%%%%

%%% Local Variables:
%%% mode: latex
%%% TeX-master: "main"
%%% End:

% \section{Generating Update-Based Explanations}
\section{Update-Based Explanations}
\label{sec:repair}

%In Section~\ref{sec:solution}, we developed techniques for generating explanations based on \textit{removal} of data points from training data.
In this section, we formalize the problem of generating explanations based on \textit{updating} training data. Given an influential subset of training data obtained through the techniques proposed in \Cref{sec:solution}, our goal is to find a homogenous update that can lead to maximum bias reduction. Toward that goal, we first discuss an approach for approximating the influence of updating or perturbing a subset of training data on model bias.

%More specifically, we obtained explanations that identify subsets of training data such that removing them from traning data results in maximum reduction of bias. This section formalized and solve the problem of generating update-based explanations. Intuitively, the goal is to find minimal perturbations in subsets of data points that are highly responsible for bias that lead to maximum bias reduction.

%such that removing subsets of data instances satisfying those patterns resulted in maximum reduction of bias. In this section, we solve the problem of updating those patterns instead of removing them such that the update will guarantee a reduction in bias.

%\babak{removed from section 4. It should be dicussed her}

%\subsection{Gradient descent approach to updates}
%\label{sec:updexp}

Consider a classifier with optimal parameters $\oparam$ trained on a training dataset $\dtrain $. Let $\mb S= \{\mb z_i = (\mb X_i, Y_i)\}_{i=1}^m$ be a subset of $\dtrain$ consisting of $m$ data points. Also, let $\mb S^p= \{\mb z_i^p = (\mb X_i^p, Y_i^p)\}_{i=1}^m$ be the result of uniformly updating each instance in $\mb S$ using a {\em perturbation vector} $\boldsymbol \delta = \{\delta_1, \ldots, \delta_{|\mb X|}\}$. The attribute values of each data point $\mb z \in \mb S$ are updated by applying the perturbation vector: $\mb z^p =  \mb z + \boldsymbol \delta $, i.e., the $j$-th attribute of $\mb z$ is updated by $\delta_j$. For example, if the value of the working hours attribute for an individual is $32$, then an update of $8$ will change it to $40$. In this example, uniform perturbation means that the working hours for all individuals from the subset $\mb S$ are increased by $8$.  Our goal is to compute the optimal parameters $\theta_{p}$ for a new model trained on $\dtrain^p=(\dtrain \setminus \mb S) \ \cup \ \mb S^p$. We approximate $\theta_{p}$ using a single step of gradient descent (Section~\ref{sec:gdapprox}), as follows:
%We assume that the perturbed model is computed as a single gradient step from the original model, and estimate $\theta_{p}$ as:
\begin{align}
    \theta_{p} = \oparam - \dfrac{\eta}{n}\left( \sum_{i=1}^{n-m} \grad \lossz(\mb z_i, \oparam) + \sum_{i=1}^{m} \grad \lossz(\mb z_i^p, \oparam)\right)
    \label{eq:delta_theta}
\end{align}
where $\eta$ is the learning rate for the gradient step.

Therefore, the influence of an update to $\mb S$ on the bias of the model can be measured using the chain rule, as follows:
\begin{align}
    \Delta \bias(\oparam, \theta_{p} , \dtest) = \grad \bias(\oparam, \dtest)^\top \left(\theta_{p} - \oparam\right)
    \label{eq:delta_f}
\end{align}

Now, given a subset of training data, $\mb S \subseteq \dtrain$, our goal is
to obtain an update on $\mb S$ that leads to the maximum bias reduction.  That is, we want to solve the following optimization problem:
\begin{align}
    \boldsymbol \delta^* = \arg\!\max_{\boldsymbol \delta}  \Delta \bias(\oparam, \theta_{p} , \dtest)
    % \text{subject to: }
    \label{eq:opt}
\end{align}
%where $\grad \lossD(\mb Z^p, \oparam) = \sum_{i=1}^{m} \grad \lossz(\mb z^p, \oparam)$. %\babak{can we replace with with 14?}

% \begin{figure}[t]
%   \centering
% 	%\includegraphics[width=0.5\columnwidth]{plots/repair-example.pdf}
% 	\caption{Suggested repair for the example in ...}
% \end{figure}

%For the homogeneous repair problem where we assume that we update data instances in the perturbed subset homogeneously, our goal is to devise homogeneous interpretable perturbations that maximize bias reduction on the test dataset.
Using Equations~\eqref{eq:delta_theta} and~\eqref{eq:delta_f}, Equation~\eqref{eq:opt} becomes:
\begin{align}
    \boldsymbol \delta^* = \arg\!\min_{\boldsymbol \delta} \grad \bias(\oparam, \dtest)^\top \grad \lossD(\mb S^p, \oparam)
    \label{eq:opt_delta}
\end{align}

% We solve this optimization problem for perturbations
% $\boldsymbol \delta$ % = \{\delta_1, \ldots, \delta_m\}$
% that homogeneously update the entire data points in $\mb S$. For homogeneous perturbations, the updated instance after perturbing data instance $\mb z$ is represented as
% $\mb z' = \mb z + \boldsymbol \delta$ where $x_i' = x_i + \delta_i$. We should mention here that we do not perturb the class attribute. Therefore, $\mb Z^p = \mb Z + \boldsymbol \delta$ actually represents $\mb X^p = \mb X + \mb \delta$ and Equation~\eqref{eq:opt} becomes:
% \begin{align}
%     \boldsymbol \delta = \arg\!\max_{\boldsymbol \delta} \grad \bias(\oparam)^\top \grad \lossD(\mb Z^p, \oparam)
%     \label{eq:opt_delta}
% \end{align}

\noindent We solve Equation~\eqref{eq:opt_delta} using the gradient descent algorithm to obtain the perturbation vector:
\begin{align}
    \boldsymbol \delta_{k+1} = \boldsymbol \delta_k - \eta\grad \bias(\dtest, \oparam)^\top \nabla_{\delta}\left(\grad \lossD(\mb S^p + \boldsymbol \delta_k, \oparam)\right)
    \label{eq:opt:perturb}
\end{align}
where $\eta$ is the learning rate of the gradient ascent step.

An updated data point is obtained as $\mb z^p = \mb z + \boldsymbol\delta_{k+1}$. Note that the preceding formulation can result in perturbations that lie outside the input domain. We add domain constraints to prevent this from happening.  % that this is not the case, . to ensure the updated attribute values lie inside the input domain.
In particular, the updates should change an attribute of a data point from one value to another in the input domain, i.e., $\mb z, \mb z^p \in \Dom{\features} \times \Dom{Y}$. We solve this \textit{constrained} optimization problem in \Cref{eq:opt:perturb} using \textit{projected} gradient descent~\cite{intro-optimization}, which works as follows: if the updated data point violates the domain constraint, i.e., $\mb z^p \notin \Dom{\features} \times \Dom{Y}$, then project it back to the input domain $\Dom{\features} \times \Dom{Y}$ as:
\begin{equation}
    \mb z_{up} = \argmin \limits_{\mb z' \in \Dom{\features} \times \Dom{Y}}
    \left\lVert \mb z' - \mb z^p \right\rVert
    \label{eq:zup}
\end{equation}
The projection ensures that $\mb z_{up}$ is the data point in the input domain $\Dom{\features} \times \Dom{Y}$ that is closest to the actual perturbation $\mb z^p$.

\newcommand{\delfigspace}{\\[-6mm]}
\vspace{-0.2cm}
\section{Experiments}
\label{sec:exp}
In our experimental evaluation of \sys, we aim
% This section presents experiments that evaluate the feasibility and efficacy of \sys. Weaim
to address the
following questions: \textbf{Q1:} How effective are the proposed techniques for approximating causal responsibility of training data subsets?
\textbf{Q2:} What is the end-to-end performance of \sys\ in generating interpretable and diverse data-based explanations? \textbf{Q3:}
What is the quality of the update-based explanations? \textbf{Q4:} How effective is our approach at detecting data errors responsible for ML model bias?
%\boris{for testing:}
\iftechreport{
Additional experiments can be found in \cite{techreport}.
}
% \babak{do we have any baseline to compaye againt? To decouple w into the quetsions Q1) how effective our approach is in generaing the top-K explanations? Q2 how is the quality of the update-based expalnations?
% Q3: how effective is our apporach at detecting adversarial attacks on fairness? We should have one section as per each question in the subsequent. }
\begin{figure}[t]
  \centering
  %%%%%%%%%%%%%%%%%%%%%%%%%%%%%%%%%%%%%%%%
  \begin{subfigure}{1.0\linewidth}
    \includegraphics[width=1.0\linewidth]{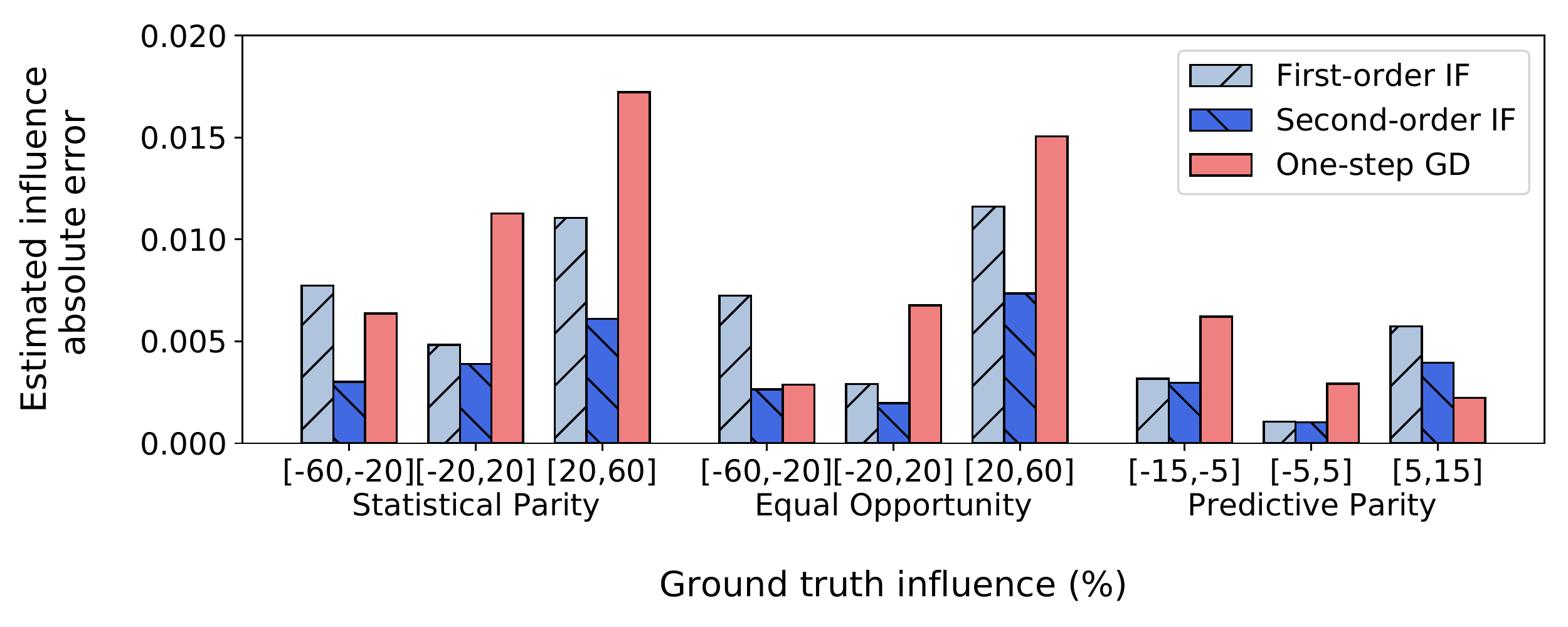}\delfigspace
    \caption{Logistic regression}
    \label{fig:exp:quality-lr}
  \end{subfigure}
  %%%%%%%%%%%%%%%%%%%%%%%%%%%%%%%%%%%%%%%%
  \begin{subfigure}{1.0\linewidth}
    \includegraphics[width=1.0\linewidth]{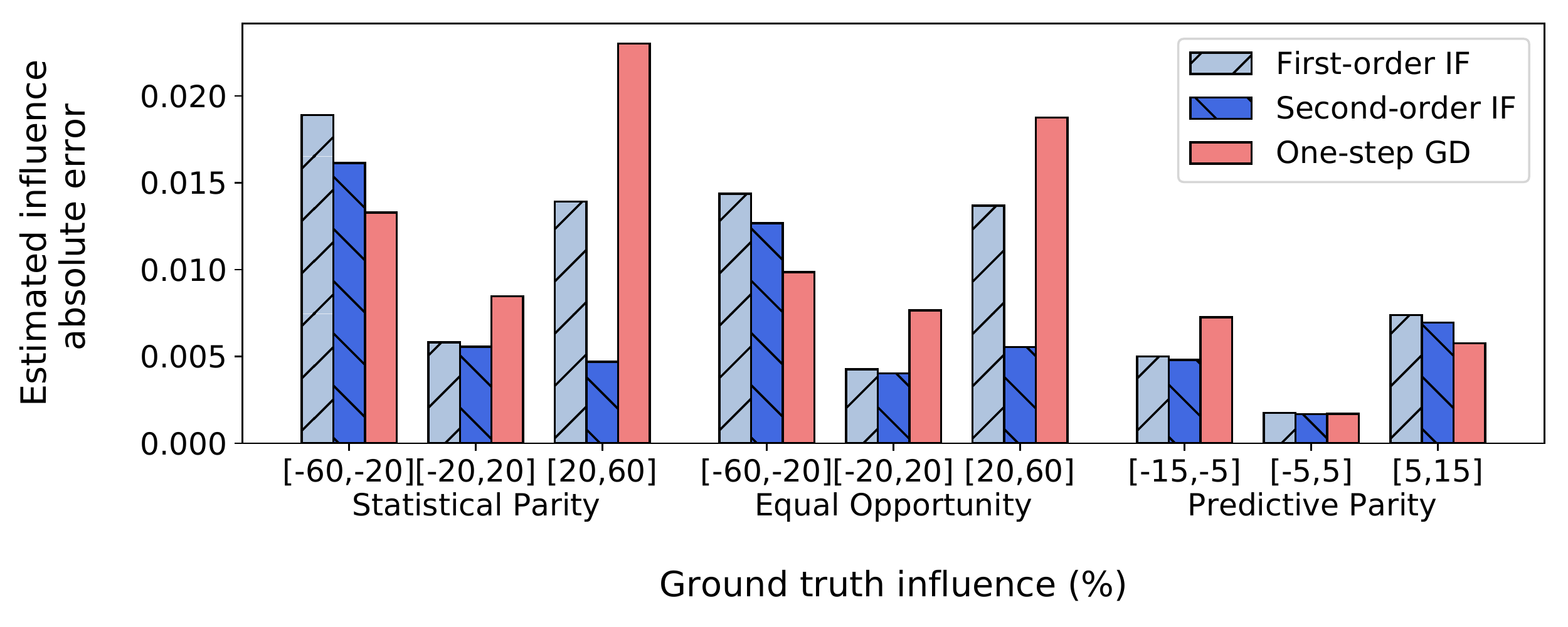}\delfigspace
    \caption{Neural networks (NNs).}
    \label{fig:exp:quality-nn}
  \end{subfigure}
  %%%%%%%%%%%%%%%%%%%%%%%%%%%%%%%%%%%%%%%%
  \begin{subfigure}{1\linewidth}
    \includegraphics[width=1.0\linewidth]{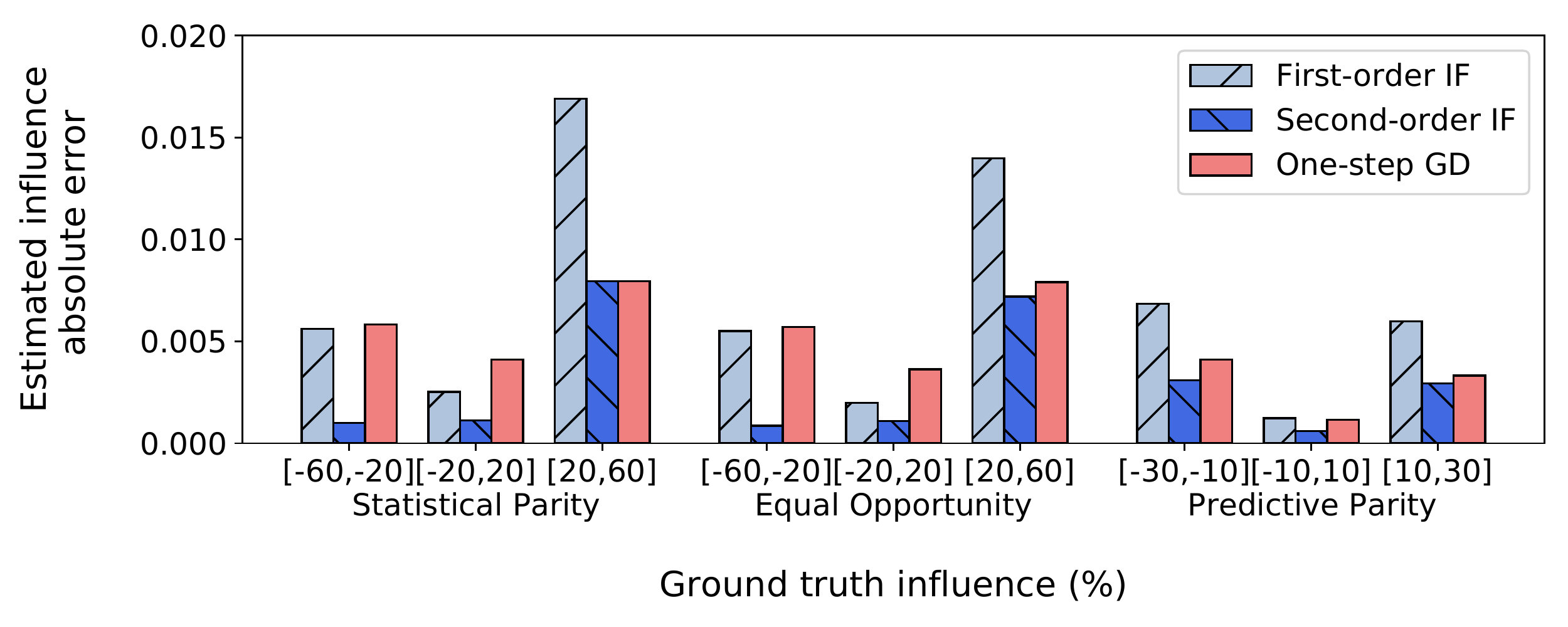}    \delfigspace
    \caption{Support vector machine (SVM).}
    \label{fig:exp:quality-svm}
  \end{subfigure}
  %%%%%%%%%%%%%%%%%%%%%%%%%%%%%%%%%%%%%%%%
  \caption{Comparing influence function approximations (\textsf{first-order IF} and \textsf{second-order IF}) against the one-step gradient descent approach (\textsf{one-step GD}) for estimating subset influence on the \textsf{German} dataset.}
\end{figure}

%%%%%%%%%%%%%%%%%%%%%%%%%%%%%%%%%%%%%%%%
% \begin{figure*}[t]
%     \centering
%     \includegraphics[scale=0.4]{plots/effectiveness.pdf}\delfigspace
%     \caption{Comparing influence function approximations (\textsf{first-order IF} and \textsf{second-order IF}) against the one-step gradient descent approach (\textsf{one-step GD}) on estimating ground truth subset influence on the \textsf{German} dataset using logistic regression. Influence function approximations better capture small changes in model parameters.
%     }\label{fig:exp:quality}
% \end{figure*}
%%%%%%%%%%%%%%%%%%%%%%%%%%%%%%%%%%%%%%%%
%%%%%%%%%%%%%%%%%%%%%%%%%%%%%%%%%%%%%%%%
\ignore{\begin{figure*}[t]
  \centering
  \vspace{-5mm}
	\includegraphics[width=0.8\linewidth]{plots/time.pdf}\delfigspace
	\caption{Time taken to compute subset influence for different subsets of the \textsf{German} dataset using logistic regression. Influence function approximations are significantly faster than model retraining, especially  % and one-step gradient descent
      for computing subset influence of smaller training data subsets. The results are averaged over ten runs.}
	\label{fig:exp:runtime-lr-nogd}
\end{figure*}}
%%%%%%%%%%%%%%%%%%%%%%%%%%%%%%%%%%%%%%%%

\begin{figure*}[t]
  \centering
    \subcaptionbox{Logistic regression. \label{fig:exp:runtime:logistic}}
    {\includegraphics[width=.33\textwidth]{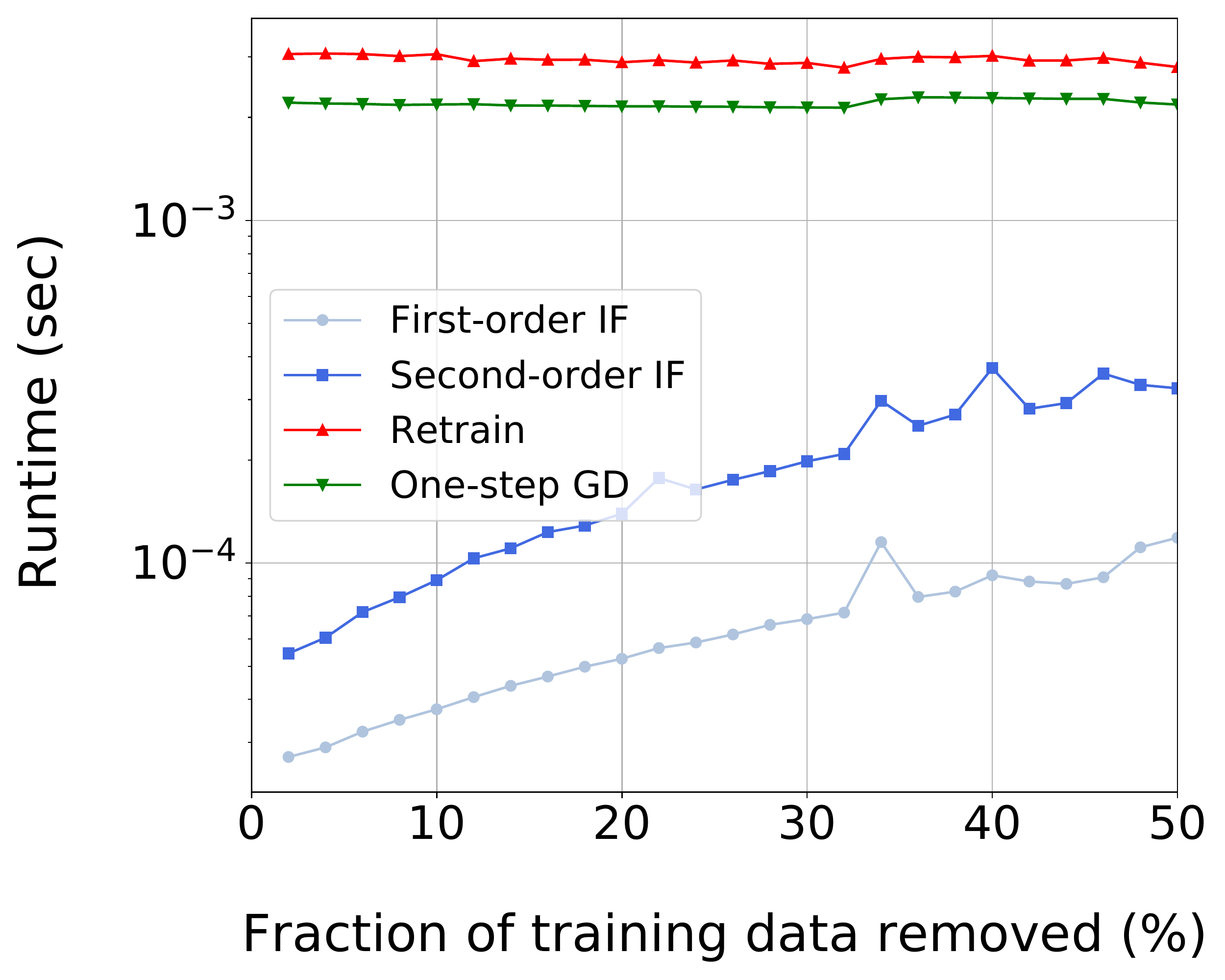}}
    \subcaptionbox{Support vector machine. \label{fig:exp:runtime:svm}}
    {\includegraphics[width=.33\textwidth]{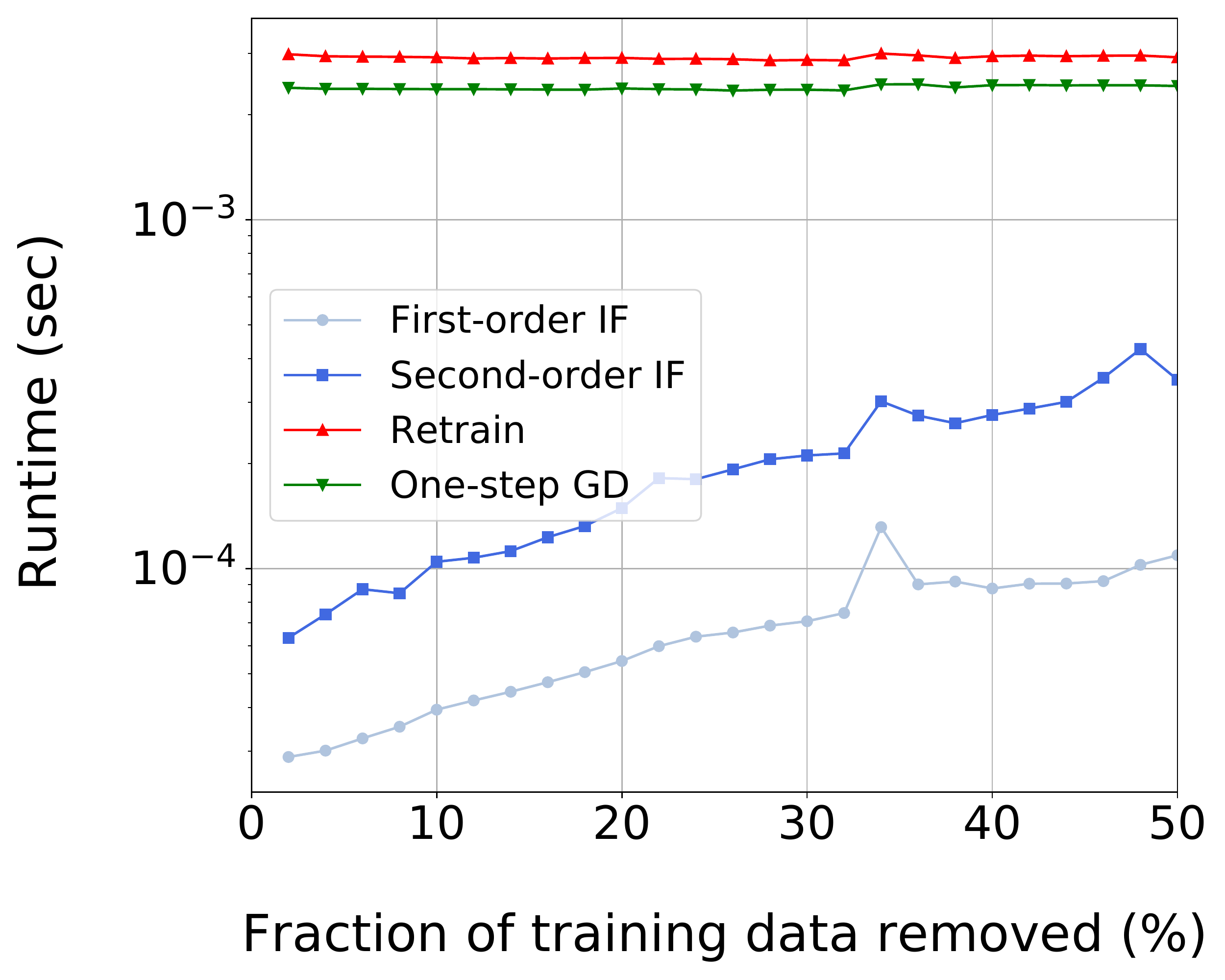}}
    \subcaptionbox{Neural network. \label{fig:exp:runtime:nn}}
    {\includegraphics[width=.33\textwidth]{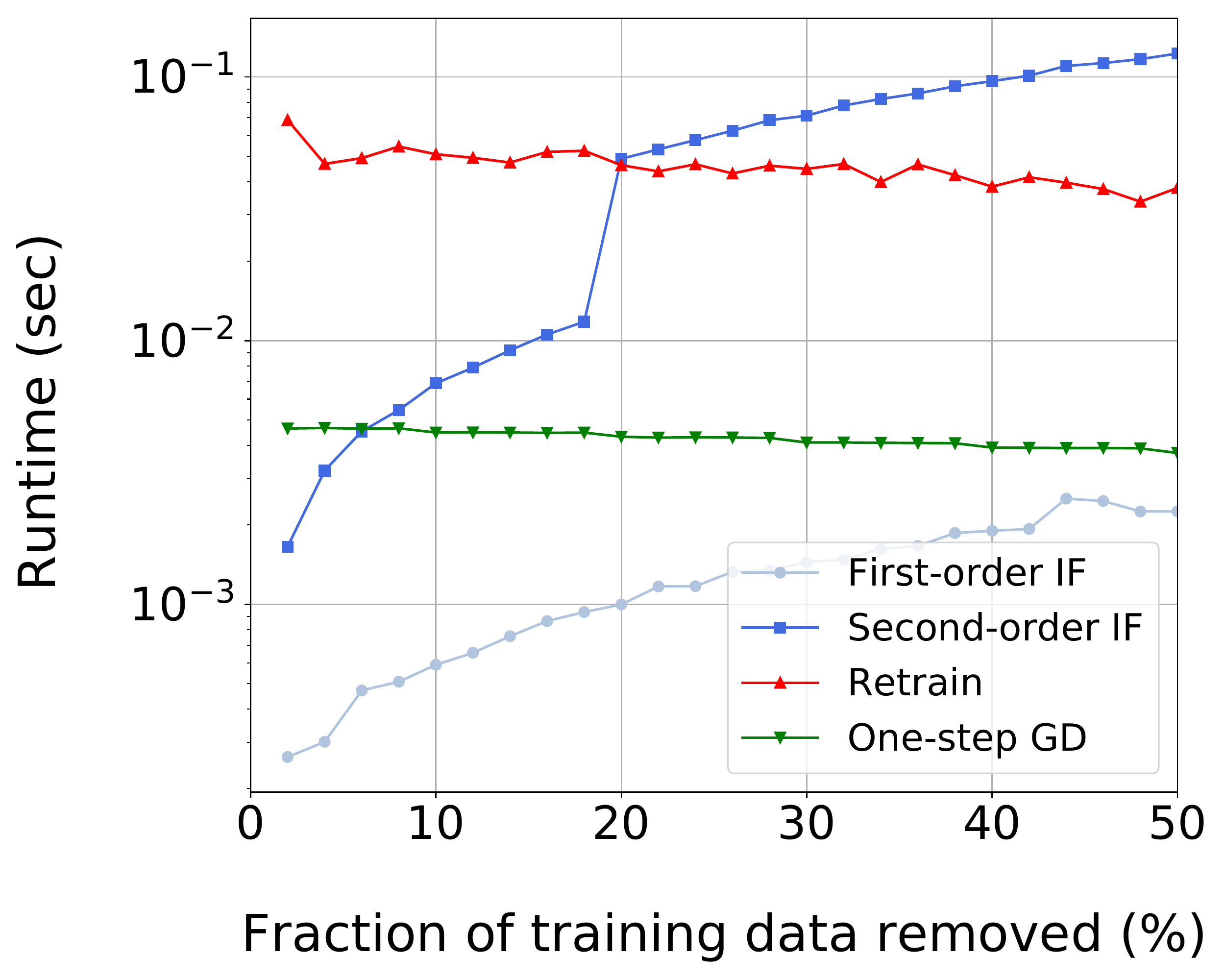}}
\\[-2mm]
    \caption{\revd{Runtime (averaged over 30 runs) for computing influence for  subsets of \textsf{German}. Influence function approximations are significantly faster than  retraining and one-step gradient descent influence for smaller  subsets.}}\label{fig:exp:runtime}
\end{figure*}

\newcommand{\codeanddataurl}{\href{https://github.com/romilapradhan/gopher}{https://github.com/romilapradhan/gopher}}

%%%%%%%%%%%%%%%%%%%%%%%%%%%%%%%%%%%%%%%%%%%%%%%%%%%%%%%%%%%%%%%%%%%%%%%%%%%%%%%%
\subsection{Datasets}
We use standard datasets from the fairness literature. The data and code for the experiments can be found at the project page\footnote{\codeanddataurl}.

\noindent\textbf{German Credit Data (German)~\cite{Dua2019}.} Personal, financial and demographic information ($20$ attributes) of $1,000$ bank account holders. The prediction task classifies individuals as good/bad credit risks. %\\
% \noindent
\textbf{Adult Income Data (Adult)~\cite{Dua2019}.} Demographic information, level of education, occupation, working hours, etc.,
  of $48,000$ individuals ($14$ attributes). % as well as information on their level of education, occupation, working hours, etc., as described by .
  The task predicts whether the annual income of an individual exceeds $\$50$K.
  % \\ \noindent
\textbf{Stop, Question, and Frisk Data (SQF)~\cite{sqf-data}.} Demographic and stop-related information for $72,548$ individuals stopped and questioned (and possibly frisked) by the NYC Police Department (NYPD). % during the years 2014-17.
    The classification task is to predict if a stopped individual % (who was stopped)
    will be frisked.\\
\iftechreport{\noindent\textbf{Salary~\cite{0df6080323bf4fe484391b75d1e1fe05}.} This dataset consists of salary and other information (on six variables) of $52$ tenure-track faculty in a small Midwestern college collected in the 1980s. The classification task predicts whether the salary of an individual exceeds $\$23,719$.\\}
%\noindent\textbf{Adult-sample.} This dataset is a sample of the \textsf{Adult} dataset and contains $4,884$ rows and $14$ attributes.
% \noindent\textbf{COMPAS~\cite{compas}.} This dataset contains information on offenders from Broward County, Florida. We consider the task of predicting whether an individual will recommit a crime within two years.

%%%%%%%%%%%%%%%%%%%%%%%%%%%%%%%%%%%%%%%%%%%%%%%%%%%%%%%%%%%%%%%%%%%%%%%%%%%%%%%%

%%%%%%%%%%%%%%%%%%%%%%%%%%%%%%%%%%%%%%%%%%%%%%%%%%%%%%%%%%%%%%%%%%%%%%%%%%%%%%%%
\vspace{-0.5cm}
\subsection{Setup}
\label{sec:exp:setup}
We considered three ML algorithms:
% linear regression~\cite{sklearn},
logistic regression, support vector machines, and \revc{a feed-forward neural network with 1 layer and 10 nodes (we provide details about the hyper-parameters for in the \sys repository\footnotemark)}. We used the PyTorch~\cite{pytorch} or sklearn~\cite{sklearn} implementation of these algorithms. In accordance with existing literature on evaluation of fairness of ML algorithms on these datasets~\cite{chiappa2019path}, the sensitive attributes are:  \attr{gender} (Adult),  \attr{age} (German), \attr{race} (SQF). We implemented our algorithms in Python and used PyTorch's autograd package to compute the gradients and Hessian. At start up, we pre-computed the Hessian and gradients for faster computation of the influence function approximations. % We report explanations generated for each dataset under different scenarios.
We split each dataset into training and test data, trained an ML model over the training data, and generated explanations using our pattern generation algorithm. We report the top-k explanations for each dataset under different scenarios. For each explanation, we report the pattern $\pat$, its support $\support(\pat)$, and the ground truth change in bias (reported in terms of statistical parity unless stated otherwise) achieved by removing the data corresponding to the pattern from the training data. We solve \Cref{eq:zup} using SciPy's \texttt{optimize} package for constrained optimization. For update-based explanations, we report the update to the data corresponding to the explanation's pattern and the change in bias resulting from the update. For both kinds of explanations, predicates that occur in more than one pattern are color-coded. We use the logistic regression model as the default ML model. % and statistical parity to indicate model bias unless we mention otherwise.

\revd{\noindent\textbf{Baseline.} As a competitor for our approach,  we trained a decision tree regressor (referred to as \textit{FO-tree}) over FO influence approximations of data points. FO-tree splits the training data into non-overlapping subsets based on the values of an attribute at a node of the tree. % the split attribute node.
% --- a data point belongs to one subset.
The path from the root to a node of the tree corresponds to a conjunction of predicates that characterize the data points represented by that node. To generate the top-$k$ explanations consisting of up to $l$ predicates, we identify the $k$ nodes from the root (level 0) to level $l$ that have the maximum combined influence of data points, and report the paths from the root node to these nodes. For example, to generate top-5 explanations with up to 3 predicates, we identify the 5 nodes up to a depth of 3 having the maximum FO influence and report the conjunction of predicates that form the path from the root to each of these node.}
% \vspace{-5mm}

% \noindent\textbf{Baselines.}
% \begin{itemize}
%     \item Overlap with ground truth explanations. For ground truth, refer HypDB and other papers where we know what the dominant biases are.
%     \item Summarize first-order explanations with decision tree
%     \item Shapley data values?
% \end{itemize}

%%%%%%%%%%%%%%%%%%%%%%%%%%%%%%%%%%%%%%%%%%%%%%%%%%%%%%%%%%

%\subsection{Evaluation of our approach}
%In this section, we evaluate the effectiveness of our approach in generating top-$k$ explanations. We assess the quality of our influence estimates, and evaluate our generated explanations.
%%%%%%%%%%%%%%%%%%%%%%%%%%%%%%%%%%%%%%%%%%%%%%%%%%%%%%%%%%%%%%%%%%%%%%%%%%%%%%%%
\vspace{-3mm}
\subsection{Causal Responsibility Approximations}
\label{exp:qual:responsibility}
In this set of experiments, we evaluated the quality of the proposed approaches for approximating influence, and thereby causal responsibility. The ground truth influence of subsets is computed by retraining the model after removing the subset from the training data. % However, as discussed in \Cref{sec:intro}, this process is expensive.
% In this set of experiments, we evaluate the proposed approximation methods.  % feasibility of influence function approximations as a substitute to the ground truth influence by comparing the effectiveness of first- and second-order influence function approximations against ground truth influence.
To estimate subset influence using FO approximations, we summed the individual influences, whereas the SO subset influence is computed using \Cref{eq:inf:second-order}. We also compared the approximations for the one-step gradient descent approach described in \Cref{sec:gdapprox}, because we use this approach for update explanations. % which approximates the updated model parameters from the original parameters using one step of gradient descent.

%%%%%%%%%%%%%%%%%%%%%%
\begin{table*}[t]
	\centering
	\begin{tabular}{|ccc|}
		\hline
	 \textsf{Pattern} & \textsf{Support} & \textsf{$\Delta_{bias}$}\\
		\hline
% 		\hline
		\begin{minipage}{.7\textwidth}
		    \includegraphics{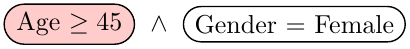}
		\end{minipage} & $5.00\%$ & $55.2\%$\\
		\begin{minipage}{.7\textwidth}
		    \includegraphics{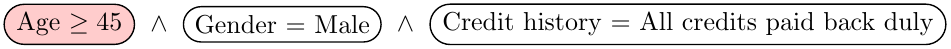}
		\end{minipage} & $6.25\%$ & $35.8\%$\\
		\begin{minipage}{.7\textwidth}
		    \includegraphics{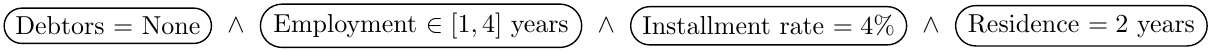}
		\end{minipage} & $5.13\%$ & $14.8\%$\\
		\hline
	\end{tabular}\\[1mm]
\caption{Top-$3$ explanations for \textsf{German} ($\tau=5\%$, logistic regression, runtime=$18$s).}
	\label{tab:german:5pc:lr}
\end{table*}
\iftechreport{\begin{table*}[t]
	\centering
	\begin{tabular}{|ccc|}
		\hline
	 \textsf{Pattern} & \textsf{Support} & \textsf{$\Delta_{bias}$}\\
		\hline
% 		\hline
		\begin{minipage}{.8\textwidth}
		    \includegraphics{plots/exp-german-pp-5pc-0-svm.pdf}
		\end{minipage} & $3.125\%$ & $91.2\%$\\
		\begin{minipage}{.8\textwidth}
		    \includegraphics{plots/exp-german-pp-5pc-1-svm.pdf}
		\end{minipage} & $3.25\%$ & $93.1\%$\\
		\begin{minipage}{.8\textwidth}
		    \includegraphics{plots/exp-german-pp-5pc-2-svm.pdf}
		\end{minipage} & $3.875\%$ & $91.0\%$\\
		\begin{minipage}{.8\textwidth}
		    \includegraphics{plots/exp-german-pp-5pc-3-svm.pdf}
		\end{minipage} & $4.75\%$ & $88.2\%$\\
		\begin{minipage}{.8\textwidth}
		    \includegraphics{plots/exp-german-pp-5pc-4-svm.pdf}
		\end{minipage} & $3.125\%$ & $90.0\%$\\
		\hline
	\end{tabular}\\[1mm]
    \caption{Top-$5$ explanations for \textsf{German} ($\tau=5\%$, $\Delta_{bias}$ represents the ground truth reduction in bias computed over predictive parity, support vector machine (SVM).)}
	\label{tab:german:5pc:svm}
\end{table*}
}
\ignore{\begin{table*}[t]
  \centering
  \vspace{-6mm}
	\begin{tabular}{|ccc|}
		\hline
	 \textsf{Pattern} & \textsf{Support} & \textsf{$\Delta_{bias}$}\\
		\hline
% 		\hline
		\begin{minipage}{.6\textwidth}
		    \includegraphics{plots/exp-adult-spd-5pc-0-nn.pdf}
		\end{minipage} & $6.83\%$ & $8.2\%$\\
		\begin{minipage}{.6\textwidth}
		    \includegraphics{plots/exp-adult-spd-5pc-1-nn.pdf}
		\end{minipage} & $7.75\%$ & $12.4\%$\\
		\begin{minipage}{.6\textwidth}
		    \includegraphics{plots/exp-adult-spd-5pc-2-nn.pdf}
		\end{minipage} & $7.95\%$ & $1.1\%$\\
		\begin{minipage}{.6\textwidth}
		    \includegraphics{plots/exp-adult-spd-5pc-3-nn.pdf}
		\end{minipage} & $6.17\%$ & $4.0\%$\\
		\begin{minipage}{.6\textwidth}
		    \includegraphics{plots/exp-adult-spd-5pc-4-nn.pdf}
		\end{minipage} & $7.42\%$ & $4.6\%$\\
		\hline
	\end{tabular}\\[1mm]
    \caption{Top-$5$ explanations on a $10\%$ sample of \textsf{Adult} ($\tau=5\%$, neural networks, runtime=$5$s).
	}
	\vspace{-8mm}
	\label{tab:adult:5pc:nn}
\end{table*}
}

\begin{table*}[t]
  \centering
  \vspace{-7mm}
	\begin{tabular}{|ccc|}
		\hline
	 \textsf{Pattern} & \textsf{Support} & \textsf{$\Delta_{bias}$}\\
		\hline
% 		\hline
		\begin{minipage}{.6\textwidth}
		    \includegraphics{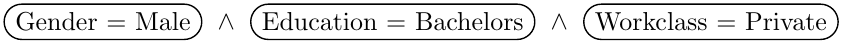}
		\end{minipage} & $7.89\%$ & $12.00\%$\\
		\begin{minipage}{.6\textwidth}
		    \includegraphics{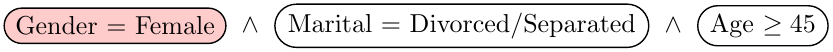}
		\end{minipage} & $6.27\%$ & $11.01\%$\\
		\begin{minipage}{.6\textwidth}
		    \includegraphics{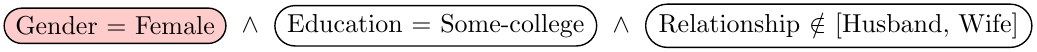}
		\end{minipage} & $7.39\%$ & $6.02\%$\\
		\hline
	\end{tabular}\\[1mm]
    \caption{Top-$3$ explanations on \textsf{Adult} ($\tau=5\%$, neural networks, runtime=$56$s).
	}
	\vspace{-4mm}
	\label{tab:adult:5pc:nn}
\end{table*}

\begin{table*}[t]
	\centering
  \vspace{-3mm}
	\begin{tabular}{|ccc|}
		\hline
	 \textsf{Pattern} & \textsf{Support} & \textsf{$\Delta_{bias}$}\\
		\hline
% 		\hline
		\begin{minipage}{.8\textwidth}
		    \includegraphics{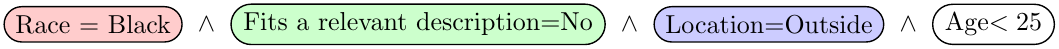}
		\end{minipage} & $16.89\%$ & $25.6\%$\\
		\begin{minipage}{.8\textwidth}
		    \includegraphics{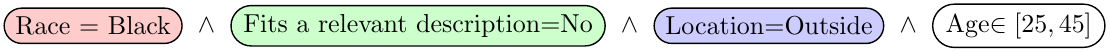}
		\end{minipage} & $12.95\%$ & $13.7\%$\\
		\begin{minipage}{.8\textwidth}
		    \includegraphics{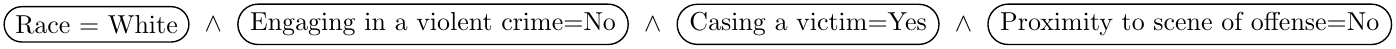}
		\end{minipage} & $7.04\%$ & $8.16\%$\\
		\hline
	\end{tabular}\\[1mm]
\caption{Top-$3$ explanations for \textsf{SQF} ($\tau=5\%$, logistic regression, runtime=$91$s).}
	\label{tab:sqf:5pc:lr}
	\vspace{-3mm}
\end{table*}

\iftechreport{
\begin{table*}[h]
	\centering
	\begin{tabular}{|p{10cm}|c|c|}
		\hline
	 \textsf{Pattern} & \textsf{Support} & \textsf{$\Delta_{bias}$} \\
		\hline \hline
		Rank=Associate Professor, Sex=Female & $2.44\%$ & $16.6\%$\\
		\hline
        Rank=Associate Professor, Years in current role $\in [5, 10)$ & $12.2\%$ & $75.1\%$\\
		\hline
        Rank=Associate Professor, Degree=Doctorate, Years in current role $\in [10, 15)$  & $4.88\%$ & $12.2\%$\\
		\hline
		Degree=Masters, Years since degree$<=10$& $2.44\%$ & $3.6\%$\\
		\hline
		Degree=Doctorate, Years since degree$<\in (10, 20]$ & $26.83\%$ & $33.42\%$\\
		\hline
	\end{tabular}
	\caption{Top-$5$ explanations for \textsf{Salary} ($\tau=1\%$, logistic regression, runtime=0.493s).}
	\label{tab:salary:bias}
\end{table*}
}
%%%%%%%%%%%%%%%%%%%%%%%%%%%%%%%%%%%%%%%%%%%%%%%%%%%%%%%%%%%%%%%%%%%%%%%%%%%%%%%%
\partitle{Effectiveness}
\revc{In \Cref{fig:exp:quality-lr,fig:exp:quality-nn,fig:exp:quality-svm}, we report the absolute deviation of the influence estimated by the different methods from the ground truth (y-axis) % with respect to the ground truth influence
  for the fairness definitions (x-axis) described in \Cref{sec:prelim}.
We trained different classifiers on the German dataset. We observed that, as expected, FO influence approximations and one-step gradient descent deviated more from ground truth influence than SO influence. FO influence approximation exhibits larger errors because it does not account for possible correlations among data points in the subset~\cite{basu2020second}; one-step gradient descent influence approximation is not accurate because it uses a single step of gradient descent instead of iterating until the model parameters converge. This behavior is especially evident when the size of the group is large, which usually corresponds to large influence (the bars on the outside of a metric).
% For SVM (\Cref{fig:exp:quality-svm}), the one-step gradient approximation outperformed influence functions for large ground truth influence but exhibited larger error for smaller ground truth influence.
% For neural networks that use gradients during training, one-step gradient consistently performs worse than SO influence approximations.
The key takeaway from this experiment is that SO influence functions closely approximate ground truth influence especially when model parameters do not change substantially (the middle bars).}

% In \Cref{fig:exp:quality}, we report the absolute deviation of the influence estimated by the different methods from ground truth (y-axis) with respect to the ground truth influence for different fairness definitions (x-axis), described in \Cref{sec:prelim}.
% For this experiment, we used the German dataset and trained a logistic regression. We observed that, as expected, first-order influence function approximations deviated the most from ground truth influence because they did not account for data correlations among points in the subset. Second-order influence functions, for the most part, exhibited less error on ground truth influence estimation compared to first-order approximations. We also observed that one-step gradient approximation outperformed influence function approximations for large ground truth influence but exhibited larger error for smaller ground truth influence. Perhaps the key takeaway from this experiment is that lower-order influence functions closely approximate ground truth influence when model parameters do not change substantially (the middle bin for each method).

%%%%%%%%%%%%%%%%%%%%%%%%%%%%%%%%%%%%%%%%%%%%%%%%%%%%%%%%%%%%%%%%%%%%%%%%%%%%%%%%
\partitle{Efficiency}
%
% \boris{We should probably say what method is used to train the models, because one of the reviewers asked about SGD.}
\revd{In \Cref{fig:exp:runtime}, we report the average time taken by each method when subsets of varying sizes are removed from the training data. The brute force approach --  retraining the model after removing the subset -- is typically more than two orders of magnitude slower than even the most expensive method, % and motivates the need for a cheaper alternative,
  especially when the fraction of removed training data points is small
% (e.g., influence functions are $\sim 20$ x faster when 0\%-10\% of the training dataset is deleted)
(e.g., influence functions are up to $4$ orders of magnitude faster when 0\%-10\% of the training dataset is deleted). We also observe that the one-step gradient descent approach, with the exception for SO influence functions on neural networks, while effective in estimating ground truth influence (as shown in \Cref{fig:exp:quality-lr}, \Cref{fig:exp:quality-svm}), is significantly slower than
influence functions for estimating subset influence. % which is a primary constituent of our task of generating interesting explanations that are usually small subsets.
Note that the time cost for retraining is close to that of one-step gradient descent because we used the initial model parameters to speed up convergence during retraining. In practice, we have to adjust the learning rate for one-step gradient descent over multiple iterations, which makes it even more expensive.
% One-step gradient descent takes so much time because we used PyTorch's autograd implementation to compute the gradients.
% On the other hand, influence function approximations are shown to be almost an order of magnitude faster than the one-step gradient descent approach for smaller subset sizes.
% Our goal for efficiency is to provide an interactive subset influence computation time for model developers and end-users of the ML algorithms. As such,
}
We conclude that retraining the model and using the one-step gradient descent approach are not feasible for generating the top-$k$ explanations. % by evaluating multiple subsets.
The one-step gradient descent, however, is useful for generating update-based explanations as described in \Cref{sec:gdapprox}: influence functions are not applicable to the optimization problem we have to solve for updates, and retraining is also not an option.

\vspace{-3mm}
\subsection{\textbf{End-to-end Performance}}
\label{sec:exp:endtoend}
Next, we evaluate the performance of \sys\ for generating the top-$k$ explanations for ML algorithms trained on different datasets.

%%%%%%%%%%%%%%%%%%%%%%%%%%%%%%%%%%%%%%%%
\partitle{German}
This dataset is biased toward older individuals and considers them less likely to be characterized as high credit risks.
In \Cref{tab:german:5pc:lr}, we report the top-$3$ explanations up to 4 predicates generated by \sys\ (with their support and ground truth influence) sorted by their interestingness score. We observe that one subset of $5\%$ of data points explains more than half of the model bias, whereas another subset of around $6\%$ of data points reduces bias by almost $36\%$. \reva{These explanations highlight fractions of training data that may have potential errors and hence, need attention. On inspection, we found that they correspond to training data points where older individuals are primarily labeled as low credit risks. By removing these individuals, the probability of an individual being classified as a high/low credit risk is uniformly distributed across the sensitive attribute age. As a result, the model's dependency on age is reduced, thus reducing overall model bias. Note that the top-2 explanations consist of predicates with the sensitive attribute for this dataset, signifying its importance in bias reduction.} \revd{In comparison, we made the following observations about the explanations generated by FO-tree: entirely different regressor trees, and hence different explanations, were generated depending upon whether sklearn or PyTorch was used to fit the model. While the sensitive attribute (age) formed the root node for the FO-tree generated on the PyTorch model, a non-intuitive attribute (installment rate) was the root in the FO-tree for the sklearn model. The top-2 explanations from FO-tree over the PyTorch model were consistent with our explanations whereas those generated from FO-tree over the sklearn model were less compact (consisting of 4 predicates each). Their (support, bias reduction) were ($6.13\%, 32.3\%$), ($5.63\%, 33.1\%$) and ($12.9\%, 8\%$), respectively.}

\begin{table*}[t]
	\centering
	\begin{tabular}{|ccc|}
		\hline
	 \textsf{Pattern} & \textsf{Support} & \textsf{$\Delta_{bias}$}\\
		\hline
% 		\hline
		\begin{minipage}{.7\textwidth}
		    \includegraphics{plots/explanations/german/exp-german-spd-5pc-0-lr.pdf}
		\end{minipage} & $5\%$ & $55.2\%$\\
		\begin{minipage}{.7\textwidth}
		    \includegraphics{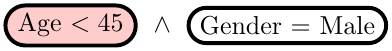}
		\end{minipage} &  & $42.0\%	\downarrow$\\
		\hline
		\begin{minipage}{.7\textwidth}
		    \includegraphics{plots/explanations/german/exp-german-spd-5pc-1-lr.pdf}
		\end{minipage} & $6.25\%$ & $35.8\%$\\
		\begin{minipage}{.7\textwidth}
		    \includegraphics{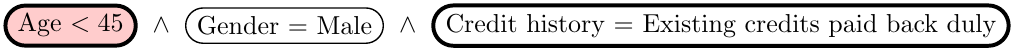}
		\end{minipage} &  & $21.6\%	\downarrow$\\
		\hline
		\begin{minipage}{.7\textwidth}
		    \includegraphics{plots/explanations/german/exp-german-spd-5pc-2-lr.pdf}
		\end{minipage} & $5.13\%$ & $14.8\%$\\
		\begin{minipage}{.7\textwidth}
		    \includegraphics{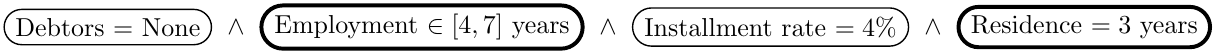}
		\end{minipage} &  & $5.4\%	\downarrow$\\
		\hline
	\end{tabular}\\[1mm]
\caption{Update-based explanations for the top-3 explanations for \textsf{German} ( $\tau=5\%$). Updates to the original explanation (top) are shown with a bold outline (bottom). Change in bias reduction due to the update $\Delta_{bias}$ is represented by $\downarrow$ (decrease) or $\uparrow$ (increase). Average time taken to update each point = $0.22$s.}
	\label{tab:german:5pc:update}
\end{table*}

%%%%%%%%%%%%%%%%%%%%%%%%%%%%%%%%%%%%%%%
\begin{table*}[t]
  \centering
  \vspace{-7mm}
	\begin{tabular}{|ccc|}
		\hline
	 \textsf{Pattern} & \textsf{Support} & \textsf{$\Delta_{bias}$}\\
		\hline
% 		\hline
		\begin{minipage}{.7\textwidth}
		    \includegraphics{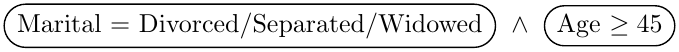}
		\end{minipage} & $9.1\%$ & $13.3\%$\\
		\begin{minipage}{.7\textwidth}
		    \includegraphics{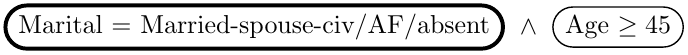}
		\end{minipage} & & $13.8\%\uparrow$\\
		\hline
		\begin{minipage}{.7\textwidth}
		    \includegraphics{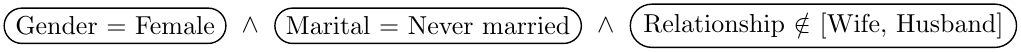}
		\end{minipage} & $14.3\%$ & $19.1\%$\\
		\begin{minipage}{.7\textwidth}
		    \includegraphics{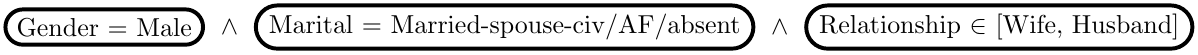}
		\end{minipage} & & $0.6\%\downarrow$\\
		\hline
		\begin{minipage}{.7\textwidth}
		    \includegraphics{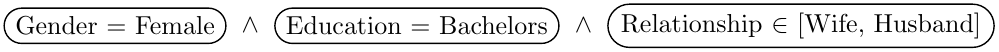}
		\end{minipage} & $7.6\%$ & $9.1\%$\\
		\begin{minipage}{.7\textwidth}
		    \includegraphics{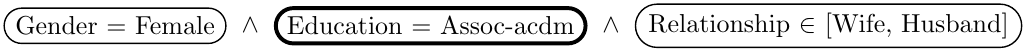}
		\end{minipage} &  & $9.5\%\uparrow$\\
		\hline
	\end{tabular}\\[1mm]
%%%%%%%%%%%%%%%%%%%%
\caption{Update-based explanations for the top-3 explanations for \textsf{Adult} ( $\tau=5\%$). Updates to the original explanation (top) are shown in bold outline (bottom). Change in bias reduction $\Delta_{bias}$ is represented by $\downarrow$ (decrease) or $\uparrow$ (increase). Average time taken to update each point = $0.24$s.}
%%%%%%%%%%%%%%%%%%%%
\label{tab:adult:5pc:lr:update}
\end{table*}

%%%%%%%%%%%%%%%%%%%%%%%%%%%%%%%%%%%%%%%
\begin{table*}[t]
  \centering
  \vspace{-7mm}
	\begin{tabular}{|ccc|}
		\hline
	 \textsf{Pattern} & \textsf{Support} & \textsf{$\Delta_{bias}$}\\
		\hline
% 		\hline
		\begin{minipage}{.8\textwidth}
		    \includegraphics{plots/explanations/sqf/exp-sqf-spd-5pc-0-lr.pdf}
		\end{minipage} & $16.9\%$ & $25.6\%$\\
		\begin{minipage}{.8\textwidth}
		    \includegraphics{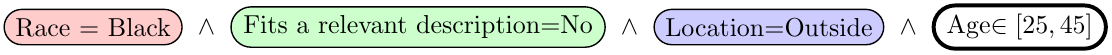}
		\end{minipage} & & $14.3\%\downarrow$\\
		\hline
		\begin{minipage}{.8\textwidth}
		    \includegraphics{plots/explanations/sqf/exp-sqf-spd-5pc-1-lr.pdf}
		\end{minipage} & $13.0\%$ & $13.7\%$\\
		\begin{minipage}{.8\textwidth}
		    \includegraphics{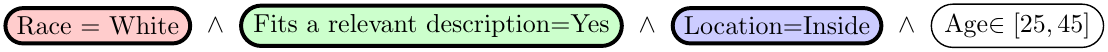}
		\end{minipage} & & $8.6\%\downarrow$\\
		\hline
		\begin{minipage}{.8\textwidth}
		    \includegraphics{plots/explanations/sqf/exp-sqf-spd-5pc-2-lr.pdf}
		\end{minipage} & $7.0\%$ & $8.2\%$\\
		\begin{minipage}{.8\textwidth}
		    \includegraphics{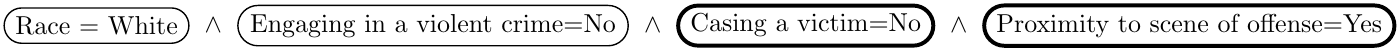}
		\end{minipage} &  & $13.7\%\uparrow$\\
		\hline
	\end{tabular}\\[1mm]
%%%%%%%%%%%%%%%%%%%%
\caption{\revd{Update-based explanations for the top-3 explanations for \textsf{SQF} ( $\tau=5\%$). Updates to the original explanation (top) are shown in bold outline (bottom). Change in bias reduction $\Delta_{bias}$ is represented by $\downarrow$ (decrease) or $\uparrow$ (increase). Average time taken to update each point = $0.5$s}} \vspace{-5mm}
%%%%%%%%%%%%%%%%%%%%
\label{tab:sqf:5pc:lr:update}
\end{table*}

%%%%%%%%%%%%%%%%%%%%%%%%%%%%%%%%%%%%%%%%
\partitle{Adult}
This dataset has been at the center of several studies that analyze the impact of gender~\cite{TAGH+17,DBLP:conf/sigmod/SalimiGS18} and has been shown to be inconsistent: income attributes for married individuals report household income. The dataset has more married males, indicating a favorable bias toward males. As seen in \Cref{tab:adult:5pc:nn}, \revc{the sensitive attribute} gender plays an important role in all of the explanations. \revc{In this set of experiments on neural networks, we observed that second-order influence functions did not estimate the model parameters accurately and greatly underestimated the ground truth influence. This observation was consistent with the analysis provided in~\cite{basu2020second} for neural networks where the influence of a group of data points has low correlation with ground truth influence, and second-order influences underestimate ground truth influence. Our approach hinges on the applicability of influence functions to correctly estimate the model parameters around the optimal parameters-- an assumption that might not hold for neural networks. \sys\ still identified patterns that reduce model bias to some extent. In comparison, the top-3 patterns returned by FO-tree had higher support and lower ground truth influence: ($10.9\%, 9.8\%$), ($13.2\%, 10.8\%$) and ($5.9\%, 11\%$), respectively.} Note that even though the dataset has single predicates (\textsf{[marital = Married]}, $\sim47\%$ data) that remove bias almost completely, these predicates do not rank in the top-$k$ explanations because of their low interestingness scores.
% A key observation was that with this dataset, subset removal does not reduce bias drastically because even with as few as $\sim6\%$ training data, a subset has more than $200$ data points.

\revd{\partitle{SQF}
This dataset highlighted that the practices of NYPD in stopping, questioning and frisking blacks (and latinos) more often compared to whites were unconstitutional and violated Fourth Amendment rights~\cite{sqf-nyclu}. In Table~\ref{tab:sqf:5pc:lr}, our top-3 explanations identify patterns consisting of the protected group that were frisked and the privileged group that were not frisked. Because of these data points, the model learns that data points belonging to the privileged (protected) group are less (more) likely to be frisked, and therefore, cause bias in the model. By removing these data points, the privileged group becomes less correlated with the `no frisk' outcome, thus reducing model bias. The topmost explanation generated by the FO-tree [(location=Outside) $\wedge$ (race=White) $\wedge$ (build<>Thin) $\wedge$ (does not fit a relevant description)] had similar support ($13.2\%$) and bias reduction ($27\%$) as our topmost explanation. The other two patterns had lower bias reduction (one of them had $0.15\%$ reduction in bias) and greater support, and hence were less interesting.}
% A key observation was that with this dataset, subset removal does not reduce bias drastically because even with as few as $\sim6\%$ training data, a subset has more than $200$ data points.

\iftechreport{
\partitle{Salary}
The dataset was presented in legal proceedings for which discrimination against women in salary was at issue, and was reported to show salary bias in favor of men~\cite{0df6080323bf4fe484391b75d1e1fe05}. We report the top-$5$ explanations generated by \sys\ in Table~\ref{tab:salary:bias}. In this dataset, women are concentrated in the lowest ranks ($54\%$ of all assistant professors are women compared to $18\%$ of associate/full professors) and have fewer years of service in the current role ($93\%$ of women have fewer than $10$ years of service compared to $76\%$ of men). Deleting data in the first two explanations also deletes women in the high ranked groups and breaks the model's dependency on sex and years of service (which in turn is correlated with sex), thus contributing to lower bias. The rest of the explanations reduces bias by removing men with higher salary.
% http://www.ru.ac.bd/wp-content/uploads/sites/25/2019/03/304_03_Weisberg-Applied-Linear-Regression-Wiley-2013.pdf
% http://math.bu.edu/people/cgineste/classes/ma575/h/Solution%20Manuel.pdf
}

\iftechreport{
\begin{table*}[h]
	\centering
	\begin{tabular}{|p{10cm}|c|c|}
		\hline
	 \textsf{Pattern} & \textsf{Support} & \textsf{$\Delta_{bias}$} \\
		\hline \hline
		Rank=Associate Professor, Sex=Female & $2.44\%$ & $16.6\%$\\
		Rank=Associate Professor, Sex=Male &  & $18.1\%\uparrow$\\
		\hline
        Rank=Associate Professor, Years in current role $\in [5, 10)$ & $12.2\%$ & $75.1\%$\\
        Rank=Full Professor, Years in current role $\in [10, 15)$ & & $88.1\%\uparrow$\\
		\hline
        Rank=Associate Professor, Degree=Doctorate, Years in current role $\in [10, 15)$  & $4.88\%$ & $12.2\%$\\
        Rank=Full Professor, Degree=Doctorate, Years in current role $\in [10, 15)$  &  & $10.2\%\downarrow$\\
		\hline
		Degree=Masters, Years since degree$<=10$& $2.44\%$ & $3.6\%$\\
		Degree=Doctorate, Years since degree$\in (10, 20]$&  & $3.3\%\downarrow$\\
		\hline
		Degree=Doctorate, Years since degree$\in (10, 20]$ & $26.83\%$ & $33.42\%$\\
		Degree=Doctorate, Years since degree$> 20$ & & $16.01\%\downarrow$\\
		\hline
	\end{tabular}
	\caption{Update-based explanations for the top-5 explanations for \textsf{Salary} ( $\tau=1\%$). Updates to the original explanation (top) are shown with a bold outline (bottom). Change in bias reduction $\Delta_{bias}$ is represented by $\downarrow$ (decrease) or $\uparrow$ (increase). Time=$1.58$s).]}
	\label{tab:salary:update}
\end{table*}
}
%%%%%%%%%%%%%%%%%%%%%%%%%%%%%%%%%%%%%%%%

%%%%%%%%%%%%%%%%%%%%%%%%%%%%%%%%%%%%%%%%%%%%%%%%%%%%%%%%%%%%%%%%%%%%%%%%%%%%%%%%
%\vspace{-0.4cm}
\subsection{Update-based Explanations}
In these experiments, we generated updates for the top-$k$ explanations. For each explanation, we provide the perturbation that would result in the maximum  bias reduction.

%%%%%%%%%%%%%%%%%%%%%%%%%%%%%%%%%%%%%%%%7
\partitle{German}
As seen in \Cref{tab:german:5pc:update},  updates typically involve perturbing the protected attribute (\attr{age}). For example, the first explanation suggests that by updating data points satisfying the pattern such that after the update they are in the protected  instead of the privileged group, and by changing the gender to increase the chance of a positive outcome,  % now has attribute values that are more likely to result in a favorable model outcome,
bias is reduced by $42\%$. In this case, we found an update
% for the pattern
that would reduce model bias but not
by as much as deleting those points would.
% by as much as it would reduce by deleting those data points.
Similarly, in explanation 2, older individuals that have a good credit history are considered low credit risks. By changing their age group to the protected group and credit history to a worse level, we now associate younger individuals with good credit risk and reduce model bias (albeit by an amount smaller than if the group was removed). The key takeaway here is that we can reduce model bias by updating the training data points referring to these patterns instead of removing them altogether.

%%%%%%%%%%%%%%%%%%%%%%%%%%%%%%%%%%%%%%%%
\partitle{Adult}
We report the update-based explanations for this dataset in \Cref{tab:adult:5pc:lr:update}. As mentioned before, this subset is biased toward married individuals and males. In explanation 1, individuals that were not married had lower income. By changing their marital status to married, we were able to reduce model bias by at least as much as would have been achieved were those patterns deleted. Note, however, that in explanation 2, even by changing the gender and marital status to the preferred attribute values, we could not find an appropriate update that would reduce the bias. After marital status, we found that education accounts for most of the bias~\cite{DBLP:conf/sigmod/SalimiGS18}-- individuals with a higher level of education are associated with higher incomes. In explanation 2, by changing the education level we were able to reduce bias by almost the same amount as would have been achieved if the subset were altogether removed.

%%%%%%%%%%%%%%%%%%%%%%%%%%%%%%%%%%%%%%%%
\revd{\partitle{SQF}
In \Cref{tab:sqf:5pc:lr:update}, we observe that changing particular attribute values can help us avoid discriminatory behavior for this dataset. For example, before the update in explanation 3, whites that appeared to be casing a victim (or studying them for probable targets) were not frisked --- a clear case of discrimination. In this case, we were able to find an update such that they did not case a victim even when close to the crime scene, and achieved even more reduction in model bias than if this subpopulation is removed altogether. Similarly, frisking blacks even when they do not fit a relevant description was biased against them. In this case, updating this subset of data points such that whites that fit a relevant description are frisked reduced the model bias but less than if the subset is removed.}

\iftechreport{
\partitle{Salary}
We report the update-based explanations for this dataset in \Cref{tab:salary:update}. The first update explanation decreases (increases) the fraction of protected (privileged) data points with negative outcome and thus, reduces bias more than before. The second update also changes the sensitive attribute of negative outcome data points from protected to privileged, thus reducing the salary bias. By changing the rank and years in service of negative outcome data points, this update also weakens the model's dependency on rank and years in service. The rest of the updates are on positive outcome data points and bolster the model's dependency on rank, degree and years since highest degree; since rank and years since highest degree are correlated with sex, these updates do not achieve a greater reduction in bias.
}

\begin{figure}[t]
  \centering
    {\includegraphics[width=\columnwidth]{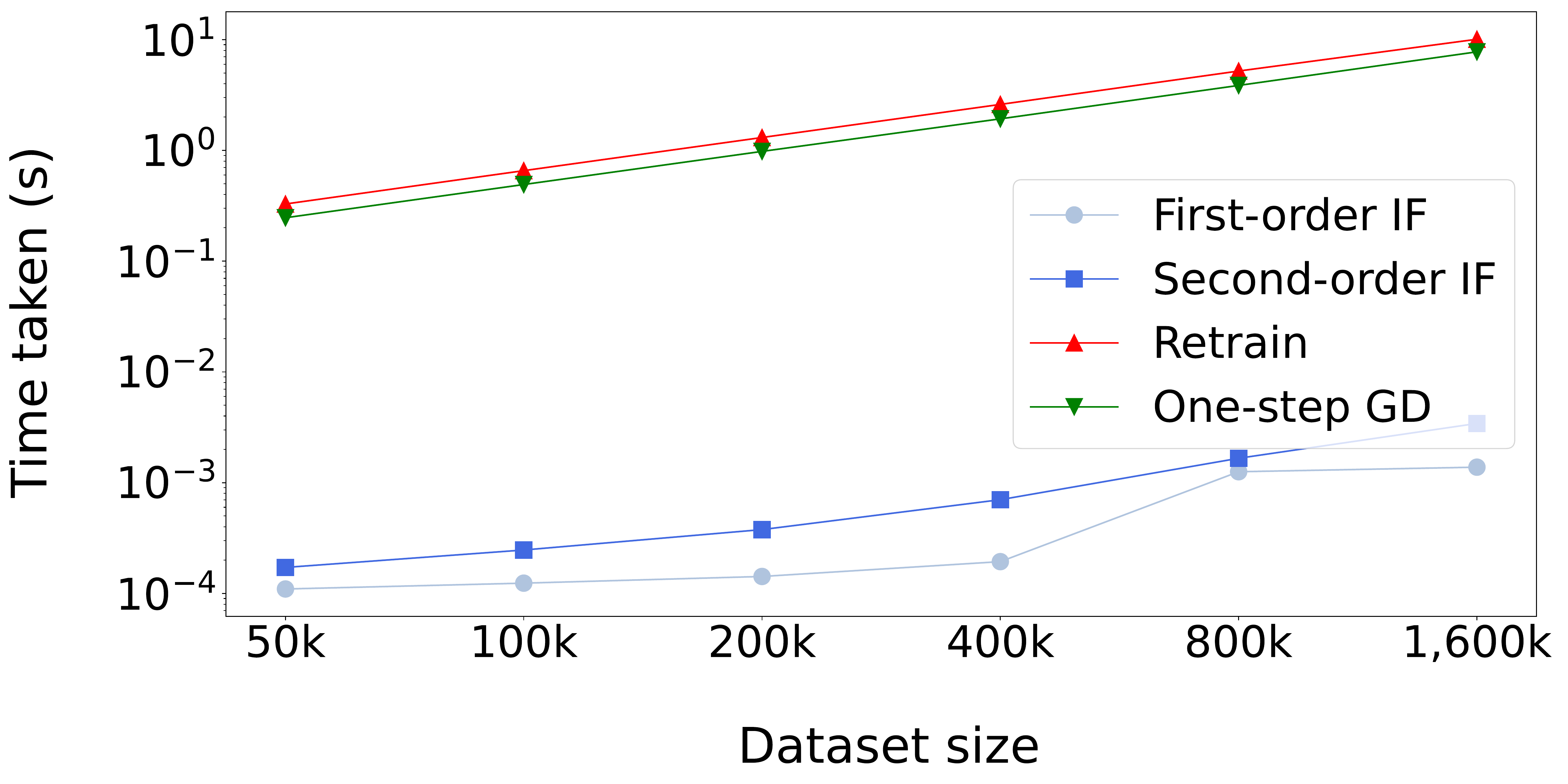}}
    \vspace{-7mm}\caption{\revd{Runtime vs. dataset size.} \label{fig:exp:runtime:datasize}}
\end{figure}

% \vspace{-0.4cm}
\revd{\subsection{Scalability Analysis}\label{sec:exp:scalability}
To evaluate the scalability of influence computations, we report in \Cref{fig:exp:runtime:datasize} the effect of dataset size on the time taken to compute influence of a subset by the approaches described in Section~\ref{sec:intervention}. We replicated \textsf{German} to increase its size by a factor of $50$ to $1,600$ yielding up to 1.6M training data points. To evaluate how dataset sizes affect influence computation runtimes, we fixed the size of the subset for which we compute the subset influence to $5\%$ (this threshold reflects our problem setting where we are interested in small fractions of training data that need attention). We observed that both FO and SO influence computations scale well in the dataset size, and achieve speed-ups of several orders of magnitude  over retraining the model or using one-step gradient descent. Note that \sys\ has an upfront cost of pre-computing the gradients of the loss function and the Hessian. Once these computations are done, the time taken to compute subset influence is negligible (as seen in Figure~\ref{fig:exp:runtime:datasize}). In contrast, model retraining to compute subset influence can be quite expensive. For example, using the feed-forward network on \textsf{Adult}, we observed that the pre-computations took $\sim 1,800s$ and the top-$3$ explanations were generated in $56s$ for a total time cost of $\sim 1,856s$. In comparison, retraining the model after removing \textit{one} subset took $>10s$. We see benefit in using \sys\ when a relatively large number of candidates are being considered to generate top-k explanations (which is the case with our datasets).
% the datasets used in this paper).

In Table~\ref{tbl:scalability:candidates}, we report the time taken to generate the top-$5$ explanations on \textsf{German} when we allow more predicates in an explanation (indicated by the level in the lattice structure), and hence consider greater number of candidates. We observe that \sys's explanation generation using the lattice structure has good scalability (runtimes of $<25$ min) for explanations with fewer predicates. In comparison, our filtering mechanism that accounts for diversity of explanations takes negligible amount of time (although considering more candidates increases the runtime of the filtering step).}
%%%%%%%%%%%%%%%%%%%%%%%%%%%%%%%%%%%%%%%%%%%%%%%%%%%%%%%%%%%%%%%%%%%%%%%%%%%%%%%%

%%%%%%%%%%%%%%%%%%%%%%%%%%%%%%%%%%%%%%%%%%%%%%%%%%%%%%%%%%%%%%%%%%%%%%%%%%%%%%%%
\begin{table}[] \centering
		\begin{tabular}{@{}lrrrrrr@{}}
		\toprule
            \textbf{Level} & 1 & 2 & 3 & 4 & 5 & 6 \\
            \toprule
            \textbf{execution (s)} & 0.03 & 0.24 & 1.59 & 17.5 & 269 & 1,472\\
            \textbf{filtering (ms)} & 36 & 45 & 70 & 75 & 76 & 90\\
            \textbf{\#candidates} & 29 & 371 & 2,770 & 13,625 & 36,704 & 62,955\\
        \hline
		\end{tabular}
	\caption{\revd{Scalability in the number of candidate patterns.}}
	\vspace{-7mm}
	\label{tbl:scalability:candidates}
\end{table}

%%%%%%%%%%%%%%%%%%%%%%%%%%%%%%%%%%%%%%%%%%%%%%%%%%%%%%%%%%%%%%%%%%%%%%%%%%%%%%%%
% \vspace{-0.4cm}
\subsection{Detecting % errors injected by
  data errors}\label{sec:exp-detect-errors}
To test the viability of \sys in detecting data errors, we applied our implementation as a detection mechanism for errors injected by
data poisoning attacks
% adversarial attacks. % (and misclassifications in \cite{techreport}).
% \boris{We mentioned different types of errors, I assume this is outdated and removed it. CHECK}
% \noindent\textbf{Data errors injected by adversarial attacks.}
% We tested whether \sys can detect errors introduced by data poisoning attacks
~\cite{mehrabi-bias, koh-data-poisoning}.  The objective was to develop techniques to detect attacks that have superior performance (measured in terms of accuracy and fairness) on training data and are targeted at exacerbating model performance on test data. The state of the art in safeguarding against data poisoning attacks is to detect anomalies that do not conform to the rest of the data. However, anomaly detection fails in the presence of sophisticated attacks that are targeted at deteriorating model accuracy and/or fairness~\cite{solans2020poisoning, jagielski2020subpopulation,mehrabi2020exacerbating,goel2021importance}. We performed experiments where we injected poisoned data points into the training data using non-random anchoring attacks~\cite{mehrabi-bias}. We found that the outlier detection mechanism supported by scikit-learn~\cite{sklearn}, LocalOutlierFactor, was not able to detect any of the poisoned data points, because they follow  a similar distribution as the original training data points. In comparison, when the data was clustered (using k-means or Gaussian mixture models clustering) and clusters were ranked in decreasing order of estimated SO influence, we observed that the top-2 clusters
% with the maximum second-order influence
contained almost $70\%$ of poisoned points. While these results are promising, a detailed study demonstrating the effectiveness of second-order influence functions in detecting adversarial attacks is out of the scope of this work and is an interesting direction for future work.

\vspace{-0.2cm}
\section{Related Work}
\label{sec:related}
Our work relates to the following lines of research.

\par
\noindent\textbf{Feature-based explanations.}
Much of XAI research focuses on explaining ML models in terms of
% patterns and
dependencies between input features and their outcomes.
% Its methods are based on
{\em Feature attribution} methods
% , such as the Shapley value,
quantify the responsibility of input features for model predictions~\cite{
% lipovetsky2001analysis,
vstrumbelj2014explaining,
lundberg2017unified,
% lundberg2018consistent,
datta2016algorithmic,merrick2019explanation,frye2019asymmetric,aas2019explaining}. Methods based on {\em surrogate explainability} approximate ML models using a simple, interpretable model
% (such as linear regression)
~\cite{lundberg2017unified,ribeiro2016should}.
% ,ribeiro2018anchors}.
{\em Contrastive and causal} methods explain ML model predictions in terms of minimal {\em interventions} or {\em perturbations} on input features that change the prediction~\cite{verma2020counterfactual,wachter2017counterfactual,
% laugel2017inverse,
karimi2019model,ustun2019actionable,mahajan2019preserving,mothilal2020explaining, DBLP:conf/sigmod/BertossiLSSV20,galhotra2021explaining}. Logic-based methods
% use tools from logic-based diagnosis that
operate on logical representations of
ML algorithms~\cite{shih2018symbolic,ignatiev2020towards,darwiche2020reasons} to compute minimal sets of features that are
sufficient and necessary for ML model predictions. These approaches fall short in generating \textit{diagnostic explanations} that help users trace an ML model's unexpected or discriminatory behavior back to its training data.

%Understanding ML model output, in particular mispredictions, has been an active area of recent research in XAI. The primary focus of prior work in this area is to reveal the internal logic of a complex decision-making model by identifying features that contribute toward particular decisions made by the algorithm, or approximating the model using a simple, interpretable model such as decision trees~\cite{lundberg2017unified,ribeiro2016should,ribeiro2018anchors}.

%-- Debugging Trusted Data items
%-- Efficient data deletion
%-- Representer points
%-- Influence functions
%-- Data Shapley
%-- Prototypes and criticisms

%While first-order approximations~\cite{pmlr-v70-koh17a} estimated how the model parameters change if a training data instance was upweighted by an infinitesimal amount, second-order influence function approximations~\cite{icml2020_4261} estimates the change when large groups of training data are removed.

\partitle{Explanations based on training data}
In contrast, data-based explanations  attribute ML model predictions to specific parts of training data~\cite{AREA-202009-Brophy}. A popular approach ranks {\em individual} training data points based on their influence on model predictions~\cite{koh2017understanding,basu2020second}
% ,yeh2018representer},
using {\em influence functions}~\cite{cook-influence}--- a classic technique from robust statistics that measure how optimal model parameters depend on training data points. Based on first-order influence functions, Rain~\cite{Wu2020ComplaintdrivenTD}
% introduced an approach for
identifies data points responsible for user constraints specified by an SQL query.
% All current approaches rank {\em individual} training data points based on their influence on ML model predictions~\cite{koh2017understanding,basu2020second,pmlr-v130-kwon21a,pmlr-v119-ghorbani20a,DBLP:conf/icml/GhorbaniZ19}
% ,yeh2018representer},
% typically using {\em influence functions}~\cite{cook-influence}. Influence functions are a classic technique from robust statistics that measure how optimal model parameters depend on training data instances. Based on first-order influence functions, \cite{Wu2020ComplaintdrivenTD} introduced an approach for identifying training data points that are responsible for user constraints specified by an SQL query.
Several recent works argue for the use of data Shapley values to quantify the contribution of individual data points~\cite{pmlr-v130-kwon21a,pmlr-v119-ghorbani20a,DBLP:conf/icml/GhorbaniZ19}, which is computationally expensive because the model needs to be retrained for each data point.
% However, there exist alternative approaches that do not use influence functions. For example,
\textit{Representer} points explain the predictions of neural networks by decomposing the pre-activation prediction into a linear combination of activations of training points~\cite{yeh2018representer}. PrIU~\cite{Wu2020PrIUAP} is a provenance-based approach that incrementally computes the effect of removing a subset of training data points.
Unlike prior methods, we generate: (1)~explanations for the bias of an ML model, (2)~interpretable explanations based on first-order predicates that pinpoint a training data subset responsible for model bias, and (3)~update-based explanations that reveal data errors in certain attributes of a training data subset.
\revc{It is worth mentioning that model bias can be reduced not only by removing subsets of training data but also by adding appropriate training samples. Searching for in-distribution data points that explain away bias upon insertion into training data is a challenging problem that we plan to explore in the future. Note that our update-based explanations act as a combination of removing existing data points and adding new ones.}

%%%%%%%%%%%%%%%%%%%%%%%%%%%%%%%%%%%%%%%%%%%%%%%%%%%%%%%%%%%%%%%%%%%%%%%%%%%%%%%%

\partitle{Debugging ML models}
Several recent works address debugging of ML models for bias, including fair-DAGs~\cite{Yang2020FairnessAwareIO} and mlinspect~\cite{10.1145/3448016.3452759} that attribute
discrimination to class imbalance
% regard class imbalance as a cause of discrimination
and address bias by tracking the distribution of sensitive attributes along ML pipelines. A closely related idea identifies regions of the input domain that are not adequately covered by training data~\cite{asudeh2021identifying, asudeh2019assessing}. MLDebugger~\cite{10.1145/3329486.3329489}
% applies a method that automatically
identifies minimal causes of unsatisfactory performance in ML pipelines using provenance of the previous runs.  These approaches cannot highlight parts of training data responsible for biased outputs.
% develop methods to identify regions of attribute space not adequately covered by training data.
DUTI~\cite{zhang2018training}
% develops an approach for
% debugs training data by
identifies
% making
the smallest set of changes to training data {\em labels} so the model trained on the updated data can correctly predict labels
of a trusted set of items from test data; this approach focuses on updating training labels, and is not applicable to our setting.  Another line of work finds data \textit{slices} in which the model performs poorly ~\cite{8713886,47966,10.1145/3448016.3457323}; slices are discovered based on their association with model error and do not capture the causal effect of interventions.  These techniques are not directly applicable
% in the context of
to
fairness where, unlike model error, bias due to a subset is not additive. Moreover, none of these
% preceding
interventions update data instances.

%In this direction, \cite{DBLP:journals/corr/abs-2102-03054} uses influence functions to identify influential training data instances responsible for mispredictions of data instances that were discriminated against.

%%%%%%%%%%%%%%%%%%%%%%%%%%%%%%%%%%%%%%%%%%%%%%%%%%%%%%%%%%%%%%%%%%%%%%%%%%%%%%%%
\partitle{Adversarial ML}
We share similarities with {\em adversarial ML} that aims to degrade ML model fairness or predictions through {\em adversarial attacks}. The most relevant classes of  attacks are based on {\em data poisoning}~\cite{steinhardt2017certified,chen2017targeted}, which injects a minimum set of synthetic data points into the training data to compromise the performance or fairness of a model trained on the contaminated data~\cite{solans2020poisoning, jagielski2020subpopulation,mehrabi2020exacerbating}. In contrast, our goal is to detect those points.
% We share similarities with {\em adversarial ML}, which explores ways in which ML can be compromised by {\em adversarial attacks}. The most relevant classes of adversarial attacks are based on {\em data poisoning}~\cite{steinhardt2017certified,chen2017targeted}, where polluting training data compromises model fairness or predictions~\cite{solans2020poisoning, jagielski2020subpopulation,mehrabi2020exacerbating}. Data poisoning attacks on fairness aim to inject a minimum set of synthetic
% % poisonous
% data points in training data that compromises the fairness of a model trained on the contaminated data.
% %\sandy{I do not understand what the following sentence means.  I assume you mean "This is in opposition to ...", but what does that mean?  Wouldn't your framework be able to highlight/identify poisoned or synthetic data points? So where is the opposition?}
% Our goal, on the other hand, is to detect those points.
% % This is in reverse to the goal of our research.
% % Note that many data poisoning attacks
% % methods
% % rely on influence functions.   %Similar techniques such as those used in this paper to approximate causal responsibility used to develop adversarial attaches of fairness too.

%which is the reverse of the goal of our research.

%%%%%%%%%%%%%%%%%%%%%%%%%%%%%%%%%%%%%%%%%%%%%%%%%%%%%%%%%%%%%%%%%%%%%%%%%%%%%%%%
\partitle{Machine unlearning}
Our research also relates to the nascent field of {\em machine unlearning}~\cite{bourtoule2019machine,gupta2021adaptive,schelter2021hedgecut,izzo2021approximate}, which addresses the ``right to be forgotten" provisioned in recent legislations, such as the General Data Protection Regulation (GDPR) in the European Union~\cite{voigt2017eu} and  the California Consumer Privacy Act in the United States~\cite{de2018guide}. Current methods are typically designed for particular classes of ML models  e.g., HedgeCut~\cite{schelter2021hedgecut} supports efficient unlearning requests for decision trees. The techniques in this sub-field could be integrated into our framework for efficient computation of causal responsibility. We defer this investigation to future study.

%%%%%%%%%%%%%%%%%%%%%%%%%%%%%%%%%%%%%%%%%%%%%%%%%%%%%%%%%%%%%%%%%%%%%%%%%%%%%%%%
\partitle{Bias detection and mitigation}
Prior work to detect and mitigate  bias in ML models
% , and
% These approaches
can be categorized into pre-, post- and in-processing methods~\cite{Caton2020FairnessIM}.
\revc{We aim to pre-process data to reduce bias by removing bias and discrimination signals from training data~\cite{NIPS2017_6988,feldman2015certifying, salimi2019interventional}, which is different from post-processing that deals with model output to handle bias, and in-processing that is aimed at building fair ML models.
% The idea is to remove bias and discrimination signals from training data by repairing or pre-processing it~\cite{NIPS2017_6988,feldman2015certifying, salimi2019interventional}.
Pre-processing methods are independent of the downstream ML model, are usually not interpretable, and hence are insufficient in generating explanations that reveal the potential source of bias.} While
% bias mitigation is
not a direct focus of our research, our approach could help develop bias mitigation algorithms that are interpretable and account for the downstream ML model; hence, they would incur minimal information loss and generalize better.  We defer this research to future work.

%\vspace{2mm}\noindent \textbf{Explanations in DBs.}
%Several works have proposed summarizing training data toward explaining ML model outcomes %e.g., Explanation Tables~\cite{10.14778/2735461.2735467},

% Amazon SageMaker: full visibility into model training by monitoring, recording, and analyzing the tensor data that captures the state of a training job

% For perturbations, mention about Koh and Liang's update

%%% Local Variables:
%%% mode: latex
%%% TeX-master: "main"
%%% End:

% \vspace{-5mm}
\section{Conclusions}\label{sec:conclusions}
We present a novel approach for debugging bias in machine learning models by identifying coherent subsets of training data that are responsible for the bias. We introduce \sys, a principled framework for reasoning about the responsibility of such subsets and develop an efficient algorithm that produces explanations, which compactly describe responsible sets of training data points through patterns. We demonstrate experimentally that \sys\ is efficient and produces explanations that are interpretable and correctly identify sources of bias for datasets where ground truth biases are well understood.
In the future, we plan to expand our approach beyond supervised ML algorithms with differentiable loss functions to support a wider range of ML algorithms such as tree-based ML models and clustering algorithms. Moreover, we plan to leverage database techniques for incremental maintenance to efficiently compute causal responsibility as opposed to approximating it. Another interesting future direction is to integrate \sys\ with database provenance to formalize the notion of provenance of ML model decisions that trace ML model outcomes all the way back to decisions made in the ML pipeline that might explain the bias and unexpected behavior of the model.

%\boris{What are interesting directions? Incremental maintenance? Tracking back into the pipeline?}

%%% Local Variables:
%%% mode: latex
%%% TeX-master: "main"
%%% End:
% \vspace{-1mm}
% \section{Acknowledgements}
% Babak acknowledges the support of the National Science Foundation (NSF) under Grant Number 2112606, and Boris acknowledges the support of NSF under Grant Numbers IIS-1956123 and IIS-2107107.
% \newpage

\bibliographystyle{plain}
\bibliography{ref}

\begin{thebibliography}{10}

\bibitem{sqf-data}
{NYPD} stop, question and frisk data.
  \url{https://www1.nyc.gov/site/nypd/stats/reports-analysis/stopfrisk.page}.
\newblock [Online; accessed 19-October-2021].

\bibitem{sqf-nyclu}
Stop-and-frisk in the de blasio era.
  \url{https://www.nyclu.org/en/publications/stop-and-frisk-de-blasio-era-2019}.
\newblock [Online; accessed 19-October-2021].

\bibitem{fb-housing}
Housing department slaps facebook with discrimination charge.
  \url{https://www.npr.org/2019/03/28/707614254/hud-slaps-facebook-with-housing-discrimination-charge},
  2019.

\bibitem{self-driving-cars}
Self-driving cars more likely to hit blacks.
  \url{https://www.technologyreview.com/2019/03/01/136808/self-driving-cars-are-coming-but-accidents-may-not-be-evenly-distributed/},
  2019.

\bibitem{aas2019explaining}
Kjersti Aas, Martin Jullum, and Anders L{\o}land.
\newblock Explaining individual predictions when features are dependent: More
  accurate approximations to shapley values.
\newblock {\em arXiv preprint arXiv:1903.10464}, 2019.

\bibitem{10.5555/645920.672836}
Rakesh Agrawal and Ramakrishnan Srikant.
\newblock Fast algorithms for mining association rules in large databases.
\newblock In {\em Proceedings of the 20th International Conference on Very
  Large Data Bases}, VLDB '94, page 487–499, San Francisco, CA, USA, 1994.
  Morgan Kaufmann Publishers Inc.

\bibitem{asudeh2019assessing}
Abolfazl Asudeh, Zhongjun Jin, and HV~Jagadish.
\newblock Assessing and remedying coverage for a given dataset.
\newblock In {\em 2019 IEEE 35th International Conference on Data Engineering
  (ICDE)}, pages 554--565. IEEE, 2019.

\bibitem{asudeh2021identifying}
Abolfazl Asudeh, Nima Shahbazi, Zhongjun Jin, and HV~Jagadish.
\newblock Identifying insufficient data coverage for ordinal continuous-valued
  attributes.
\newblock In {\em Proceedings of the 2021 International Conference on
  Management of Data}, pages 129--141, 2021.

\bibitem{icml2020_4261}
Samyadeep Basu, Xuchen You, and Soheil Feizi.
\newblock On second-order group influence functions for black-box predictions.
\newblock In {\em Proceedings of Machine Learning and Systems 2020}, pages
  7503--7512. 2020.

\bibitem{basu2020second}
Samyadeep Basu, Xuchen You, and Soheil Feizi.
\newblock On second-order group influence functions for black-box predictions.
\newblock In {\em International Conference on Machine Learning}, pages
  715--724. PMLR, 2020.

\bibitem{DBLP:conf/sigmod/BertossiLSSV20}
Leopoldo~E. Bertossi, Jordan Li, Maximilian Schleich, Dan Suciu, and Zografoula
  Vagena.
\newblock Causality-based explanation of classification outcomes.
\newblock In {\em Proceedings of the Fourth Workshop on Data Management for
  End-To-End Machine Learning, In conjunction with the 2020 {ACM} {SIGMOD/PODS}
  Conference, DEEM@SIGMOD 2020, Portland, OR, USA, June 14, 2020}, pages
  6:1--6:10, 2020.

\bibitem{binns2018s}
Reuben Binns, Max Van~Kleek, Michael Veale, Ulrik Lyngs, Jun Zhao, and Nigel
  Shadbolt.
\newblock 'it's reducing a human being to a percentage' perceptions of justice
  in algorithmic decisions.
\newblock In {\em Proceedings of the 2018 Chi conference on human factors in
  computing systems}, pages 1--14, 2018.

\bibitem{bourtoule2019machine}
Lucas Bourtoule, Varun Chandrasekaran, Christopher~A Choquette-Choo, Hengrui
  Jia, Adelin Travers, Baiwu Zhang, David Lie, and Nicolas Papernot.
\newblock Machine unlearning.
\newblock {\em arXiv preprint arXiv:1912.03817}, 2019.

\bibitem{AREA-202009-Brophy}
Jonathan Brophy.
\newblock Exit through the training data: A look into instance-attribution
  explanations and efficient data deletion in machine learning.
\newblock Area exam, University of Oregon, Computer and Information Sciences
  Department, 9 2020.
\newblock Available at
  \url{https://www.cs.uoregon.edu/Reports/AREA-202009-Brophy.pdf}.

\bibitem{Bubeck2015ConvexOA}
S{\'e}bastien Bubeck.
\newblock Convex optimization: Algorithms and complexity.
\newblock {\em Found. Trends Mach. Learn.}, 8:231--357, 2015.

\bibitem{NIPS2017_6988}
Flavio Calmon, Dennis Wei, Bhanukiran Vinzamuri, Karthikeyan
  Natesan~Ramamurthy, and Kush~R Varshney.
\newblock Optimized pre-processing for discrimination prevention.
\newblock In I.~Guyon, U.~V. Luxburg, S.~Bengio, H.~Wallach, R.~Fergus,
  S.~Vishwanathan, and R.~Garnett, editors, {\em Advances in Neural Information
  Processing Systems 30}, pages 3992--4001. Curran Associates, Inc., 2017.

\bibitem{Caton2020FairnessIM}
Simon Caton and C.~Haas.
\newblock Fairness in machine learning: A survey.
\newblock {\em ArXiv}, abs/2010.04053, 2020.

\bibitem{chen2017targeted}
Xinyun Chen, Chang Liu, Bo~Li, Kimberly Lu, and Dawn Song.
\newblock Targeted backdoor attacks on deep learning systems using data
  poisoning.
\newblock {\em arXiv preprint arXiv:1712.05526}, 2017.

\bibitem{chiappa2019path}
Silvia Chiappa.
\newblock Path-specific counterfactual fairness.
\newblock In {\em Proceedings of the AAAI Conference on Artificial
  Intelligence}, volume~33, pages 7801--7808, 2019.

\bibitem{intro-optimization}
Edwin Kah~Pin Chong and Stanislaw~H. Zak.
\newblock {\em An introduction to optimization}.
\newblock John Wiley \& Sons, 2013.

\bibitem{8713886}
Y.~Chung, T.~Kraska, N.~Polyzotis, K.~Tae, and S.~Whang.
\newblock Automated data slicing for model validation: A big data - ai
  integration approach.
\newblock {\em IEEE Transactions on Knowledge \& Data Engineering},
  32(12):2284--2296, dec 2020.

\bibitem{cook-influence}
Dennis~R. Cook and Sanford Weisberg.
\newblock Characterizations of an empirical influence function for detecting
  influential cases in regression.
\newblock {\em Technometrics}, 22, 1980.

\bibitem{darwiche2020reasons}
Adnan Darwiche and Auguste Hirth.
\newblock On the reasons behind decisions.
\newblock {\em arXiv preprint arXiv:2002.09284}, 2020.

\bibitem{datta2016algorithmic}
Anupam Datta, Shayak Sen, and Yair Zick.
\newblock Algorithmic transparency via quantitative input influence: Theory and
  experiments with learning systems.
\newblock In {\em 2016 IEEE symposium on security and privacy (SP)}, pages
  598--617. IEEE, 2016.

\bibitem{de2018guide}
Lydia de~la Torre.
\newblock A guide to the california consumer privacy act of 2018.
\newblock {\em Available at SSRN 3275571}, 2018.

\bibitem{Dua2019}
Dheeru Dua and Casey Graff.
\newblock {UCI} machine learning repository, 2017.

\bibitem{farrand2020neither}
Tom Farrand, Fatemehsadat Mireshghallah, Sahib Singh, and Andrew Trask.
\newblock Neither private nor fair: Impact of data imbalance on utility and
  fairness in differential privacy.
\newblock In {\em Proceedings of the 2020 Workshop on Privacy-Preserving
  Machine Learning in Practice}, pages 15--19, 2020.

\bibitem{feldman2015certifying}
Michael Feldman, Sorelle~A. Friedler, John Moeller, Carlos Scheidegger, and
  Suresh Venkatasubramanian.
\newblock Certifying and removing disparate impact.
\newblock In {\em {KDD}}, pages 259--268. {ACM}, 2015.

\bibitem{fernando2021missing}
Mart{\'\i}nez-Plumed Fernando, Ferri C{\`e}sar, Nieves David, and
  Hern{\'a}ndez-Orallo Jos{\'e}.
\newblock Missing the missing values: The ugly duckling of fairness in machine
  learning.
\newblock {\em International Journal of Intelligent Systems}, 2021.

\bibitem{frye2019asymmetric}
Christopher Frye, Ilya Feige, and Colin Rowat.
\newblock Asymmetric shapley values: incorporating causal knowledge into
  model-agnostic explainability.
\newblock {\em arXiv preprint arXiv:1910.06358}, 2019.

\bibitem{Fu2020AIAA}
Runshan Fu, Yan Huang, and P.~Singh.
\newblock Ai and algorithmic bias: Source, detection, mitigation and
  implications.
\newblock {\em Social Science Research Network}, 2020.

\bibitem{galhotra2021explaining}
Sainyam Galhotra, Romila Pradhan, and Babak Salimi.
\newblock Explaining black-box algorithms using probabilistic contrastive
  counterfactuals.
\newblock In {\em Proceedings of the 2021 International Conference on
  Management of Data}, pages 577--590, 2021.

\bibitem{pmlr-v119-ghorbani20a}
Amirata Ghorbani, Michael Kim, and James Zou.
\newblock A distributional framework for data valuation.
\newblock In Hal~Daumé III and Aarti Singh, editors, {\em Proceedings of the
  37th International Conference on Machine Learning}, volume 119 of {\em
  Proceedings of Machine Learning Research}, pages 3535--3544. PMLR, 13--18 Jul
  2020.

\bibitem{DBLP:conf/icml/GhorbaniZ19}
Amirata Ghorbani and James~Y. Zou.
\newblock Data shapley: Equitable valuation of data for machine learning.
\newblock In Kamalika Chaudhuri and Ruslan Salakhutdinov, editors, {\em
  Proceedings of the 36th International Conference on Machine Learning, {ICML}
  2019, 9-15 June 2019, Long Beach, California, {USA}}, volume~97 of {\em
  Proceedings of Machine Learning Research}, pages 2242--2251. {PMLR}, 2019.

\bibitem{gianfrancesco2018potential}
Milena~A Gianfrancesco, Suzanne Tamang, Jinoos Yazdany, and Gabriela Schmajuk.
\newblock Potential biases in machine learning algorithms using electronic
  health record data.
\newblock {\em JAMA internal medicine}, 178(11):1544--1547, 2018.

\bibitem{goel2021importance}
Naman Goel, Alfonso Amayuelas, Amit Deshpande, and Amit Sharma.
\newblock The importance of modeling data missingness in algorithmic fairness:
  A causal perspective.
\newblock In {\em Proceedings of the AAAI Conference on Artificial
  Intelligence}, volume~35, pages 7564--7573, 2021.

\bibitem{10.1145/3448016.3452759}
Stefan Grafberger, Shubha Guha, Julia Stoyanovich, and Sebastian Schelter.
\newblock Mlinspect: A data distribution debugger for machine learning
  pipelines.
\newblock In {\em Proceedings of the 2021 International Conference on
  Management of Data}, SIGMOD/PODS '21, page 2736–2739, New York, NY, USA,
  2021. Association for Computing Machinery.

\bibitem{gupta2021adaptive}
Varun Gupta, Christopher Jung, Seth Neel, Aaron Roth, Saeed Sharifi-Malvajerdi,
  and Chris Waites.
\newblock Adaptive machine unlearning.
\newblock {\em arXiv preprint arXiv:2106.04378}, 2021.

\bibitem{10.1145/335191.335372}
Jiawei Han, Jian Pei, and Yiwen Yin.
\newblock Mining frequent patterns without candidate generation.
\newblock {\em SIGMOD Rec.}, 29(2):1–12, May 2000.

\bibitem{ignatiev2020towards}
Alexey Ignatiev.
\newblock Towards trustable explainable ai.
\newblock In {\em 29th International Joint Conference on Artificial
  Intelligence}, pages 5154--5158, 2020.

\bibitem{izzo2021approximate}
Zachary Izzo, Mary~Anne Smart, Kamalika Chaudhuri, and James Zou.
\newblock Approximate data deletion from machine learning models.
\newblock In {\em International Conference on Artificial Intelligence and
  Statistics}, pages 2008--2016. PMLR, 2021.

\bibitem{jacobs2021measurement}
Abigail~Z Jacobs and Hanna Wallach.
\newblock Measurement and fairness.
\newblock In {\em Proceedings of the 2021 ACM Conference on Fairness,
  Accountability, and Transparency}, pages 375--385, 2021.

\bibitem{jagielski2020subpopulation}
Matthew Jagielski, Giorgio Severi, Niklas~Pousette Harger, and Alina Oprea.
\newblock Subpopulation data poisoning attacks.
\newblock {\em arXiv preprint arXiv:2006.14026}, 2020.

\bibitem{DBLP:journals/corr/abs-2006-14026}
Matthew Jagielski, Giorgio Severi, Niklas~Pousette Harger, and Alina Oprea.
\newblock Subpopulation data poisoning attacks.
\newblock {\em CoRR}, abs/2006.14026, 2020.

\bibitem{karimi2019model}
Amir-Hossein Karimi, Gilles Barthe, Borja Belle, and Isabel Valera.
\newblock Model-agnostic counterfactual explanations for consequential
  decisions.
\newblock {\em arXiv preprint arXiv:1905.11190}, 2019.

\bibitem{kasirzadeh2021reasons}
Atoosa Kasirzadeh.
\newblock Reasons, values, stakeholders: A philosophical framework for
  explainable artificial intelligence.
\newblock In {\em Proceedings of the 2021 ACM Conference on Fairness,
  Accountability, and Transparency}, pages 14--14, 2021.

\bibitem{10.1145/2702123.2702520}
Matthew Kay, Cynthia Matuszek, and Sean~A. Munson.
\newblock Unequal representation and gender stereotypes in image search results
  for occupations.
\newblock In {\em Proceedings of the 33rd Annual ACM Conference on Human
  Factors in Computing Systems}, CHI '15, page 3819–3828, New York, NY, USA,
  2015. Association for Computing Machinery.

\bibitem{kemper2019transparent}
Jakko Kemper and Daan Kolkman.
\newblock Transparent to whom? no algorithmic accountability without a critical
  audience.
\newblock {\em Information, Communication \& Society}, 22(14):2081--2096, 2019.

\bibitem{kilbertus2017avoiding}
Niki Kilbertus, Mateo~Rojas Carulla, Giambattista Parascandolo, Moritz Hardt,
  Dominik Janzing, and Bernhard Sch{\"o}lkopf.
\newblock Avoiding discrimination through causal reasoning.
\newblock In {\em Advances in Neural Information Processing Systems}, pages
  656--666, 2017.

\bibitem{pmlr-v70-koh17a}
Pang~Wei Koh and Percy Liang.
\newblock Understanding black-box predictions via influence functions.
\newblock volume~70 of {\em Proceedings of Machine Learning Research}, pages
  1885--1894, International Convention Centre, Sydney, Australia, 06--11 Aug
  2017. PMLR.

\bibitem{koh2017understanding}
Pang~Wei Koh and Percy Liang.
\newblock Understanding black-box predictions via influence functions.
\newblock In {\em International Conference on Machine Learning}, pages
  1885--1894. PMLR, 2017.

\bibitem{koh-data-poisoning}
Pang~Wei Koh, Jacob Steinhardt, and Percy Liang.
\newblock Stronger data poisoning attacks break data sanitization defenses.
\newblock {\em arXiv 2018}.

\bibitem{kusner2017counterfactual}
Matt~J Kusner, Joshua Loftus, Chris Russell, and Ricardo Silva.
\newblock Counterfactual fairness.
\newblock In {\em Advances in Neural Information Processing Systems}, pages
  4069--4079, 2017.

\bibitem{pmlr-v130-kwon21a}
Yongchan Kwon, Manuel A.~Rivas, and James Zou.
\newblock Efficient computation and analysis of distributional shapley values.
\newblock In Arindam Banerjee and Kenji Fukumizu, editors, {\em Proceedings of
  The 24th International Conference on Artificial Intelligence and Statistics},
  volume 130 of {\em Proceedings of Machine Learning Research}, pages 793--801.
  PMLR, 13--15 Apr 2021.

\bibitem{laugel2017inverse}
Thibault Laugel, Marie-Jeanne Lesot, Christophe Marsala, Xavier Renard, and
  Marcin Detyniecki.
\newblock Inverse classification for comparison-based interpretability in
  machine learning.
\newblock {\em arXiv preprint arXiv:1712.08443}, 2017.

\bibitem{lepri2018fair}
Bruno Lepri, Nuria Oliver, Emmanuel Letouz{\'e}, Alex Pentland, and Patrick
  Vinck.
\newblock Fair, transparent, and accountable algorithmic decision-making
  processes.
\newblock {\em Philosophy \& Technology}, 31(4):611--627, 2018.

\bibitem{lipovetsky2001analysis}
Stan Lipovetsky and Michael Conklin.
\newblock Analysis of regression in game theory approach.
\newblock {\em Applied Stochastic Models in Business and Industry},
  17(4):319--330, 2001.

\bibitem{10.1145/3329486.3329489}
Raoni Louren\c{c}o, Juliana Freire, and Dennis Shasha.
\newblock Debugging machine learning pipelines.
\newblock In {\em Proceedings of the 3rd International Workshop on Data
  Management for End-to-End Machine Learning}, DEEM'19, New York, NY, USA,
  2019. Association for Computing Machinery.

\bibitem{lundberg2018consistent}
Scott~M Lundberg, Gabriel~G Erion, and Su-In Lee.
\newblock Consistent individualized feature attribution for tree ensembles.
\newblock {\em arXiv preprint arXiv:1802.03888}, 2018.

\bibitem{lundberg2017unified}
Scott~M Lundberg and Su-In Lee.
\newblock A unified approach to interpreting model predictions.
\newblock In {\em Advances in neural information processing systems}, pages
  4765--4774, 2017.

\bibitem{mahajan2019preserving}
Divyat Mahajan, Chenhao Tan, and Amit Sharma.
\newblock Preserving causal constraints in counterfactual explanations for
  machine learning classifiers.
\newblock {\em arXiv preprint arXiv:1912.03277}, 2019.

\bibitem{martinez2019fairness}
Fernando Mart{\'\i}nez-Plumed, C{\`e}sar Ferri, David Nieves, and Jos{\'e}
  Hern{\'a}ndez-Orallo.
\newblock Fairness and missing values.
\newblock {\em arXiv preprint arXiv:1905.12728}, 2019.

\bibitem{mehrabi2019survey}
Ninareh Mehrabi, Fred Morstatter, Nripsuta Saxena, Kristina Lerman, and Aram
  Galstyan.
\newblock A survey on bias and fairness in machine learning.
\newblock {\em arXiv preprint arXiv:1908.09635}, 2019.

\bibitem{mehrabi-bias}
Ninareh Mehrabi, Muhammad Naveed, Fred Morstatter, and Aram Galstyan.
\newblock Exacerbating algorithmic bias through fairness attacks.
\newblock {\em To appear in Proceedings of AAAI 2021}.

\bibitem{mehrabi2020exacerbating}
Ninareh Mehrabi, Muhammad Naveed, Fred Morstatter, and Aram Galstyan.
\newblock Exacerbating algorithmic bias through fairness attacks.
\newblock {\em arXiv preprint arXiv:2012.08723}, 2020.

\bibitem{merrick2019explanation}
Luke Merrick and Ankur Taly.
\newblock The explanation game: Explaining machine learning models with
  cooperative game theory.
\newblock {\em arXiv preprint arXiv:1909.08128}, 2019.

\bibitem{molnar2020interpretable}
Christoph Molnar.
\newblock {\em Interpretable Machine Learning}.
\newblock Lulu. com, 2020.

\bibitem{mothilal2020explaining}
Ramaravind~K Mothilal, Amit Sharma, and Chenhao Tan.
\newblock Explaining machine learning classifiers through diverse
  counterfactual explanations.
\newblock In {\em Proceedings of the 2020 Conference on Fairness,
  Accountability, and Transparency}, pages 607--617, 2020.

\bibitem{nabi2018fair}
Razieh Nabi and Ilya Shpitser.
\newblock Fair inference on outcomes.
\newblock In {\em Proceedings of the... AAAI Conference on Artificial
  Intelligence. AAAI Conference on Artificial Intelligence}, volume 2018, page
  1931. NIH Public Access, 2018.

\bibitem{parikh2019addressing}
Ravi~B Parikh, Stephanie Teeple, and Amol~S Navathe.
\newblock Addressing bias in artificial intelligence in health care.
\newblock {\em Jama}, 322(24):2377--2378, 2019.

\bibitem{pytorch}
Adam Paszke, Sam Gross, Francisco Massa, Adam Lerer, James Bradbury, Gregory
  Chanan, Trevor Killeen, Zeming Lin, Natalia Gimelshein, Luca Antiga, Alban
  Desmaison, Andreas Kopf, Edward Yang, Zachary DeVito, Martin Raison, Alykhan
  Tejani, Sasank Chilamkurthy, Benoit Steiner, Lu~Fang, Junjie Bai, and Soumith
  Chintala.
\newblock Pytorch: An imperative style, high-performance deep learning library.
\newblock In H.~Wallach, H.~Larochelle, A.~Beygelzimer, F.~d\textquotesingle
  Alch\'{e}-Buc, E.~Fox, and R.~Garnett, editors, {\em Advances in Neural
  Information Processing Systems 32}, pages 8024--8035. Curran Associates,
  Inc., 2019.

\bibitem{sklearn}
F.~Pedregosa, G.~Varoquaux, A.~Gramfort, V.~Michel, B.~Thirion, O.~Grisel,
  M.~Blondel, P.~Prettenhofer, R.~Weiss, V.~Dubourg, J.~Vanderplas, A.~Passos,
  D.~Cournapeau, M.~Brucher, M.~Perrot, and E.~Duchesnay.
\newblock Scikit-learn: Machine learning in {P}ython.
\newblock {\em Journal of Machine Learning Research}, 12:2825--2830, 2011.

\bibitem{47966}
Neoklis Polyzotis, Steven Whang, Tim~Klas Kraska, and Yeounoh Chung.
\newblock Slice finder: Automated data slicing for model validation.
\newblock In {\em Proceedings of the IEEE Int' Conf. on Data Engineering
  (ICDE), 2019}, 2019.

\bibitem{rezaei2020robust}
Ashkan Rezaei, Anqi Liu, Omid Memarrast, and Brian Ziebart.
\newblock Robust fairness under covariate shift.
\newblock {\em arXiv preprint arXiv:2010.05166}, 2020.

\bibitem{ribeiro2016should}
Marco~Tulio Ribeiro, Sameer Singh, and Carlos Guestrin.
\newblock " why should i trust you?" explaining the predictions of any
  classifier.
\newblock In {\em Proceedings of the 22nd ACM SIGKDD international conference
  on knowledge discovery and data mining}, pages 1135--1144, 2016.

\bibitem{ribeiro2018anchors}
Marco~Tulio Ribeiro, Sameer Singh, and Carlos Guestrin.
\newblock Anchors: High-precision model-agnostic explanations.
\newblock In {\em AAAI}, volume~18, pages 1527--1535, 2018.

\bibitem{10.1145/3448016.3457323}
Svetlana Sagadeeva and Matthias Boehm.
\newblock Sliceline: Fast, linear-algebra-based slice finding for ml model
  debugging.
\newblock In {\em Proceedings of the 2021 International Conference on
  Management of Data}, SIGMOD/PODS '21, page 2290–2299, New York, NY, USA,
  2021. Association for Computing Machinery.

\bibitem{DBLP:conf/sigmod/SalimiGS18}
Babak Salimi, Johannes Gehrke, and Dan Suciu.
\newblock Bias in {OLAP} queries: Detection, explanation, and removal.
\newblock In {\em Proceedings of the 2018 International Conference on
  Management of Data, {SIGMOD} Conference 2018, Houston, TX, USA, June 10-15,
  2018}, pages 1021--1035, 2018.

\bibitem{salimi2019interventional}
Babak Salimi, Luke Rodriguez, Bill Howe, and Dan Suciu.
\newblock Interventional fairness: Causal database repair for algorithmic
  fairness.
\newblock In {\em Proceedings of the 2019 International Conference on
  Management of Data}, pages 793--810. ACM, 2019.

\bibitem{schelter2021hedgecut}
Sebastian Schelter, Stefan Grafberger, and Ted Dunning.
\newblock Hedgecut: Maintaining randomised trees for low-latency machine
  unlearning.
\newblock In {\em Proceedings of the 2021 International Conference on
  Management of Data}, pages 1545--1557, 2021.

\bibitem{shih2018symbolic}
Andy Shih, Arthur Choi, and Adnan Darwiche.
\newblock A symbolic approach to explaining bayesian network classifiers.
\newblock {\em arXiv preprint arXiv:1805.03364}, 2018.

\bibitem{singh2021fairness}
Harvineet Singh, Rina Singh, Vishwali Mhasawade, and Rumi Chunara.
\newblock Fairness violations and mitigation under covariate shift.
\newblock In {\em Proceedings of the 2021 ACM Conference on Fairness,
  Accountability, and Transparency}, pages 3--13, 2021.

\bibitem{solans2020poisoning}
David Solans, Battista Biggio, and Carlos Castillo.
\newblock Poisoning attacks on algorithmic fairness.
\newblock {\em arXiv preprint arXiv:2004.07401}, 2020.

\bibitem{steinhardt2017certified}
Jacob Steinhardt, Pang~Wei Koh, and Percy Liang.
\newblock Certified defenses for data poisoning attacks.
\newblock In {\em Proceedings of the 31st International Conference on Neural
  Information Processing Systems}, pages 3520--3532, 2017.

\bibitem{Stoyanovich2020ResponsibleDM}
Julia Stoyanovich, Bill Howe, and H.~V. Jagadish.
\newblock Responsible data management.
\newblock {\em Proceedings of the VLDB Endowment}, 13:3474 -- 3488, 2020.

\bibitem{vstrumbelj2014explaining}
Erik {\v{S}}trumbelj and Igor Kononenko.
\newblock Explaining prediction models and individual predictions with feature
  contributions.
\newblock {\em Knowledge and information systems}, 41(3):647--665, 2014.

\bibitem{TAGH+17}
Florian Tram{\`e}r, Vaggelis Atlidakis, Roxana Geambasu, Daniel Hsu,
  Jean-Pierre Hubaux, Mathias Humbert, Ari Juels, and Huang Lin.
\newblock Fairtest: Discovering unwarranted associations in data-driven
  applications.
\newblock In {\em IEEE European Symposium on Security and Privacy (EuroS\&P)}.
  IEEE, 2017.

\bibitem{fastai}
\url{https://docs.fast.ai/tabular.learner.htm}.
\newblock Fastai neural network.

\bibitem{ustun2019actionable}
Berk Ustun, Alexander Spangher, and Yang Liu.
\newblock Actionable recourse in linear classification.
\newblock In {\em Proceedings of the Conference on Fairness, Accountability,
  and Transparency}, pages 10--19, 2019.

\bibitem{verma2020counterfactual}
Sahil Verma, John Dickerson, and Keegan Hines.
\newblock Counterfactual explanations for machine learning: A review.
\newblock {\em arXiv preprint arXiv:2010.10596}, 2020.

\bibitem{10.1145/3194770.3194776}
Sahil Verma and Julia Rubin.
\newblock Fairness definitions explained.
\newblock In {\em Proceedings of the International Workshop on Software
  Fairness}, FairWare '18, page 1–7, New York, NY, USA, 2018. Association for
  Computing Machinery.

\bibitem{voigt2017eu}
Paul Voigt and Axel Von~dem Bussche.
\newblock The eu general data protection regulation (gdpr).
\newblock {\em A Practical Guide, 1st Ed., Cham: Springer International
  Publishing}, 10:3152676, 2017.

\bibitem{wachter2017counterfactual}
Sandra Wachter, Brent Mittelstadt, and Chris Russell.
\newblock Counterfactual explanations without opening the black box: Automated
  decisions and the gdpr.
\newblock {\em Harv. JL \& Tech.}, 31:841, 2017.

\bibitem{wang2021fair}
Jialu Wang, Yang Liu, and Caleb Levy.
\newblock Fair classification with group-dependent label noise.
\newblock In {\em Proceedings of the 2021 ACM Conference on Fairness,
  Accountability, and Transparency}, pages 526--536, 2021.

\bibitem{Wu2020ComplaintdrivenTD}
Weiyuan Wu, Lampros Flokas, Eugene Wu, and Jiannan Wang.
\newblock Complaint-driven training data debugging for query 2.0.
\newblock {\em Proceedings of the 2020 ACM SIGMOD International Conference on
  Management of Data}, 2020.

\bibitem{Wu2020PrIUAP}
Yinjun Wu, V.~Tannen, and S.~Davidson.
\newblock Priu: A provenance-based approach for incrementally updating
  regression models.
\newblock {\em Proceedings of the 2020 ACM SIGMOD International Conference on
  Management of Data}, 2020.

\bibitem{Yang2020FairnessAwareIO}
K.~Yang, Biao Huang, Julia Stoyanovich, and Sebastian Schelter.
\newblock Fairness-aware instrumentation of preprocessing~pipelines for machine
  learning.
\newblock 2020.

\bibitem{yeh2018representer}
Chih-Kuan Yeh, Joon~Sik Kim, Ian~EH Yen, and Pradeep Ravikumar.
\newblock Representer point selection for explaining deep neural networks.
\newblock {\em arXiv preprint arXiv:1811.09720}, 2018.

\bibitem{zhang2018training}
Xuezhou Zhang, Xiaojin Zhu, and Stephen Wright.
\newblock Training set debugging using trusted items.
\newblock In {\em Thirty-second AAAI conference on artificial intelligence},
  2018.

\end{thebibliography}
\end{document}